\documentclass{article}

\usepackage{iclr2024_conference,times}
\iclrfinalcopy 

\usepackage[utf8]{inputenc} 
\usepackage[T1]{fontenc}    
\usepackage{hyperref}       
\usepackage{url}            
\usepackage{booktabs}       
\usepackage{amsfonts}       
\usepackage{amsmath}
\usepackage{nicefrac}       
\usepackage{microtype}      
\usepackage{xcolor}         
\usepackage{graphicx}
\usepackage{comment}
\usepackage{multirow}
\usepackage{threeparttable}
\usepackage{tabularx}
\usepackage{arydshln}
\usepackage{placeins}
\usepackage{makecell}
\usepackage{threeparttablex}
\usepackage{svg}
\usepackage{adjustbox}
\usepackage{pythonhighlight}
\usepackage[acronym,shortcuts]{glossaries}
\usepackage{listings}
\usepackage{natbib}[maxnames=1]
\lstdefinestyle{pycharm}{
  language=Python,
  basicstyle=\ttfamily\small,
  keywordstyle=[1]\color{blue}\bfseries,
  keywordstyle=[2]\color{cyan}\bfseries,
  keywordstyle=[3]\color{violet}\bfseries,
  stringstyle=\color{green!70!black},
  commentstyle=\color{gray},
  morecomment=[l][\color{orange}]{@}
  showstringspaces=false,
  numbers=right,
  numberstyle=\tiny\ttfamily\color{gray},
  numbersep=5pt,
  tabsize=2,
  breaklines=true,
  morekeywords=[1]{import,from,class,def,return,if,elif,else,for,in,while,with,as,is,not,or,and,pass,break,continue,global,nonlocal,try,except,finally,raise,assert},
  morekeywords=[2]{True,False,None},
  morekeywords=[3]{int,float,str,list,tuple,dict,set,bool},
  emph={self},
  emphstyle=\color{red}\bfseries,
  basicstyle=\ttfamily
}
\definecolor{delim}{RGB}{20,105,176}
\definecolor{numb}{RGB}{106, 109, 32}
\definecolor{string}{rgb}{0.64,0.08,0.08}

\lstdefinelanguage{json}{
    numbers=right,
    numberstyle=\tiny\ttfamily\color{gray},
    frame=single,
    rulecolor=\color{black},
    showspaces=false,
    showtabs=false,
    breaklines=true,
    postbreak=\raisebox{0ex}[0ex][0ex]{\ensuremath{\color{gray}\hookrightarrow\space}},
    breakatwhitespace=true,
    basicstyle=\ttfamily\small,
    upquote=true,
    morestring=[b]",
    stringstyle=\color{string},
    literate=
     *{0}{{{\color{numb}0}}}{1}
      {1}{{{\color{numb}1}}}{1}
      {2}{{{\color{numb}2}}}{1}
      {3}{{{\color{numb}3}}}{1}
      {4}{{{\color{numb}4}}}{1}
      {5}{{{\color{numb}5}}}{1}
      {6}{{{\color{numb}6}}}{1}
      {7}{{{\color{numb}7}}}{1}
      {8}{{{\color{numb}8}}}{1}
      {9}{{{\color{numb}9}}}{1}
      {\{}{{{\color{delim}{\{}}}}{1}
      {\}}{{{\color{delim}{\}}}}}{1}
      {[}{{{\color{delim}{[}}}}{1}
      {]}{{{\color{delim}{]}}}}{1},
}

\usepackage{array}
\usepackage[tableposition=top]{caption}
\usepackage{subcaption}
\usepackage{colortbl}
\usepackage{rotating}
\usepackage{tabularray}
\usepackage{orcidlink}
\usepackage{wrapfig}
\DeclareCaptionFormat{custom}
{
\small\textsc{#1#2}#3
}
\usepackage{pifont}
\newcommand{\cmark}{\ding{51}}%
\definecolor{lightergray}{gray}{0.95}
\newcommand{\xmark}{{\color{lightergray}\ding{55}}}%
\svgpath{{../figures/}}
\title{Yet Another ICU Benchmark:
A Flexible \\ Multi-Center Framework for Clinical ML}

\definecolor{Gray}{gray}{0.9}
\newcolumntype{g}{>{\columncolor{Gray}}c}

\author{Robin van de Water$^1$\orcidlink{0000-0002-2895-4872}\thanks{Corresponding author email: robin.vandewater@hpi.de} \qquad 
   Hendrik Schmidt$^1$\orcidlink{0009-0007-9501-7567}\qquad
   Paul Elbers$^2$\orcidlink{0000-0003-0447-6893}\\
   \textbf{Patrick Thoral$^2$\orcidlink{0000-0001-6140-7195}}\qquad
   \textbf{Bert Arnrich$^1$\orcidlink{0000-0001-8380-7667}}\qquad
   \textbf{Patrick Rockenschaub$^3$\orcidlink{0000-0002-6499-7933}}\\ 
   $^1$Hasso Plattner Institute, University of Potsdam, Germany\\
    $^2$Amsterdam UMC, Vrije Universiteit, Amsterdam, The Netherlands\\
   $^3$Lab for AI in Medicine, Charité - Universitätsmedizin Berlin, Germany\\
}

\begin{document}
\newacronym{icu}{ICU}{intensive care unit}
\newacronym{ehr}{EHR}{Electronic Health Record}

\newacronym{ml}{ML}{machine learning}
\newacronym{dl}{DL}{deep learning}


\newacronym{cnn}{CNN}{Convolutional Neural Network}
\newacronym{fnn}{FNN}{Feed-forward Neural Network}
\newacronym{rnn}{RNN}{Recurrent Neural Network}
\newacronym{lstm}{LSTM}{Long Short-Term Memory}
\newacronym{gru}{GRU}{Gated Recurrent Unit}
\newacronym{tcn}{TCN}{Temporal Convolutional Network}
\newacronym{lgbm}{LGBM}{Light Gradient Boosting Machine}
\newacronym{en}{EN}{elastic net}
\newacronym{lr}{LR}{logistic regression}
\newacronym{rf}{RF}{random forest}
\newacronym{da}{DA}{Domain Adaptation}
\newacronym{sda}{SDA}{Supervised Domain Adaptation}
\newacronym{uda}{UDA}{Unsupervised Domain Adaptation}
\newacronym{dg}{DG}{Domain Generalization}
\newacronym{tf}{TF}{transformer}
\newacronym{yaib}{\textsl{YAIB}}{\textsl{Yet Another ICU Benchmark}}

\newacronym{aki}{AKI}{acute kidney injury}
\newacronym{los}{LoS}{length of stay}
\newacronym{kf}{KF}{kidney function}
\newacronym{format}{HDF}{Harmonized Data Format}
\newacronym{auroc}{AUROC}{Area Under the Reciever Operating Characteristic}
\newacronym{auprc}{AUPRC}{Area Under the Precision Recall Curve}
\newacronym{mimic}{MIMIC}{Medical Information Mart for Intensive Care}
\newacronym{miiv}{MIMIC-IV}{Medical Information Mart for Intensive Care IV}
\newacronym{miiii}{MIMIC-III}{Medical Information Mart for Intensive Care III}
\newacronym{eicu}{eICU}{eICU Collaborative Research Database}
\newacronym{hirid}{HiRID}{High Time Resolution ICU Dataset}
\newacronym{aumc}{AUMCdb}{AmsterdamUMCdb}
\newacronym{sicdb}{SICdb}{Salzburg Intensive Care Database}

\newacronym{roc}{ROC}{Receiver Operating Characteristic}
\newacronym{prc}{PRC}{Precision Recall Curve}
\newacronym{json}{JSON}{JavaScript Object Notation}
\newacronym{wandb}{WandB}{Weights and Biases}
\newacronym{mae}{MAE}{Mean Absolute Error}
\newacronym{pl}{PL}{Pytorch-lighting}
\newacronym{csdi}{CSDI}{Conditional Score-based Diffusion models for Imputation}

\glsdisablehyper
\newcommand{\pycu}{\textsc{PyICU}}
\newcommand{\preproc}{\textsc{ReciPys}}
\newcommand{\bench}{\textsc{YAIB}}
\newcommand{\format}{harmonized ICU data}
\newcommand{\lt}{\ensuremath <}
\newcommand{\gt}{\ensuremath >}
\begin{center}
    \maketitle
\end{center}
\defcitealias{savelievTemporAIFacilitatingMachine2023}{Saveliev et al.}
\defcitealias{vandewaterClosingGapsImputation2023}{van de Water et al. (2023)}


\begin{abstract}
Medical applications of \acrfull{ml} have experienced a surge in popularity in recent years. The intensive care unit (ICU) is a natural habitat for ML given the abundance of available data from electronic health records. Models have been proposed to address numerous ICU prediction tasks like the early detection of complications. While authors frequently report state-of-the-art performance, it is challenging to verify claims of superiority. Datasets and code are often not published, and cohort definitions, preprocessing pipelines, and training setups are difficult to reproduce. This work introduces \acrfull{yaib}, a modular framework that allows researchers to define reproducible and comparable clinical \acrshort{ml} experiments; we offer an end-to-end solution from cohort definition to model evaluation.
The framework natively supports most open-access ICU datasets (\acrshort{mimic} III/IV, \acrshort{eicu}, \acrshort{hirid}, \acrshort{aumc}) and is easily adaptable to future and custom ICU datasets. Combined with a transparent preprocessing pipeline and extensible training code for multiple \acrshort{ml} and deep learning models, \acrshort{yaib} enables unified model development, transfer, and evaluation. Our benchmark comes with five predefined established prediction tasks (mortality, acute kidney injury, sepsis, kidney function, and length of stay) developed in collaboration with clinicians. Adding further tasks is straightforward by design. 
Using \acrshort{yaib}, we demonstrate that the choice of dataset, cohort definition, and preprocessing have a major impact on the prediction performance, often more so than model class, indicating an urgent need for \acrshort{yaib} as a holistic benchmarking tool.
We provide our work to the clinical \acrshort{ml} community to accelerate method development and enable real-world implementations.\\
\textbf{Software Repository:} \url{https://github.com/rvandewater/YAIB}

\end{abstract}
\vspace{-2mm}
\section{Introduction}
\vspace{-2mm}

\label{sec:introduction}
The \acrfull{icu} has long been a focus for research into data-driven decision support, owing to the impact of medical decisions as well as the breadth and depth of data collected in this setting~\citep{johnsonReproducibilityCriticalCare2017}. The COVID-19 pandemic confirmed the need for reliable \acrfull{ml}-based clinical decision support that can alert healthcare professionals to worsening patient states, help them make a clinical diagnosis, or recommend treatment \citep{medicEvidencebasedClinicalDecision2019}. 

Despite a steep increase in the number of published \acrshort{icu} prediction models \citep{shillanUseMachineLearning2019}, hardly any have made their way into clinical practice \citep{eini-poratTellMeSomething2022, fleurenMachineLearningIntensive2020}. A major obstacle to translation is an ongoing lack of comparability and reproducibility \citep{johnsonReproducibilityCriticalCare2017}. By using custom datasets and definitions, preprocessing pipelines, and evaluation schemes, the benefits of novel models are conflated with differences between patient case mix, task definitions, and cohort selection \citep{sarwarSecondaryUseElectronic2023,kellyKeyChallengesDelivering2019}. Reviewing models for early prediction of sepsis, for example, \cite{moorEarlyPredictionSepsis2021} found that the definition of sepsis, the time of prediction, and the available features differed substantially between the 22 included studies; similar results were found in an earlier review~\citep{fleurenMachineLearningPrediction2020}. Even among studies from the same research group~\citep{hylandEarlyPredictionCirculatory2020, yecheHiRIDICUBenchmarkComprehensiveMachine2022}, cohort definitions may vary substantially, precluding a meaningful comparison. Inconsistencies in imputation and feature extraction further complicate an objective evaluation of research progress.

The increasing availability of open-access \acrshort{icu} datasets is a first, important step towards urgently needed model comparability \citep{sauerLeveragingElectronicHealth2022}. However, models derived from the same dataset may still vary considerably in their analytical setup. Earlier work has therefore created benchmarks that establish a single pipeline for preprocessing and modeling \citep{yecheHiRIDICUBenchmarkComprehensiveMachine2022, harutyunyanMultitaskLearningBenchmarking2019}
. These benchmarks are hard-coded for a given dataset, following proprietary formats and supporting a limited, fixed set of tasks. Extending an existing benchmark to include new datasets or tasks requires changes to the benchmark's --- often lightly documented --- source code. Despite the existence of multiple benchmarks, new models are therefore rarely evaluated on more than one dataset or do not use \textit{any} benchmark~\citep{shillanUseMachineLearning2019}. 

We address this gap by providing \acrfull{yaib} as a modular multi-dataset framework specifically designed for extensibility. Building on recent work to harmonize \acrshort{icu} data \citep{bennettRicuInterfaceIntensive2023} (i.e., match time-scale, clinical definitions, and units across datasets), we standardize the entire modeling workflow from the definition of clinical concepts (a medical abstraction to facilitate patient care) and data extraction to model fitting and evaluation across several established open-source \acrshort{icu} datasets \citep{sauerLeveragingElectronicHealth2022}. We provide a predefined set of common prediction tasks, developed in collaboration with clinical intensivists, that can be easily extended to fit user needs. Our benchmark, by default, provides endpoint prediction for ICU mortality, sepsis~\citep{singerThirdInternationalConsensus2016}, \acrfull{aki} ~\citep{kdigoKidneyDiseaseImproving2012}, \acrfull{kf}, and \acrfull{los}. With this work, we aim to \textbf{(1)} dramatically reduce the overhead of developing new ICU prediction methods, \textbf{(2)} provide a transparent, open-source, and reproducible definition of experiments, and \textbf{(3)} unify \acrshort{ml} workflows for ICU prediction modeling.   
\vspace{-2mm}
\section{Related work}
\vspace{-2mm}

Our work builds upon several previous efforts to harmonize the definition, development, and evaluation of \acrshort{icu} prediction models. \acrshort{yaib} combines these existing works in a novel, end-to-end fashion to enable quick, reproducible, and comparable model development.

\textbf{Publicly available ICU datasets} \hspace{1em} Our benchmark currently supports four established \acrshort{icu} datasets ~\citep{sauerSystematicReviewComparison2022}: the \acrfull{mimic} version III~\citep{johnsonMIMICIIIFreelyAccessible2016} and IV~\citep{johnsonMIMICIVFreelyAccessible2023}, the \acrfull{eicu}~\citep{pollardEICUCollaborativeResearch2018}, the \acrfull{hirid}~\citep{hylandEarlyPredictionCirculatory2020}, and the \acrfull{aumc}~\citep{thoralSharingICUPatient2021}. These datasets contain similar data items but differ in size and scope (\autoref{tab:dataset_comparison_extended}). Together, they cover 334,812 \acrshort{icu} stays. 
We plan to integrate two recently released ICU datasets in the future~\citep{rodemundSalzburgIntensiveCare2023, jinEstablishmentChineseCritical2023}. 

\defcitealias{savelievTemporAIFacilitatingMachine2023}{Saveliev et al.}
\begin{table}[t]
\caption{\textit{Comparison of existing benchmarks and \acrshort{yaib} on \acrshort{icu} data, ordered by publication date.} }
    \centering
    \small
    \begin{threeparttable}
    \begin{tabularx}{\textwidth}{|>{\hsize=0.1\hsize\linewidth=\hsize}X|X|lcccccccccccccc>{\hsize=0.01\hsize\linewidth=\hsize\centering\arraybackslash\bfseries}c|}
        \multicolumn{3}{c<{\tiny}|}{}
        & \rotatebox[origin=l]{90}{\citeauthor{johnsonReproducibilityCriticalCare2017}}
        & \rotatebox[origin=l]{90}{\citeauthor{purushothamBenchmarkingDeepLearning2018}}
        & \rotatebox[origin=l]{90}{\citeauthor{harutyunyanMultitaskLearningBenchmarking2019}}
        & \rotatebox[origin=l]{90}{\citeauthor{barbieriBenchmarkingDeepLearning2020}}
        & \rotatebox[origin=l]{90}{\citeauthor{wangMIMICExtractDataExtraction2020}}
        & \rotatebox[origin=l]{90}{\citeauthor{jarrettCLAIRVOYANCEPIPELINETOOLKIT2021}}
        & \rotatebox[origin=l]{90}{\citeauthor{sheikhalishahiBenchmarkingMachineLearning2020} }
        & \rotatebox[origin=l]{90}{\citeauthor{tangDemocratizingEHRAnalyses2020}}
        & \rotatebox[origin=l]{90}{\citeauthor{yecheHiRIDICUBenchmarkComprehensiveMachine2022}}
        & \rotatebox[origin=l]{90}{\citeauthor{mandyamCOPECATCleaningOrganization2021}}
        & \rotatebox[origin=l]{90}{\citeauthor{guptaExtensiveDataProcessing2022a}}
        & \rotatebox[origin=l]{90}{\citeauthor{yangPyHealthDeepLearning2023}}
        & \rotatebox[origin=l]{90}{\citetalias{savelievTemporAIFacilitatingMachine2023}}
        & \rotatebox[origin=l]{90}{\citeauthor{oliverIntroducingBlendedICUDataset2023}}
        & \rotatebox[origin=l]{90}{YAIB (ours)}
        \\
        \toprule
        \parbox[t]{2mm}{\multirow{5}{*}{\rotatebox[origin=c]{90}{\textbf{Datasets}}}}
        & \multicolumn{2}{l|}{\acrshort{miiii}}     & \cmark & \cmark & \cmark & \cmark & \cmark & \cmark &  \xmark & \cmark &  \xmark &\xmark & \xmark &\cmark &\xmark & \cmark& \cmark\\
        & \multicolumn{2}{l|}{\acrshort{miiv}}      & \xmark & \xmark & \xmark & \xmark  &  \xmark &  \xmark & \xmark  &  \xmark &  \xmark &\cmark & \cmark &\cmark &\xmark & \cmark& \cmark\\
        & \multicolumn{2}{l|}{\acrshort{eicu}}      & \xmark  & \xmark  & \xmark  & \xmark  & \xmark  & \xmark  & \cmark & \cmark & \xmark &\xmark & \xmark & \cmark &\xmark & \cmark&\cmark\\
        & \multicolumn{2}{l|}{\acrshort{hirid}}     & \xmark & \xmark & \xmark & \xmark & \xmark & \xmark & \xmark & \xmark & \cmark &\xmark & \xmark &\xmark &\xmark & \cmark& \cmark\\
        & \multicolumn{2}{l|}{\acrshort{aumc}}      & \xmark & \xmark & \xmark & \xmark & \xmark & \xmark & \xmark & \xmark & \xmark &\xmark & \xmark &\xmark &\xmark & \cmark& \cmark\\
        \cline{1-18}

        \parbox[t]{2mm}{\multirow{10}{*}{\rotatebox[origin=c]{90}{\textbf{Prediction tasks}}}}
        & \multicolumn{2}{l|}{Mortality risk}        & \cmark & \cmark & \cmark & \xmark & \cmark & \xmark & \cmark & \cmark & \cmark &\cmark & \xmark &\cmark & \xmark &\xmark & \cmark\\
        & \multicolumn{2}{l|}{Circulatory failure}   & \xmark & \xmark & \xmark & \xmark & \xmark & \xmark & \xmark & \xmark & \cmark &\xmark &\cmark & \xmark & \xmark &\xmark & *\\
        & \multicolumn{2}{l|}{\Acf{kf}}       & \xmark & \xmark & \xmark & \xmark & \xmark & \xmark & \xmark & \xmark & \cmark &\xmark & \cmark &\xmark & \xmark &\xmark & \cmark\\
        & \multicolumn{2}{l|}{Respiratory failure}   & \xmark & \xmark & \xmark & \xmark & \xmark & \cmark & \xmark & \cmark & \cmark &\xmark & \cmark & \xmark & \xmark &\xmark & *\\
        & \multicolumn{2}{l|}{Sepsis}                & \xmark & \xmark & \xmark & \xmark & \xmark & \xmark & \xmark & \xmark & \xmark &\xmark &\xmark &\xmark & \xmark & \xmark &\cmark\\
        & \multicolumn{2}{l|}{\Acrlong{aki}}             & \xmark & \xmark & \xmark & \xmark & \xmark & \xmark & \xmark & \xmark & \xmark &\xmark & \xmark & \xmark & \xmark &\xmark &\cmark\\
        & \multicolumn{2}{l|}{Phenotyping$^\S$}           & \xmark & \cmark & \cmark & \xmark & \xmark & \xmark & \cmark & \xmark & \cmark &\xmark & \xmark &\xmark & \xmark &\xmark & *\\
        & \multicolumn{2}{l|}{Interventions}         & \xmark & \xmark & \xmark & \xmark & \cmark & \cmark & \xmark & \xmark & \xmark &\xmark & \xmark & \xmark & \xmark &\xmark &*\\
        & \multicolumn{2}{l|}{\Acf{los}}        & \xmark & \cmark & \cmark & \xmark & \cmark & \xmark & \cmark & \xmark & \cmark &\xmark & \xmark &\cmark & \xmark &\xmark & \cmark\\
        & \multicolumn{2}{l|}{Readmission$^\S$}           & \xmark & \xmark & \xmark & \cmark & \xmark & \xmark & \xmark & \xmark & \xmark &\xmark & \xmark &\cmark & \xmark &\xmark & *\\
        \cline{1-18}
        \parbox[t]{2mm}{\multirow{4}{*}{\rotatebox[origin=c]{90}{\textbf{Preproc.}}}}
         & \multicolumn{2}{l|}{Feature engineering} & \xmark &\cmark &\cmark &\cmark &\cmark &\cmark & \xmark & \xmark &\cmark &\cmark &\cmark & \xmark &\cmark &\xmark &\cmark\\
         & \multicolumn{2}{l|}{Temporal imputation} & \xmark & \xmark & \xmark & \xmark &\cmark & \cmark & \xmark &\cmark &\cmark &\cmark &\cmark & \xmark &\cmark &\xmark &\cmark\\
         & \multicolumn{2}{l|}{Temporal resampling} & \xmark & \xmark & \xmark & \xmark & \xmark & \xmark & \xmark &\cmark &\cmark &\xmark &\cmark & \xmark &\xmark &\xmark &\cmark\\
         & \multicolumn{2}{l|}{Modular pipeline} & \xmark & \xmark & \xmark & \xmark & \xmark & \cmark & \xmark & \cmark & \xmark &\xmark &\cmark &\xmark &\cmark &\xmark & \cmark\\
        \cline{1-18}
        \parbox[t]{2mm}{\multirow{8}{*}{\rotatebox[origin=c]{90}{\textbf{Model architectures}}}}
            & \multirow{3}{*}{\shortstack[l]{ML}}
        & LR & \cmark & \cmark & \cmark & \cmark & \cmark & \xmark & \cmark & \cmark & \cmark &\xmark & \cmark &\xmark &\xmark &\xmark & \cmark\\
        && Random forest                  & \xmark & \cmark & \xmark & \xmark & \cmark & \xmark & \xmark & \cmark & \xmark &\xmark & \cmark &\xmark &\xmark & \xmark &\cmark \\
        && Gradient boost                 & \cmark & \cmark & \xmark & \xmark & \xmark & \xmark & \xmark & \xmark & \cmark & \cmark &\cmark &\xmark &\xmark & \xmark &\cmark\\
        \cline{2-18}
        & \parbox[t]{22mm}{\multirow{5}{*}{\shortstack[l]{DL}}}
        & \acrshort{rnn}                     & \xmark & \cmark & \xmark & \cmark & \xmark & \cmark & \xmark & \xmark & \xmark &\xmark & \xmark &\cmark &\cmark & \xmark &\cmark \\
        && \acrshort{lstm}                   & \xmark & \xmark & \cmark & \xmark & \xmark & \cmark & \cmark & \cmark & \cmark & \xmark &\cmark & \cmark &\cmark &\xmark &\cmark\\
        && \acrshort{gru}                    & \xmark & \xmark & \xmark & \xmark & \cmark & \cmark & \xmark & \xmark & \cmark & \xmark &\xmark &\cmark &\cmark &\xmark & \cmark\\
        && Temporal CNN                     & \xmark & \xmark & \xmark & \xmark & \xmark & \cmark & \xmark & \cmark & \cmark & \xmark &\cmark &\cmark &\cmark & \xmark &\cmark\\
        && Transformer                       & \xmark & \xmark & \xmark & \cmark  & \xmark & \cmark & \xmark & \xmark & \cmark & \xmark &\xmark &\cmark &\cmark &\xmark & \cmark\\
        \cline{1-18}

        \multicolumn{3}{|l|}{Code available} & \xmark & \cmark & \cmark & \cmark & \cmark & \cmark & \cmark & \cmark & \cmark & \cmark &\cmark &\cmark &\cmark &\cmark & \cmark\\
        \cline{1-18}
        \multicolumn{3}{|l|}{Extensible$^\dagger$}     & \xmark & \xmark& \xmark& \xmark & \xmark & \cmark & \xmark & \xmark & \xmark &\xmark & \xmark &\xmark &\cmark &\xmark & \cmark\\
        \cline{1-18}
        \multicolumn{3}{|l|}{Dataset interoperability$^\ddagger$}     & \xmark & \xmark& \xmark& \xmark & \xmark & \xmark & \xmark & \xmark & \xmark &\xmark & \xmark &\xmark &\xmark &\xmark & \cmark\\
        \bottomrule
    \end{tabularx}
        \end{threeparttable}
        \begin{tablenotes}[flushleft]\setlength\labelsep{0pt} 
        \scriptsize
        \item *: These tasks are not included by default but may be easily added through our cohort definition pipeline. 
        \item $\S$: Due to lack of recorded database information, these tasks can only be defined for \acrshort{mimic} III and IV.
        \item $\dagger$: Interface and extensive instructions to add interoperable modules following a provided abstraction (datasets, prediction tasks, models) and adjust existing modules without extensive rewriting or refactoring.
        \item$\ddagger$: Provides an uncoupled interoperable dataset definition, allowing a.o. transfer learning and domain adaption. 
        \end{tablenotes}
        \vspace{-5mm}
    \label{tab:benchmark_comparison}
\end{table}

\textbf{Benchmarks} \hspace{1em}
To improve comparability between models trained on these \acrshort{icu} datasets, several benchmarks or benchmark-like applications have been developed 
 (\autoref{tab:benchmark_comparison}). These solutions mainly differ in the tasks and models they support. Notably, existing benchmarks heavily focus on benchmarking results, often hardcoding key steps like data extraction, task definition, preprocessing, feature generation, and sometimes model training. While they may reduce implementation overhead when evaluating new ML approaches, present benchmarks are difficult to adapt to user requirements. Core code base changes are often necessary if the users' problems do not fit into the provided task definitions. Even advanced modeling frameworks such as \cite{jarrettCLAIRVOYANCEPIPELINETOOLKIT2021} and \cite{savelievTemporAIFacilitatingMachine2023} share this weakness, as they do not support reproducible data extraction or task definitions; thus, they do not provide an end-to-end solution like \acrshort{yaib}.

\textbf{Multi-dataset support} \hspace{1em} Due to considerable heterogeneity in data structure, existing benchmarks tend to focus on a single dataset, most frequently \acrshort{miiii}. 
As \acrshort{miiii} also has a large existing user base~\citep{syedApplicationMachineLearning2021}, it thus often becomes the default choice~\citep{shillanUseMachineLearning2019}. This has potentially resulted in a self-enforcing bias towards the \acrshort{miiii} datasets, which represent a single-center US population. Even frameworks that work with its successor \acrshort{miiv} lack backward compatibility \citep{mandyamCOPECATCleaningOrganization2021, guptaExtensiveDataProcessing2022a}. Among the few multi-dataset solutions, \citep{tangDemocratizingEHRAnalyses2020} operates on both \acrshort{eicu} and \acrshort{miiii}, but lacks many of the model architectures found in others works and does no longer appear to be in active development. \cite{oliverIntroducingBlendedICUDataset2023} provides a hardcoded pipeline to combine several datasets without providing cohort definitions, benchmarking, or an end-to-end pipeline. Finally, \cite{yangPyHealthDeepLearning2023} recently proposed PyHealth as a comprehensive deep learning toolkit for both ML researchers and healthcare practitioners; it is perhaps most closely related to our work. 
Unfortunately, PyHealth only supports subsets of the full datasets, and tasks must be defined anew for each dataset. It also does not currently include time series or ways to deal with missing data, limiting its use for novel clinical or ML developments.

\section{Benchmark design}
\label{sec:design}
\begin{figure}
    \centering
    \includegraphics[width=\textwidth]{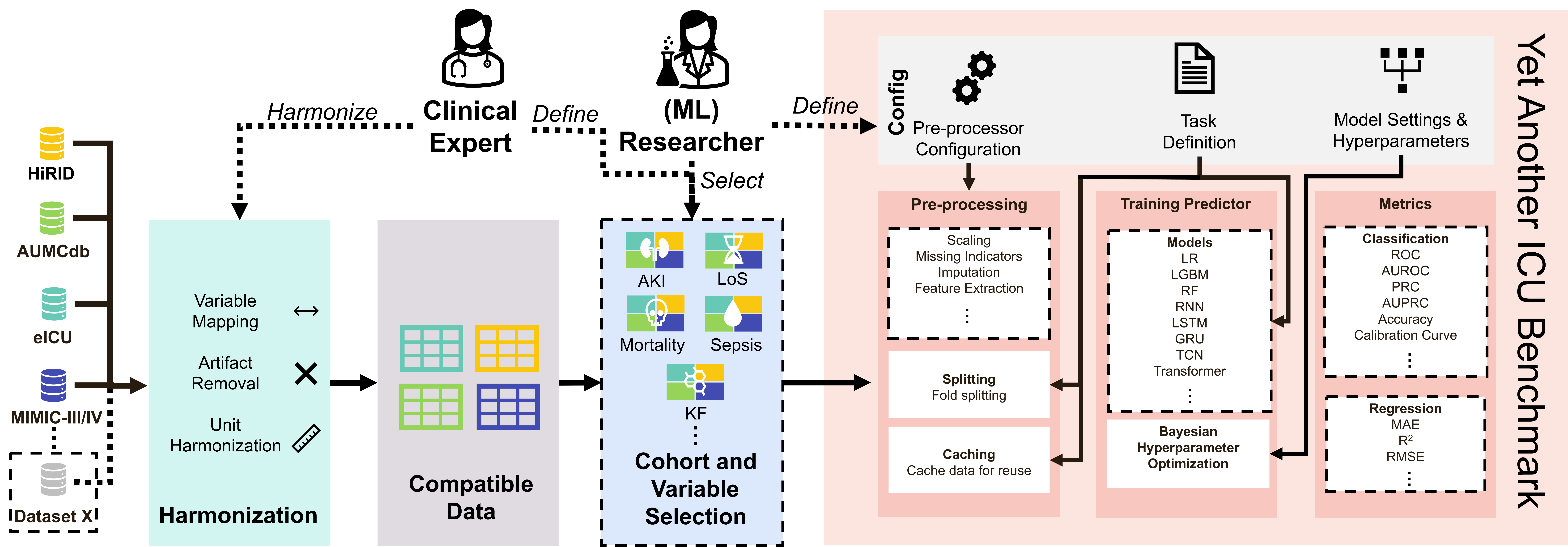}
 \caption{\textit{Schematic overview of benchmark pipeline.} On the left side, the creation of harmonized ICU cohorts is shown. Note that the domain expertise of clinicians is often necessary for defining clinically useful tasks. The schematic overview of the benchmark stages can be found on the right. Note that the dotted line indicates that this component can be easily extended, as it follows an abstracted interface.
 \vspace*{-5mm}
}
\label{fig:yaib_flow}
\end{figure}

\acrshort{yaib} addresses the issues identified above and provides a unified interface to develop clinical prediction models for the ICU. An experiment in \acrshort{yaib} consists of four steps: \textbf{1)} define clinical concepts from the raw data; \textbf{2)} extract the patient cohort and specify the prediction task; \textbf{3)} preprocess the data and generate features; and \textbf{4)} train and evaluate the \acrshort{ml} models (\autoref{fig:yaib_flow}). 
\vspace{-2mm}
\subsection{Design philosophy}
\vspace{-2mm}
We strongly believe that medical research is inherently complex and that --- rather than providing a rigid benchmark --- there lies most value in providing a modular setup where the user can exchange any part with something that better suits their needs and, importantly, do so reproducibly. For example, users frequently want to highlight a particular aspect of their model, prompting them to adapt the default tasks. Changes, however minor, can render results incomparable. We, therefore, prioritized extensibility across the entire experiment lifecycle. This high level of extensibility may increase the complexity of our benchmark. We mitigate this by providing a range of default experiments for users with limited access to medical expertise or who are content with a fixed set of medical tasks. The experiments were designed to be directly comparable and provide a common benchmark. This allows for a standardized evaluation of models similar to existing benchmarks but still benefits from out-of-the-box support for multiple datasets and easy adaptability if need be. While we did our best to ensure extensibility, \acrshort{yaib} cannot currently support all possible use cases. Specialized use cases like federated learning or reinforcement learning currently require custom code. However, we keep adding functionality to \acrshort{yaib}, and users may nevertheless benefit from using parts of our framework. We provide detailed documentation on how to implement any extensions (\autoref{app:extend}). We strongly request users of YAIB to provide their code and a detailed list of the changes they have made to the repository to accurately and transparently provide results for their experiments. 
\vspace{-2mm}
\subsection{Clinical concepts}
\vspace{-2mm}

We ensured that our benchmark supports existing and future ICU datasets. Working with multiple datasets requires careful data harmonization, as datasets are collected in different locations, with different clinical recording, and may have completely different data structures. We use the \texttt{ricu}~\citep{bennettRicuInterfaceIntensive2023} R package to bring datasets into a common, semantically interoperable format. This harmonization relies on two things: \textbf{1)} a common temporal reference point and \textbf{2)} a dataset-independent definition of clinical concepts. \texttt{ricu} by default distinguishes measurements recorded for a patient, a hospital admission, or an ICU admission, and supports conversion between these levels of measurement. Through definition of reference points, it facilitates temporal comparability between datasets. \texttt{ricu} also allows defining clinical concepts such as heart rate or SOFA score independently of any particular dataset, specifying their meaning, plausible min/max ranges, and units of measurement. A concept can be enabled for a dataset by specifying how it should be extracted from the data, for example, by selecting an entire column or subsetting a table based on an item identifier. \texttt{ricu} thus acts as an interface to the raw data (stored in a fast, compressed column format), on command returning the data for a concept in a table of ID-time-value pairs. This is still no panacea to make ICU datasets immediately interoperable, but it provides a helpful framework for harmonization. For users unfamiliar with R, we provide an interface to access \texttt{ricu} concepts directly from Python. 
\pycu{}, a native Python implementation of \texttt{ricu}, is in development.

\vspace{-2mm}
\subsection{Patient cohort and task definition}
\vspace{-2mm}

Once in a common format, the same task definition can be applied across datasets. This facilitates code reuse and eliminates opportunities for error. Even so, care must be taken to combine clinical concepts, define meaningful prediction targets, and apply appropriate exclusion criteria. We provide default workflows and helper functions to support this process, including a transparent pipeline for applying exclusion criteria and reporting patient attrition. We supplied this functionality in a standalone repository
 to facilitate its use with other modeling frameworks such as Clairvoyance \citep{jarrettCLAIRVOYANCEPIPELINETOOLKIT2021}. The specification of our adaptive and re-definable pipeline is found in \autoref{appendix:outc-detail}. 

\vspace{-2mm}
\subsection{Preprocessing and feature extraction}
\vspace{-2mm}

Further preprocessing is often required at runtime, including data normalization, generation of missingness indicators, and imputation. We provide a transparent, flexible way for users to define their preprocessing pipeline (also available as a standalone package)
, including default implementations of historical aggregation (e.g., mean or variance), resampling of the time resolution, imputation methods, and a wrapper for any Scikit-learn~\citep{scikit-learn} preprocessing step. Custom steps can be added by subtyping an abstracted step interface or providing a callable object to a generic step. 

\vspace{-2mm}
\subsection{Training and evaluation}
\vspace{-2mm}

\label{subsec:training}
A single \acrshort{yaib} experiment creates and optimizes a model for a given task and preprocessing pipeline. Experiments are defined using the \verb|gin-config| library~\citep{google_ginconfig} in simple Python-like text files. 
The model configuration defines the model architecture and contains information on hyperparameters and optimizers. Every aspect of a model is fully configurable. The task configuration defines the target, the data source, the features, and the preprocessing. Additionally, one can define the cross-validation splits and the number of iterations. By defining the model and task separately, they can be mixed and matched, training the same architecture for multiple tasks or training multiple models for a single task. We provide details for adding new datasets, preprocessing, models, and an example of sepsis prediction in \autoref{appendix:yaib_example}.
Training is supervised by PyTorch Lightning~\citep{falcon2019pytorch}, which uses standardized training and logging, GPU parallelism, and advanced debugging. 
Users can configure hyperparameter ranges and sampling methods for model optimization. A Gaussian Process is fit to the hyperparameters using \verb|scikit-optimize|~\citep{headScikitoptimizeScikitoptimize2021} as a robust alternative to random search~\citep{snoekPracticalBayesianOptimization2012}. 

\textbf{Result tracking} 
Results are automatically aggregated and written to a \acrshort{json} file, in addition to optional Tensorboard~\citep{tensorflow2015-whitepaper}, PyTorch Lighting~\citep{falcon2019pytorch}, and \acrshort{wandb}~\citep{wandb} logging for easy experiment tracking. 
Performance evaluation records widely-used metrics out of the box (\acrshort{auroc}, \acrshort{auprc}, calibration curve, accuracy, loss) and supports multiple evaluation libraries: TorchMetrics~\citep{nickiskaftedetlefsenTorchMetricsMeasuringReproducibility2022}, Pytorch-Ignite~\citep{pytorch-ignite}, and Scikit-Learn~\citep{scikit-learn} metrics. 
New metrics, either developed by the user or from existing libraries, can be easily added (see Appendix \ref{app:ext-eval}).
\begin{table}
\small
 \caption{\textit{Prediction task overview.}  Note that the related work is non-exhaustive.} 
 \begin{adjustbox}{center}
 \begin{threeparttable}
 \begin{tabular}{>{\centering\arraybackslash} m{0.01\textwidth}>{\centering\arraybackslash} m{0.07\textwidth}>{\centering\arraybackslash} m{0.14\textwidth}>{\centering\arraybackslash}m{0.01\textwidth}>{\centering\arraybackslash}m{0.63\textwidth}}
 \toprule
 \textbf{No} & \textbf{Task} & \textbf{Frequency} & \textbf{Type}& \textbf{Related work} \\ \hline
 1           & Mortality            & Once per stay*   & C           &   \tiny{\cite{bakerContinuousAutomaticMortality2020, luMachineLearningBased2022, medicEvidencebasedClinicalDecision2019, sharmaMortalityPredictionICU2017, syedApplicationMachineLearning2021}}                   \\ \hline
 2           & \acrshort{aki} & Hourly                &  C & \tiny{\cite{huangInterpretableTemporalConvolutional2021, nikkinenDevelopingSupervisedMachine2022, panSelfCorrectingDeepLearning2019, rankDeeplearningbasedRealtimePrediction2020, shamoutMachineLearningClinical2021, wangRealTimePredictionAKI2020, koynerDevelopmentMachineLearning2018}}                      \\ \hline
 3           & Sepsis                    & Hourly               &  C & \tiny{\cite{kokAutomatedPredictionSepsis2020, lauritsenEarlyDetectionSepsis2020, merathUseMachineLearning2020, fleurenMachineLearningIntensive2020, moorPredictingSepsisMultisite2021, moorEarlyRecognitionSepsis2019, muralitharanMachineLearningBased2021, reynaEarlyPredictionSepsis2019, shamoutMachineLearningClinical2021, wangIntegratingPhysiologicalTime2022}}                      \\ \hline
 4 & \acrshort{kf} & Once per stay* & R &  \tiny{\cite{tomasevClinicallyApplicableApproach2019,futomaPredictingDiseaseProgression2016, perotteRiskPredictionChronic2015,chengPredictingInpatientAcute2018}}\\
 \hline
 5           & \acrshort{los}      & Hourly               & R     &          \tiny{\cite{ shillanUseMachineLearning2019, guoEvaluationTimeSeries2020}}             \\ \bottomrule
 \end{tabular}
 \end{threeparttable}
 \end{adjustbox}
 \label{tab:prediction_task_description}
 \begin{tablenotes}[flushleft]\setlength\labelsep{0pt} 
     \item[]\scriptsize{C: Classification, R: Regression, * Using data from 0-24 hours.}
 \end{tablenotes}
 \vspace{-5mm}
 \end{table}
\vspace{-2mm}
\section{Experiments}
\vspace{-2mm}


We ran experiments for five common prediction tasks: ICU mortality, onset of \acrfull{aki}, onset of sepsis, \acrfull{kf} on day 2, and remaining \acrfull{los} (\autoref{tab:prediction_task_description}). Mortality and KF used data from 0-24 hours. All other task used all available data until the event or discharge. We ensured adequate data quality by excluding: \textbf{1)} patients younger than 18 years; \textbf{2)} stays with missing discharge times; \textbf{3)} stays with less than six hours in the ICU; \textbf{4)} stays with measurements in less than four time bins; and \textbf{5)} stays with no measurement for more than 12 consecutive hours in the ICU. We also applied task-specific exclusion criteria. For example, we excluded stays of less than 30 hours for the ICU mortality task, as this could introduce causal leakage from patients already dead or about to die at the time of prediction. For each task, we included 52 features, of which 4 were static and 48 were time series. Various additional features, including prescriptions and diagnoses, can be directly used in \acrshort{yaib} by adjusting the cohort generation module (YAIB-cohorts); if features are not available, their implementation is straightforward (\autoref{app:extend}). 
Information on the datasets, features, and individual cohort definitions can be found in \autoref{appendix:datasets} and \ref{appendix:outc-detail}. The code to define these cohorts is publicly available. 
In addition to the baseline performance for each task, dataset, and model, we used \acrshort{yaib} to investigate the effects of small variations in task definitions on predictive performance --- a common obstacle to model comparability \citep{moorEarlyPredictionSepsis2021, fleurenMachineLearningIntensive2020}. Specifically, we \textbf{i)} only excluded stays of less than 24 hours to assess the effects of causal leakage by aligning our mortality task with \cite{yecheHiRIDICUBenchmarkComprehensiveMachine2022}, \textbf{ii)} omitted static and dynamic historical features (i.e., \texttt{min}, \texttt{max}, \texttt{count}, \texttt{mean}) to simulate access to fewer input data, and \textbf{iii)} compared alternative definitions for sepsis 
. We, additionally, evaluated transfer learning with the harmonized datasets (\textbf{iv}). 


\textbf{Preprocessing} 
\textbf{1. Scaling:} The data was scaled to zero mean and unit variance.
\textbf{2. Imputation:} After adding missing indicators, we forward-filled all columns for the dynamic data, replacing missing values with the last known values of the same stay. Missing values without a prior measurement were filled with the sample mean. To prevent data leakage, we used the mean of the train split as the sample mean for all splits. 
\textbf{3. Feature generation:} We generated the \verb|min|, \verb|max|, \verb|mean|, and \verb|count| of measurements for each feature in the dynamic data. We only applied this step for the conventional \acrshort{ml} models, e.g., \acrfull{lgbm}, as they cannot capture sequential information natively.

\vspace{-2mm}
\subsection{Models and experimental setup}
\vspace{-2mm}
\label{subsec:ml_algorithms}
We considered a range of algorithms used in previous benchmarks~(\autoref{tab:benchmark_comparison})
 and applied work~\citep{hylandEarlyPredictionCirculatory2020,pirracchioMortalityPredictionIntensive2015a,silvaPredictingInHospitalMortality2012,syedApplicationMachineLearning2021}, including regularized \acrfull{lr} and \acrfull{en} (used for classification and regression, respectively~\citep{scikit-learn}), \acrshort{lgbm} \citep{keLightGBMHighlyEfficient2017}, and four variations of neural networks: \acrfull{gru}~\citep{choPropertiesNeuralMachine2014}, \acrfull{lstm}~\citep{hochreiterLongShortTermMemory1997}, \acrfull{tcn}~\citep{baiEmpiricalEvaluationGeneric2018a} and \acrfull{tf}~\citep{vaswaniAttentionAllYou2017}. 
\acrshort{lr}, \acrshort{en}, and \acrshort{lgbm} were used with the feature generation described above, as they are unable to utilize time series.
The implementation of neural networks was adapted from \cite{yecheHiRIDICUBenchmarkComprehensiveMachine2022}. 

\label{subsec:exp_settings}
For our experiments, unless stated otherwise, we used 5 iterations of 5-fold cross-validation. 
Hyperparameters were tuned on the training set using 30/50 (DL/ML, respectively) iterations of Bayesian hyperparameter optimization \citep{snoekPracticalBayesianOptimization2012}. For computational reasons, hyperparameter tuning used only the first 2/3 folds, respectively (see \autoref{app:hyperparams} for a definition of all searched and selected hyperparameters). The final validation of the best hyperparameters used all 5 folds. 
Each model was optimized for a maximum of 1000 epochs.
Training was stopped early if performance on the validation set did not improve for 10 epochs. The epoch with the best performance on the validation set was retained and evaluated on the test set. 
This process was repeated for 5 iterations, after which the results were averaged, and the standard deviation was calculated. 

\renewcommand{\arraystretch}{0.95}
\setlength{\tabcolsep}{4.5pt}
\begin{table}[t]
    
        \caption{\textit{Baseline performance on the classification tasks.} We \textbf{embolden} the best mean AUROC $ \times \ 100 $ ($\uparrow$, i.e., higher is better) and {\color{darkgray!60}AUPRC} $ \times \ 100 $ ($\uparrow$) per dataset and those within a standard deviation (±).}
        \centering
    \small
        \begin{tabularx}{\textwidth}{l>{\centering\arraybackslash}X>{\centering\arraybackslash\leavevmode\color{darkgray!60}}X>{\centering\arraybackslash}X>{\centering\arraybackslash\leavevmode\color{darkgray!60}}X>{\centering\arraybackslash}X>{\centering\arraybackslash\leavevmode\color{darkgray!60}}X>{\centering\arraybackslash}X>{\centering\arraybackslash\leavevmode\color{darkgray!60}}X}
            \toprule
            \addlinespace[0.2em]
             & \multicolumn{2}{c}{\textbf{\acrshort{aumc}}} & \multicolumn{2}{c}{\textbf{\acrshort{hirid}}} & \multicolumn{2}{c}{\textbf{\acrshort{eicu}}} & \multicolumn{2}{c}{\textbf{\acrshort{miiv}}}\\
            \addlinespace[0.1em]
             \cline{2-9}
            \addlinespace[0.2em]
                \textbf{Algorithm} & AUROC & AUPRC & AUROC & AUPRC & AUROC & AUPRC & AUROC & AUPRC \\
            \cmidrule{1-9}\morecmidrules\cmidrule{1-9}            
            \textbf{Mortality}\\
            LR & 83.7±0.6 & 52.9±1.2 &  84.0±0.3 & 36.9±1.1 & 84.8±0.2 & 33.0±0.7 & 86.1±0.1 & 39.7±0.6 \\
            LGBM & \textbf{84.5±0.5} & \textbf{53.7±1.2} & 84.4±0.3& \textbf{40.6±0.8} & \textbf{85.7±0.2} & \textbf{36.0±0.6} & \textbf{87.7±0.2} & \textbf{44.2±0.7} \\
                \addlinespace[0.3em]
            GRU & 83.9±0.3 & \textbf{53.8±0.7} & \textbf{84.8±0.2} & 39.4±0.4 & \textbf{86.0±0.1} & \textbf{35.6±0.1} & \textbf{87.6±0.1} & 42.8±0.3 \\
            LSTM & 83.7±0.7 & \textbf{53.6±1.4} & 84.0±0.7 & 37.8±1.0 & 85.5±0.2 & \textbf{35.7±0.8} & 86.7±0.4 & 41.0±0.7 \\
            TCN & \textbf{84.0±0.6} & \textbf{54.2±1.4} & \textbf{84.6±0.7} & 39.2±1.3 & 85.4±0.2 & 34.3±0.6 & 87.1±0.3 & 41.4±0.8 \\
            TF & 84.1±0.2 & \textbf{54.4±1.1} & \textbf{84.9±0.7} & 39.3±1.5 & \textbf{85.9±0.2} & 34.7±0.8 & 86.9±0.3 & 42.2±0.3 \\
            \cmidrule{1-9}\morecmidrules\cmidrule{1-9} 
            \textbf{AKI}\\
            LR & 85.5±0.3 & 45.1±0.4 & 79.6±0.1 & 31.8±0.8 & 72.8±0.1 & 32.2±0.2 & 77.1±0.2 & 37.7±0.3 \\
            LGBM & 85.8±0.3  & 48.4±0.6 & 80.2±0.2 & 32.8±0.4 & 84.6±0.1 & 50.8±0.2 & 83.8±0.1 & 53.3±0.2 \\
                \addlinespace[0.3em]
            GRU & \textbf{90.6±0.3} & \textbf{52.8±0.7} & \textbf{82.2±0.2} & \textbf{33.9±0.4} & \textbf{90.9±0.0} & \textbf{72.2±0.1} & \textbf{90.7±0.1} & \textbf{69.6±0.2} \\
            LSTM & 86.5±0.4 & 40.6±0.6 & 81.0±0.4 & 31.8±0.4 & 90.2±0.1 & 69.9±0.2 & 89.7±0.1 & 66.5±0.2 \\
            TCN & 89.6±0.2 & 50.0±0.9& 81.2±0.2 & 32.3±0.4 & 90.4±0.0 & 70.4±0.2 & 89.8±0.1 & 66.8±0.2 \\
            TF & 88.2±0.2 & 48.2±0.7 & 81.5±0.2 & 33.4±0.5 & 89.9±0.1 & 68.0±0.3 & 89.6±0.1 & 65.6±0.2 \\
            \cmidrule{1-9}\morecmidrules\cmidrule{1-9} 
            \textbf{Sepsis}\\
            LR & 74.7±1.0 & 4.0±0.4 & 76.5±0.6 & 8.4±0.3& 71.8±0.3 & 2.9±0.1 & 77.1±0.4 & 4.6±0.1 \\
            LGBM & 74.0±0.8 & 5.2±0.7 & 76.1±0.4 & 10.4±0.5 & 69.1±0.3 & 3.3±0.1& 77.5±0.3 &5.9±0.2 \\
                \addlinespace[0.3em]
            GRU & 79.7±0.9 & 7.7±0.7 & \textbf{80.6±0.5} & \textbf{12.6±0.5} &\textbf{77.4±0.2} & \textbf{5.1±0.1} & \textbf{83.6±0.3} & \textbf{9.1±0.3} \\
            LSTM & 77.1±0.8 & 6.4±0.5 & 78.8±0.4 & 11.1±0.5 & 74.0±0.2 & 4.0±0.1 & 82.0±0.3 & 8.0±0.2 \\
            TCN & 78.7±0.7 & 7.1±0.6 & \textbf{80.8±0.5} & \textbf{13.0±0.4} & 76.7±0.1& 4.9±0.1 & 82.7±0.3 & \textbf{8.8±0.2} \\
            TF & \textbf{80.7±0.9} & \textbf{8.6±0.8} & \textbf{80.8±0.3} & \textbf{12.6±0.6} & 76.2±0.1 & 4.6±0.1 & 80.0±0.8 & 6.6±0.2 \\
            \bottomrule
        \end{tabularx}
        \label{tab:baselines}
        \vspace{-2.5mm}
\end{table}

\vspace{-2mm}
\subsection{Benchmarking baseline models on major ICU datasets}
\vspace{-2mm}

Baseline results for all tasks can be found in \autoref{tab:baselines} and \ref{tab:reg_baselines}. Note that we have also benchmarked our tasks for two openly available demo datasets from \acrshort{miiii} and \acrshort{eicu}; these can be directly accessed without completing a credentialing procedure (see \autoref{tab:demo_baselines} and \ref{tab:reg_baselines_demo}).
\begin{table}
\setlength{\aboverulesep}{0pt}
\setlength\tabcolsep{0.3pt}
    \center
     \caption{\textit{Baseline performance on the regression tasks.} Results are reported in \acrlong{mae} ($\downarrow$)}
        \small
        \begin{tabularx}{\textwidth}{lX>{\centering\arraybackslash}X>{\centering\arraybackslash}X>{\centering\arraybackslash}X>{\hsize=.20\hsize}X>{\hsize=.97\hsize\centering\arraybackslash\leavevmode}X>{\hsize=.97\hsize\centering\arraybackslash\leavevmode}X>{\hsize=.97\hsize\centering\arraybackslash\leavevmode}X>{\hsize=.97\hsize\centering\arraybackslash\leavevmode}X}
            \toprule \addlinespace[0.2em]  
            \textbf{} & \multicolumn{4}{c}{\textbf{\Acl{kf}} \textit{in mg/dL}} & & \multicolumn{4}{c}{\textbf{Length of Stay} \textit{in hours}}\\
            \cmidrule{2-5}\cmidrule{7-10}
            \textbf{Algo.} & 
             \textbf{\acrshort{aumc}} & \textbf{\acrshort{hirid}} & \textbf{\acrshort{eicu}} & \textbf{\acrshort{miiv}}   & &  \textbf{\acrshort{aumc}} & \textbf{\acrshort{hirid}} & \textbf{\acrshort{eicu}} & \textbf{\acrshort{miiv}}\\
            \cmidrule{1-5}\cmidrule{7-10}\morecmidrules\cmidrule{1-10}
            EN & \textbf{0.24±0.00} & 0.28±0.00 & 0.31±0.00 & 0.25±0.00 & & 54.9±0.0 & 47.2±0.1 & 43.6±0.0 & 46.5±0.0\\
            LGBM & 0.32±0.00 & 0.34±0.00 & \textbf{0.29±0.00} & \textbf{0.24±0.00} & & 44.7±0.0 & \textbf{39.2±0.1} & 39.3±0.0 & 40.1±0.0\\
            \addlinespace[0.3em]
            GRU & 0.29±0.00 & 0.32±0.01 & 0.34±0.01 & 0.30±0.01 & & 42.9±0.1 & 39.6±0.1 & 38.9±0.1 & 39.9±0.1\\
            LSTM  & 0.29±0.00 & 0.33±0.00 & \textbf{0.28±0.01} & 0.28±0.01 & & 44.8±0.1 &  39.8±0.1 & 39.2±0.1 &  40.6±0.1\\
            TCN  & 0.28±0.01 & \textbf{0.23±0.01} & 0.31±0.00 & 0.28±0.01 & &43.7±0.1 & 39.9±0.1 & 38.9±0.0 & 40.4±0.1\\
            TF & 0.26±0.00 & 0.31±0.01 & 0.33±0.01 & 0.32±0.01 & & \textbf{41.8±0.1} & \textbf{39.1±0.1} & \textbf{38.2±0.1} &  \textbf{39.0±0.1} \\
            \bottomrule
        \end{tabularx}
        \begin{tablenotes}[flushleft]\setlength\labelsep{0pt} 
        \scriptsize
        \item[] We provide the average and Interquartile range for Kidney Function and Length of Stay in \autoref{tab:characteristics}.
        \end{tablenotes}
    \label{tab:reg_baselines}
\end{table}

\textbf{ICU mortality}
The performance of traditional \acrshort{ml} and \acrshort{dl} models was highly comparable among each other and across datasets when predicting mortality based on data from the first 24 hours. Notably, AUPRC was higher in \acrshort{aumc} due to a higher outcome prevalence (\autoref{tab:dataset_comparison_extended}). 

\textbf{\Acf{aki}}
Maximum achievable performance was also similar across datasets when predicting the hourly onset of \acrshort{aki}, with the notable exception of \acrshort{hirid}, which had both lower \acrshort{auroc} and \acrshort{auprc} for all models. GRU models consistently achieved the best performance.

\textbf{Sepsis}
The performance of baseline models was worst for the hourly onset of sepsis, both for \acrshort{auroc} and especially \acrshort{auprc}. This may be explained by the particularly low prevalence of $\sim1$\% hourly bins classified as septic and the relative difficulty of predicting sepsis in general~\citep{moorEarlyPredictionSepsis2021}.

\textbf{\Acf{kf}} 
Classical \acrshort{ml} models achieved relatively good performance for this task, which may reflect the dependence of KF on a limited number of features~\citep{grinsztajnWhyTreebasedModels2022}. 

\textbf{Remaining \acf{los}}
The performance of \acrshort{ml} and \acrshort{dl} models was also comparable across datasets. Nevertheless, predicting the length of stay seems difficult, given that the average MAE is almost two days.
Transformers consistently outperformed most other model types.



\begin{table}[t]
\center
\caption{\textit{ICU mortality prediction on \acrshort{hirid} with (>24h) and without (>30h) possibility of causal leakage.}
}
\small
\begin{tabular}{lc>{\color{darkgray!60}}cc>{\color{darkgray!60}}cc>{\color{darkgray!60}}c}
\toprule
\addlinespace[0.2em]
& \multicolumn{6}{c}{\textbf{Cohort definition}}\\
\cline{2-7}
\addlinespace[0.2em]
& \multicolumn{2}{c}{\textbf{w/o leakage}} & \multicolumn{2}{c}{\textbf{w/ leakage}} & \multicolumn{2}{c}{\textbf{\cite{yecheHiRIDICUBenchmarkComprehensiveMachine2022}}} \\
\hline
\addlinespace[0.2em]
 \multicolumn{1}{l}{\textbf{Algorithm}} & AUROC         & AUPRC        & AUROC         & AUPRC        & AUROC            & AUPRC         
 \\     \cmidrule{1-7}\morecmidrules\cmidrule{1-7} 
\addlinespace[0.3em]
\multicolumn{1}{l}{LR}& 84.0±0.3 &  36.9±1.1 & 87.2±0.4 & 43.1±1.3 & 89.0±0.0 & 58.1±0.0\\
\multicolumn{1}{l}{LGBM} & \textbf{84.5±0.3} & 40.6±0.9 & \textbf{87.9±0.5} & \textbf{47.7±1.2} & 88.8±0.2 & 54.6±0.8 \\
 \addlinespace[0.5em]
\multicolumn{1}{l}{GRU} & \textbf{84.8±0.2} & 39.4±0.4 & \textbf{88.2±0.3} & 46.1±1.2 & 90.0±0.4 & \textbf{60.3±1.6}\\ 
\multicolumn{1}{l}{TCN} & \textbf{84.6±0.7} & 39.2±1.3 & 87.8±0.2 & 45.2±1.0 & 89.7±0.4 & \textbf{60.2±1.1}\\ 
\multicolumn{1}{l}{TF} & \textbf{84.9±0.7} & 39.4±1.5 & \textbf{88.2±0.3} & \textbf{47.1±1.2} & \textbf{90.8±0.2} & \textbf{61.0±0.8}\\ 
\bottomrule 
\end{tabular}
\label{tab:extended_cohort}
\end{table}
\vspace{-2mm}
\subsection{Using YAIB as an experimental ML framework}
\vspace{-2mm}
\textbf{Changing exclusion criteria for mortality cohorts} As hypothesized, the choice of exclusion criteria could majorly impact achievable prediction performance (\autoref{tab:extended_cohort}).
Compared to the peak performance achieved with the \acrshort{hirid}-benchmark~\citep{yecheHiRIDICUBenchmarkComprehensiveMachine2022}, our baseline performance for the mortality task was noticeably lower. 
Aligning the exclusion criteria accounted for half of the performance difference. The remaining difference was likely due to the inclusion of additional predictors --- most notably drug usage --- in the \acrshort{hirid}-benchmark. 
This highlights the difficulties of comparing works that ostensibly address the same task, even using the same dataset and model implementation.
\textbf{Restricting input features} 
We observed that dynamic feature generation consistently outperformed task definitions that did not include them (\autoref{tab:without_dynamic} and \ref{tab:without_dynamic_auprc}). 
 LR on \acrshort{miiv} showed a considerable performance gap, whereas \acrshort{aumc} remained stable. We noted a performance decrease that ranges between 4.0\% and 19.1\%  for LR and between 5.2\% and 13.1\% for LGBM.
 Omitting static features led to minor drops in performance (\autoref{tab:without_static_auroc} and \ref{tab:without_static_auprc}); averaged across datasets, we observe a performance differences ranging between 0.5\% and 0.2\% for the transformer model. 

\textbf{Comparing sepsis definitions} 
Label definitions also had a considerable impact on \acrshort{auroc} and/or \acrshort{auprc} (\autoref{tab:sepsis_comparison_results}), which was not always apparent from the definition alone. 
Sepsis has been defined in several ways \citep{fleurenMachineLearningIntensive2020}, mainly because a clinical gold standard that can be transferred to \acrshort{ml} models is currently lacking. 
Our sepsis definition (adapted from \cite{seymourAssessmentClinicalCriteria2016}, see \autoref{appendix:outc-detail}) can be considered closely related to that used by \cite{moorPredictingSepsisMultisite2021}, who implement a variant of Sepsis-3~\citep{singerThirdInternationalConsensus2016}. However, we required that antibiotics were administered continuously for $\geq 3$ days~\citep{reynaEarlyPredictionSepsis2019}. We judged that this would increase the clinical usability of the task but found that it also severely reduced the achievable \acrshort{auprc} --- likely due to a much lower prevalence (\autoref{tab:sepsis_definition}). The definition used by \cite{calvertComputationalApproachEarly2016} on the other hand adapted Sepsis-2~\citep{levy2001SCCMESICM2003}, which differs fundamentally from Sepsis-3 and resulted in a notably higher \acrshort{auroc}~\citep{engorenComparisonSepsis2Systemic2020}. This highlights the importance of precise cohort definitions, as some definitions may, by design, be more difficult to predict.


\begin{table}
\center
\caption{\textit{Sepsis prediction on \acrshort{miiv} for different definitions of sepsis.}}
\begin{threeparttable}
\small
\begin{tabular}{lc>{\color{darkgray!60}}cc>{\color{darkgray!60}}cc>{\color{darkgray!60}}c}
\toprule
& \multicolumn{6}{c}{\textbf{Sepsis definition}}\\
\addlinespace[0.2em]
\cline{2-7}
\addlinespace[0.2em]
& \multicolumn{2}{c}{\textbf{\cite{seymourAssessmentClinicalCriteria2016}}*} & \multicolumn{2}{c}{\textbf{\cite{moorPredictingSepsisMultisite2021}}} & \multicolumn{2}{c}{\textbf{\cite{calvertComputationalApproachEarly2016}}} \\
\hline
\addlinespace[0.2em]
 \multicolumn{1}{l}{\textbf{Algorithm}} & AUROC         & AUPRC        & AUROC         & AUPRC        & AUROC            & AUPRC        
 \\ 
\cmidrule{1-7}\morecmidrules\cmidrule{1-7}
\addlinespace[0.3em]
\multicolumn{1}{l}{LGBM} & 75.9±0.2 & 4.3±0.0 & 72.4±0.0 & 10.5±0.0 & 62.2±0.2 & 1.8±0.0 \\
\multicolumn{1}{l}{GRU} & \textbf{79.2±0.1} & \textbf{6.1±0.0} & \textbf{80.9±0.0} & \textbf{17.7±0.0} & \textbf{89.2±0.0} & \textbf{9.3±0.2}\\ 
\bottomrule 
\end{tabular}

\begin{tablenotes}[flushleft, para]\setlength\labelsep{0pt}  
\scriptsize
    \item * Our definition; adapted to be more clinically actionable, see \autoref{appendix:outc-detail}.
\end{tablenotes}
\vspace{-5mm}
\end{threeparttable}
\label{tab:sepsis_comparison_results}
\end{table}
\begin{figure}[b]
\vspace{-3mm}
\centering
\begin{subfigure}{0.46\textwidth}
  \label{fig:ext_val_auroc_mortality}
  \centering
        \includegraphics[width=\textwidth]{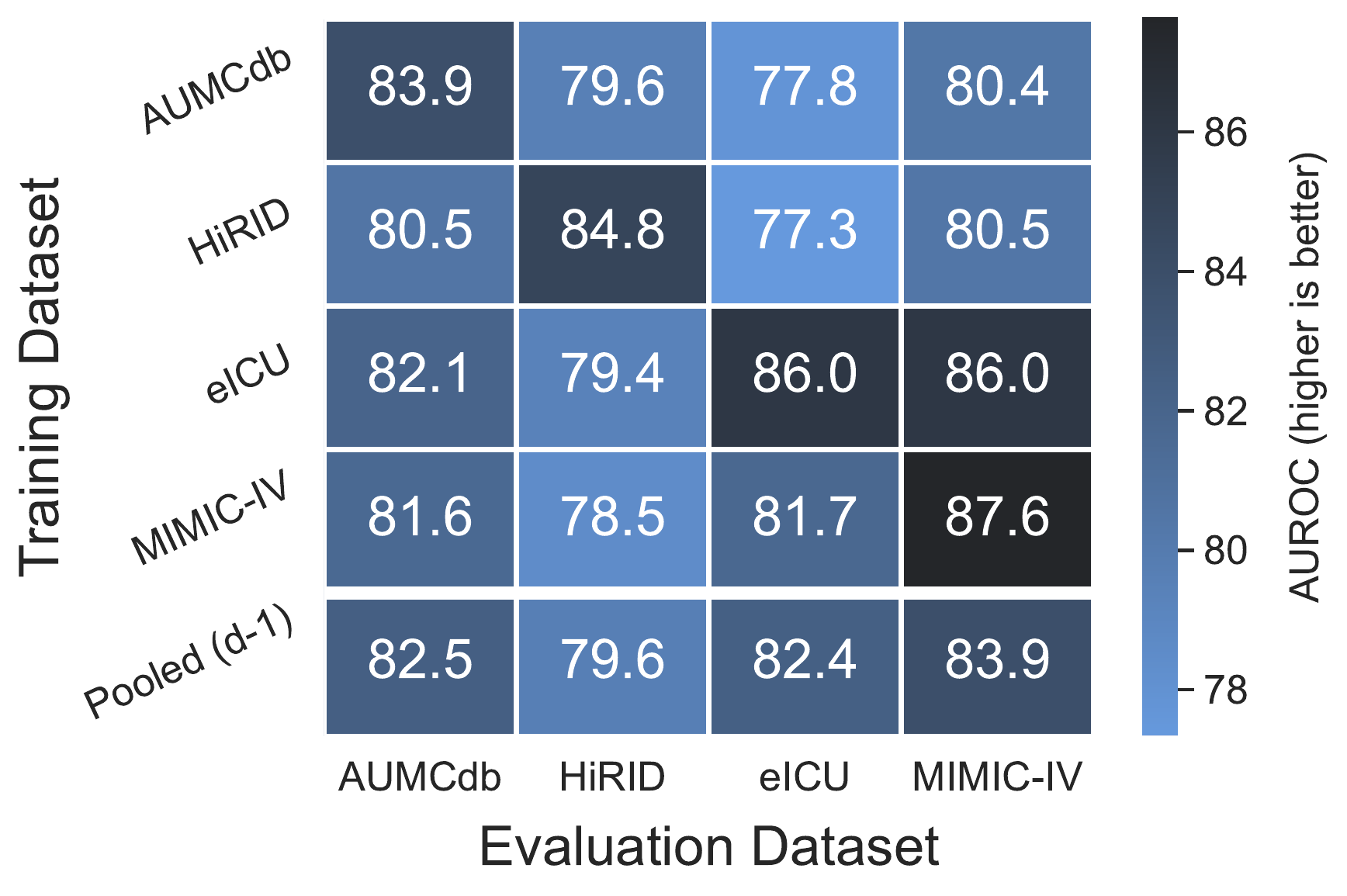}
\end{subfigure}\hfill
\begin{subfigure}{0.46\textwidth}
  \centering
  \label{fig:ext_val_mae_los}
        \includegraphics[width=\textwidth]{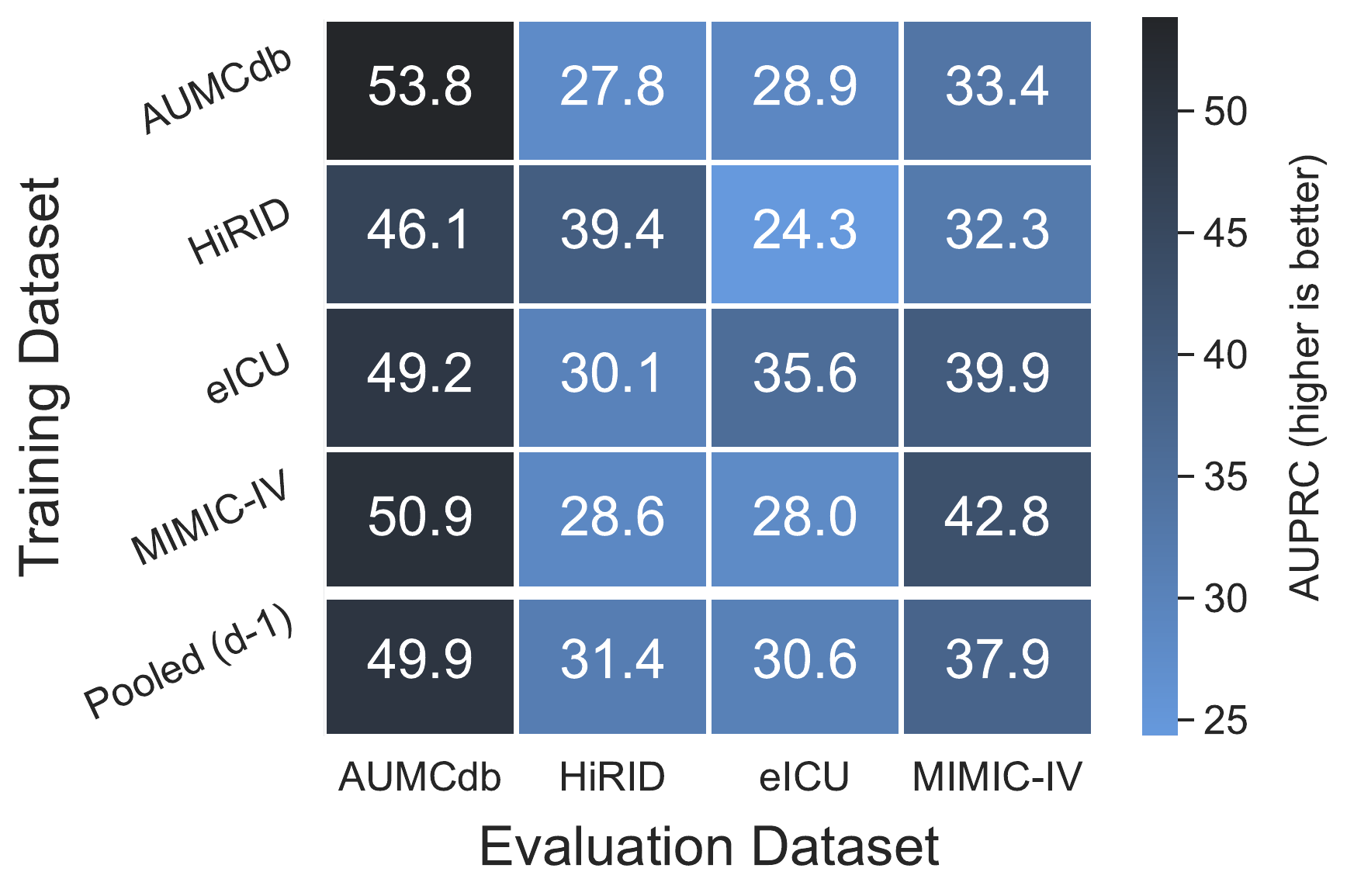}
\end{subfigure}
\caption{\textit{Performance of prediction models when trained on one dataset (row) and evaluated on all others (columns).} \textbf{Left}: Performance in AUROC of the GRU model on \textit{ICU mortality}. \textbf{Right}: Performance in AUPRC for the same models. Pooled (d-1) refers to training a model on every dataset except the evaluation dataset.}
\label{fig:ext_validation}
\end{figure}
\vspace{-3mm}
\subsection{Transfer learning}
\vspace{-2mm}

\textbf{External validation}
\acrshort{yaib}'s common dataset format allowed us to evaluate a model trained on an equal sample of one dataset on data from all other datasets. 
We additionally trained a model on pooled (d-1) data from three datasets and evaluated on the fourth, held-out dataset. 
For the ICU mortality task (\autoref{fig:ext_validation}), models, as expected, performed best on independent test data from their training dataset (diagonal). Performance could drop considerably when models were evaluated in another database (off-diagonal). Notably, AUPRC performance could increase in the evaluation dataset (rows) but always remained lower than the highest achievable performance for that dataset (columns). We found that \acrshort{miiv} and \acrshort{eicu} transferred well among each other. The pooled model usually performed as well as the best single-dataset model. 
Notably, \acrshort{aumc} AUPRC results demonstrate decidedly higher performance than evaluation on other datasets, which could be the result of a patient case mix and outcome prevalence (see \autoref{tab:characteristics}).

\begin{wrapfigure}{R}{0.40\textwidth}
\vspace{-0.6cm}
    \includegraphics[width=0.35\textwidth]{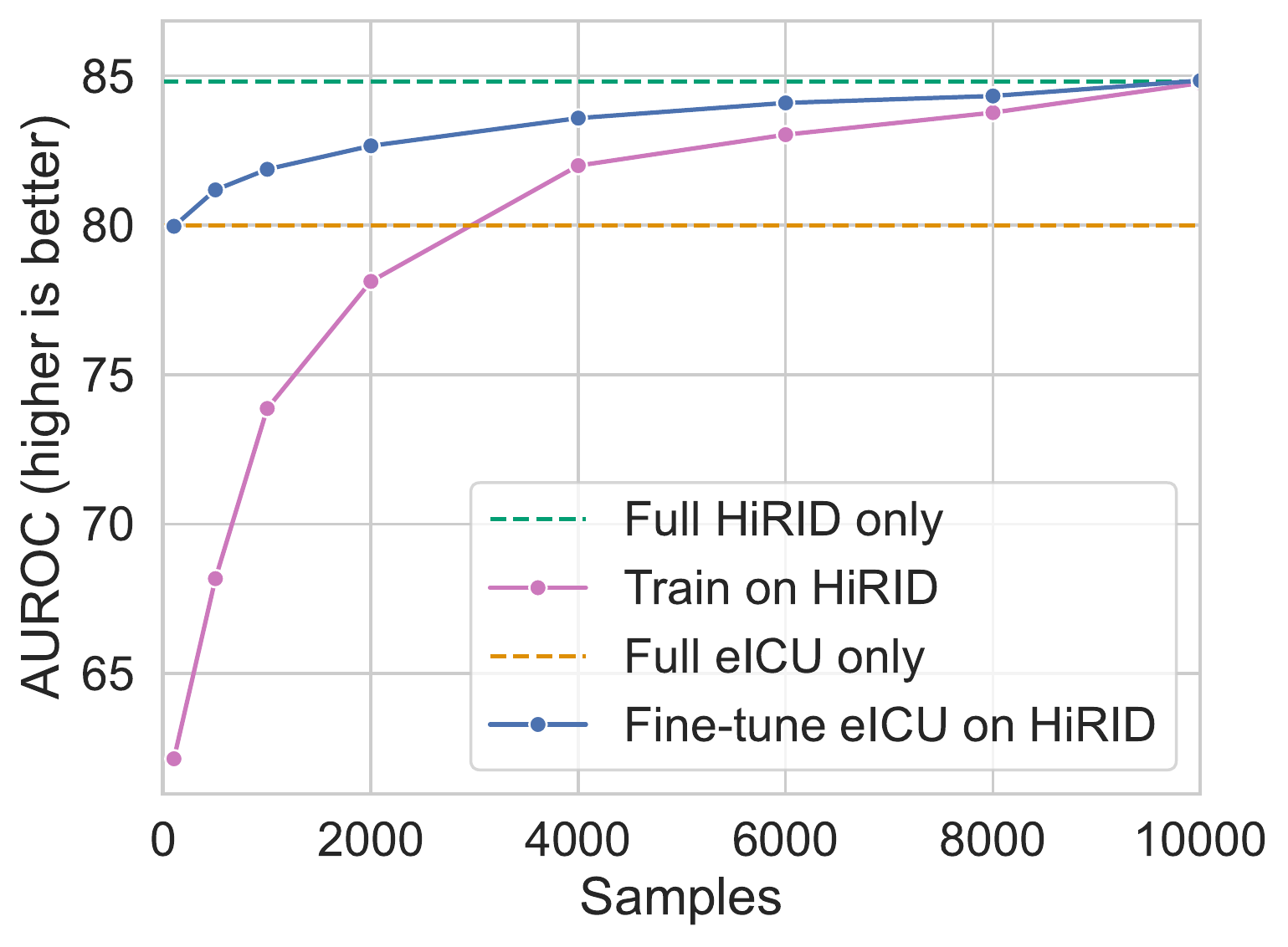}
    \caption{\textit{Fine-tuning an eICU model for ICU mortality prediction on HiRID.}}
    \label{fig:fine-tuning}
    \vspace{-0.7cm} 
\end{wrapfigure}
\textbf{Fine-tuning} In \autoref{fig:ext_validation}, we saw that \acrshort{eicu} resulted in the most generalizable model for ICU mortality, which may serve as a strong pre-training for transfer learning. Since it worked worst for \acrshort{hirid}, we further fine-tuned the \acrshort{eicu} GRU model (source) for \acrshort{hirid} (target) by retraining it using an increasing number of samples from the \acrshort{hirid} dataset. We compared the results to a model trained from scratch on the same amount of \acrshort{hirid} samples (\autoref{fig:fine-tuning}). Fine-tuning was profitable for any number of additional samples and especially for <4,000 samples.
\vspace{-3.5mm}
\section{Discussion}
\vspace{-2mm}

\label{sec:discussion}
We provide extensive \acrshort{ml} and \acrshort{dl} baselines for five clinical prediction tasks trained across four major open-source ICU datasets.
While we frequently obtained comparable results across model architectures, seemingly small differences in cohort definition could substantially impact the achieved accuracy. 
Our findings highlight not only the need for standardized training pipelines but also for harmonized cohort definitions to allow for a meaningful comparison of clinical prediction models. Our work provides the first international, multi-center ICU benchmark, including the first-ever benchmark for the \acrlong{aumc} dataset.
It naturally facilitates sorely needed external validation of model performances and allows fine-tuning of pre-trained models for new datasets.
This makes \acrshort{yaib} relevant to a wide range of research areas beyond classical supervised learning, including domain adaption and generalization.
We hope this broad reach encourages ICU data providers to ensure compatibility with \acrshort{yaib}, as they can expect a larger overall research impact.
This simplifies the use of novel datasets by the clinical and \acrshort{ml} community. 

\acrshort{yaib} aids researchers in training baseline models by providing them with ready-to-use implementations of state-of-the-art model architectures; new model implementations can therefore be easily compared.
While most existing benchmarking studies are hard-coded, we utilize flexible, \textit{dataset-independent} cohort definitions and configurable preprocessing facilities linked via a common, shareable syntax. 
This setup acknowledges that task definitions inevitably involve arbitrary decisions, without one “size” that fits all. 
In our work, we embrace this idea and aim to equip researchers — both applied and theoretical — with the tools to quickly adapt a task to their individual needs (including the use of custom proprietary data) while maintaining reproducibility and reusability across studies. 
Models can thus be compared across multiple, slightly different task definitions and datasets, still ensuring an apples-to-apples comparison. 
We hope this lowers the bar for researchers to test their approaches across a range of configurations and datasets.

\acrshort{yaib} is currently limited to ICU settings, where several datasets are publicly available. A similar setup could be beneficial for data from other medical settings, such as inpatient wards. 
Although created for critical care, \acrshort{yaib} is not specific to the ICU and can be readily extended to other settings, provided a suitable configuration is defined. Features included in \acrshort{yaib}, at the time of writing, mainly relate to vital signs, lab tests, and data relevant to outcome definitions. Further clinician-assisted harmonization efforts will be necessary to increase the breadth of features, most notably medications and comorbidities. If \acrshort{yaib} is adapted to general EHR, including clinical notes and medical imaging is a logical next step. 
We also note that we compared these models on the basis of commonly used \acrshort{ml} metrics; we leave the comparison with respect to clinical fairness and bias as an easy future extension to our framework (see \autoref{app:extend}).
Finally, we advise users of our benchmark to carefully consider the compromises made to allow for cohort harmonization; we strongly recommend clinical validation before making practical decisions based on the developed models. 

\vspace{-3mm}
\section{Conclusion}
\vspace{-2mm}
Routine medical data is highly complex. Without clear ground truth, researchers are inevitably forced to make arbitrary design choices when defining outcomes and populations of interest. To promote comparable and reproducible models in this setting, we believe that further tools are needed that allow researchers to define clinical prediction tasks transparently, share experimental setups easily, and validate results against various data sources. As a flexible and extensible framework for clinical modeling on ICU data, \acrshort{yaib} is meant to be a step towards that goal.
\newpage
\section{Acknowledgements}
Robin van de Water is funded by the “Gemeinsamer Bundesausschuss (G-BA)
Innovationsausschuss” in the framework of “CASSANDRA - Clinical ASSist AND aleRt Algorithms”
(project number 01VSF20015). We would like to acknowledge the work of Alisher Turubayev, Anna Shopova, Fabian Lange, Mahmut Kamalak, Paul Mattes, and Victoria Ayvasky for adding Pytorch Lightning, Weights and Biases compatibility, and several optional imputation methods to a later version of the benchmark repository. 
\section{Ethics statement}
 We do not manage access and do not provide access to any of the full medical datasets included in this work, and we adhere to the usage licenses for each dataset. Users can follow the credentialing procedures outlined in \autoref{appendix:datasets}. However, we provide two preprocessed demo datasets out of the box for reproducibility and experimentation. 
 The demo task cohorts for MIMIC-III and eICU mentioned in that section are derived from the official demo datasets published on PhysioNet by the original authors of the respective databases. Each demo dataset represents a small, curated subset of data that is freely accessible without any need for human subject training. 
 Both demo datasets are published under an Open Data Commons Open Database License v1.0, which explicitly permits the adoption and sharing of the data. The original demo data, as well as further information, can be found at the \href{https://physionet.org/content/mimiciii-demo/1.4/}{MIMIC-III demo} and \href{https://physionet.org/content/eicu-crd-demo/2.0.1/}{eICU demo} Physionet pages.

\section{Reproducibility statement}
We include the source code of \texttt{YAIB}\footnote{\url{https://github.com/rvandewater/YAIB}} (main benchmark), \texttt{YAIB-cohorts}\footnote{\url{https://github.com/rvandewater/YAIB-cohorts}} (adaptable cohort extraction) and \texttt{ReciPys}\footnote{\url{https://github.com/rvandewater/ReciPys}} (extensible preprocessing package) in our submission. Models for each task and architecture are publicly available\footnote{\url{https://github.com/rvandewater/YAIB-models}}.
In the included source code, a file called \texttt{PAPER.md}\footnote{\url{https://github.com/rvandewater/YAIB/blob/master/PAPER.md}} describes the reproducibility steps of the experiments in this paper. Specifically, one requires the standalone codebase of \texttt{YAIB-cohorts} to first create the cohorts from the acquired data, once you have completed the required credentialing (see \autoref{appendix:datasets} for details). As mentioned, we include demo cohort data for each task (results for these cohorts are shown in \autoref{app:ext-results}).   \autoref{appendix:outc-detail} describes the data processing and task creation. The usage of \acrshort{yaib} is detailed in \autoref{app:usage}. \autoref{app:extend} shows how \acrshort{yaib} can be extended with new datasets, clinical concepts, tasks, models, and evaluation metrics.  Additionally, we refer to the \texttt{README.md}\footnote{\url{https://github.com/rvandewater/YAIB/blob/master/README.md}} and the wiki\footnote{\url{https://github.com/rvandewater/YAIB/wiki/YAIB-wiki-home}} for the usage of \acrshort{yaib}. \autoref{app:exp_setup} and \ref{app:hyperparams} detail the experiment design and chosen hyperparameters, respectively. Finally, \autoref{app:checklist} contains the machine learning reproducibility checklist for our work.

\newpage
\bibliography{yaib_bibliography}

\begin{thebibliography}{92}
\providecommand{\natexlab}[1]{#1}
\providecommand{\url}[1]{\texttt{#1}}
\expandafter\ifx\csname urlstyle\endcsname\relax
  \providecommand{\doi}[1]{doi: #1}\else
  \providecommand{\doi}{doi: \begingroup \urlstyle{rm}\Url}\fi

\bibitem[Abadi et~al.(2016)Abadi, Barham, Chen, Chen, Davis, Dean, Devin,
  Ghemawat, Irving, Isard, Kudlur, Levenberg, Monga, Moore, Murray, Steiner,
  Tucker, Vasudevan, Warden, Wicke, Yu, and Zheng]{tensorflow2015-whitepaper}
Martin Abadi, Paul Barham, Jianmin Chen, Zhifeng Chen, Andy Davis, Jeffrey
  Dean, Matthieu Devin, Sanjay Ghemawat, Geoffrey Irving, Michael Isard,
  Manjunath Kudlur, Josh Levenberg, Rajat Monga, Sherry Moore, Derek~G. Murray,
  Benoit Steiner, Paul Tucker, Vijay Vasudevan, Pete Warden, Martin Wicke, Yuan
  Yu, and Xiaoqiang Zheng.
\newblock {{TensorFlow}}: {{A}} system for large-scale machine learning.
\newblock In \emph{12th {{USENIX}} Symposium on Operating Systems Design and
  Implementation ({{OSDI}} 16)}, pp.\  265--283, 2016.

\bibitem[Bai et~al.(2018)Bai, Kolter, and
  Koltun]{baiEmpiricalEvaluationGeneric2018a}
Shaojie Bai, J.~Zico Kolter, and Vladlen Koltun.
\newblock An {{Empirical Evaluation}} of {{Generic Convolutional}} and
  {{Recurrent Networks}} for {{Sequence Modeling}}, April 2018.

\bibitem[Baker et~al.(2020)Baker, Xiang, and
  Atkinson]{bakerContinuousAutomaticMortality2020}
Stephanie Baker, Wei Xiang, and Ian Atkinson.
\newblock Continuous and automatic mortality risk prediction using vital signs
  in the intensive care unit: A hybrid neural network approach.
\newblock \emph{Scientific Reports}, 10\penalty0 (1):\penalty0 21282, December
  2020.
\newblock ISSN 2045-2322.
\newblock \doi{10.1038/s41598-020-78184-7}.

\bibitem[Barbieri et~al.(2020)Barbieri, Kemp, {Perez-Concha}, Kotwal,
  Gallagher, Ritchie, and Jorm]{barbieriBenchmarkingDeepLearning2020}
Sebastiano Barbieri, James Kemp, Oscar {Perez-Concha}, Sradha Kotwal, Martin
  Gallagher, Angus Ritchie, and Louisa Jorm.
\newblock Benchmarking {{Deep Learning Architectures}} for {{Predicting
  Readmission}} to the {{ICU}} and {{Describing Patients-at-Risk}}.
\newblock \emph{Scientific Reports}, 10\penalty0 (1):\penalty0 1111, January
  2020.
\newblock ISSN 2045-2322.
\newblock \doi{10.1038/s41598-020-58053-z}.

\bibitem[Bennett et~al.(2023)Bennett, Plecko, Ukor, Meinshausen, and
  B{\"u}hlmann]{bennettRicuInterfaceIntensive2023}
Nicolas Bennett, Drago Plecko, Ida-Fong Ukor, Nicolai Meinshausen, and Peter
  B{\"u}hlmann.
\newblock Ricu: {{R}}'s interface to intensive care data.
\newblock \emph{GigaScience}, 12, June 2023.
\newblock \doi{10.1093/gigascience/giad041}.

\bibitem[Biewald(2020)]{wandb}
Lukas Biewald.
\newblock Experiment tracking with weights and biases, 2020.

\bibitem[Calders et~al.(2009)Calders, Kamiran, and
  Pechenizkiy]{caldersBuildingClassifiersIndependency2009}
Toon Calders, Faisal Kamiran, and Mykola Pechenizkiy.
\newblock Building {{Classifiers}} with {{Independency Constraints}}.
\newblock In \emph{2009 {{IEEE International Conference}} on {{Data Mining
  Workshops}}}, pp.\  13--18, December 2009.
\newblock \doi{10.1109/ICDMW.2009.83}.

\bibitem[Calvert et~al.(2016)Calvert, Price, Chettipally, Barton, Feldman,
  Hoffman, Jay, and Das]{calvertComputationalApproachEarly2016}
Jacob~S. Calvert, Daniel~A. Price, Uli~K. Chettipally, Christopher~W. Barton,
  Mitchell~D. Feldman, Jana~L. Hoffman, Melissa Jay, and Ritankar Das.
\newblock A computational approach to early sepsis detection.
\newblock \emph{Computers in Biology and Medicine}, 74:\penalty0 69--73, July
  2016.
\newblock ISSN 00104825.
\newblock \doi{10.1016/j.compbiomed.2016.05.003}.

\bibitem[Cheng et~al.(2018)Cheng, Waitman, Hu, and
  Liu]{chengPredictingInpatientAcute2018}
Peng Cheng, Lemuel~R. Waitman, Yong Hu, and Mei Liu.
\newblock Predicting {{Inpatient Acute Kidney Injury}} over {{Different Time
  Horizons}}: {{How Early}} and {{Accurate}}?
\newblock \emph{AMIA Annual Symposium Proceedings}, 2017:\penalty0 565--574,
  April 2018.
\newblock ISSN 1942-597X.

\bibitem[Cho et~al.(2014)Cho, {van Merrienboer}, Bahdanau, and
  Bengio]{choPropertiesNeuralMachine2014}
Kyunghyun Cho, Bart {van Merrienboer}, Dzmitry Bahdanau, and Yoshua Bengio.
\newblock On the {{Properties}} of {{Neural Machine Translation}}:
  {{Encoder-Decoder Approaches}}, October 2014.

\bibitem[{Dan Holtmann-Rice} et~al.(2018){Dan Holtmann-Rice}, Guadarrama, and
  Silberman]{google_ginconfig}
{Dan Holtmann-Rice}, Sergio Guadarrama, and Nathan Silberman.
\newblock Gin-config.
\newblock Google, 2018.

\bibitem[Du(2023)]{duPyPOTSPythonToolbox2023}
Wenjie Du.
\newblock {{PyPOTS}}: {{A Python Toolbox}} for {{Data Mining}} on
  {{Partially-Observed Time Series}}, May 2023.

\bibitem[{Eini-Porat} et~al.(2022){Eini-Porat}, Amir, Eytan, and
  Shalit]{eini-poratTellMeSomething2022}
Bar {Eini-Porat}, Ofra Amir, Danny Eytan, and Uri Shalit.
\newblock Tell me something interesting: {{Clinical}} utility of machine
  learning prediction models in the {{ICU}}.
\newblock \emph{Journal of Biomedical Informatics}, 132:\penalty0 104107,
  August 2022.
\newblock ISSN 15320464.
\newblock \doi{10.1016/j.jbi.2022.104107}.

\bibitem[Engoren et~al.(2020)Engoren, Seelhammer, Freundlich, Maile, Sigakis,
  and Schwann]{engorenComparisonSepsis2Systemic2020}
Milo Engoren, Troy Seelhammer, Robert~E. Freundlich, Michael~D. Maile, Matthew
  J.~G. Sigakis, and Thomas~A. Schwann.
\newblock A {{Comparison}} of {{Sepsis-2}} ({{Systemic Inflammatory Response
  Syndrome Based}}) to {{Sepsis-3}} ({{Sequential Organ Failure Assessment
  Based}}) {{Definitions}}---{{A Multicenter Retrospective Study}}*.
\newblock \emph{Critical Care Medicine}, 48\penalty0 (9):\penalty0 1258--1264,
  September 2020.
\newblock ISSN 0090-3493.
\newblock \doi{10.1097/CCM.0000000000004449}.

\bibitem[Falcon \& {team}(2023)Falcon and {team}]{falcon2019pytorch}
William Falcon and The PyTorch~Lightning {team}.
\newblock {{PyTorch}} lightning.
\newblock Zenodo, April 2023.

\bibitem[Fleuren et~al.(2020{\natexlab{a}})Fleuren, Klausch, Zwager,
  Schoonmade, Guo, Roggeveen, Swart, Girbes, Thoral, Ercole, Hoogendoorn, and
  Elbers]{fleurenMachineLearningPrediction2020}
Lucas~M. Fleuren, Thomas L.~T. Klausch, Charlotte~L. Zwager, Linda~J.
  Schoonmade, Tingjie Guo, Luca~F. Roggeveen, Eleonora~L. Swart, Armand R.~J.
  Girbes, Patrick Thoral, Ari Ercole, Mark Hoogendoorn, and Paul W.~G. Elbers.
\newblock Machine learning for the prediction of sepsis: A systematic review
  and meta-analysis of diagnostic test accuracy.
\newblock \emph{Intensive Care Medicine}, 46\penalty0 (3):\penalty0 383--400,
  2020{\natexlab{a}}.
\newblock ISSN 0342-4642.
\newblock \doi{10.1007/s00134-019-05872-y}.

\bibitem[Fleuren et~al.(2020{\natexlab{b}})Fleuren, Thoral, Shillan, Ercole,
  Elbers, Hoogendoorn, Gibbison, Klausch, Guo, Roggeveen, Swart, Girbes, and
  {Right Data Right Now Collaborators}]{fleurenMachineLearningIntensive2020}
Lucas~M. Fleuren, Patrick Thoral, Duncan Shillan, Ari Ercole, Paul W.~G.
  Elbers, Mark Hoogendoorn, Ben Gibbison, Thomas L.~T. Klausch, Tingjie Guo,
  Luca~F. Roggeveen, Eleonora~L. Swart, Armand R.~J. Girbes, and {Right Data
  Right Now Collaborators}.
\newblock Machine learning in intensive care medicine: Ready for take-off?
\newblock \emph{Intensive Care Medicine}, 46\penalty0 (7):\penalty0 1486--1488,
  July 2020{\natexlab{b}}.
\newblock ISSN 1432-1238.
\newblock \doi{10.1007/s00134-020-06045-y}.

\bibitem[Fomin et~al.(2020)Fomin, Anmol, Desroziers, Kriss, and
  Tejani]{pytorch-ignite}
V.~Fomin, J.~Anmol, S.~Desroziers, J.~Kriss, and A.~Tejani.
\newblock High-level library to help with training neural networks in
  {{PyTorch}}, 2020.

\bibitem[Futoma et~al.(2016)Futoma, Sendak, Cameron, and
  Heller]{futomaPredictingDiseaseProgression2016}
Joseph Futoma, Mark Sendak, Blake Cameron, and Katherine Heller.
\newblock Predicting {{Disease Progression}} with a {{Model}} for
  {{Multivariate Longitudinal Clinical Data}}.
\newblock In \emph{Proceedings of the 1st {{Machine Learning}} for {{Healthcare
  Conference}}}, pp.\  42--54. PMLR, December 2016.

\bibitem[Grinsztajn et~al.(2022)Grinsztajn, Oyallon, and
  Varoquaux]{grinsztajnWhyTreebasedModels2022}
Leo Grinsztajn, Edouard Oyallon, and Gael Varoquaux.
\newblock Why do tree-based models still outperform deep learning on typical
  tabular data?
\newblock In \emph{Thirty-Sixth {{Conference}} on {{Neural Information
  Processing Systems Datasets}} and {{Benchmarks Track}}}, September 2022.

\bibitem[Guo et~al.(2020)Guo, Lu, and Chen]{guoEvaluationTimeSeries2020}
Chonghui Guo, Menglin Lu, and Jingfeng Chen.
\newblock An evaluation of time series summary statistics as features for
  clinical prediction tasks.
\newblock \emph{BMC Medical Informatics and Decision Making}, 20\penalty0
  (1):\penalty0 48, December 2020.
\newblock ISSN 1472-6947.
\newblock \doi{10.1186/s12911-020-1063-x}.

\bibitem[Gupta et~al.(2022)Gupta, Gallamoza, Cutrona, Dhakal, Poulain, and
  Beheshti]{guptaExtensiveDataProcessing2022a}
Mehak Gupta, Brennan Gallamoza, Nicolas Cutrona, Pranjal Dhakal, Raphael
  Poulain, and Rahmatollah Beheshti.
\newblock An {{Extensive Data Processing Pipeline}} for {{MIMIC-IV}}, September
  2022.

\bibitem[Hardt et~al.(2016)Hardt, Price, Price, and
  Srebro]{hardtEqualityOpportunitySupervised2016}
Moritz Hardt, Eric Price, Eric Price, and Nati Srebro.
\newblock Equality of {{Opportunity}} in {{Supervised Learning}}.
\newblock In \emph{Advances in {{Neural Information Processing Systems}}},
  volume~29. Curran Associates, Inc., 2016.

\bibitem[Harutyunyan et~al.(2019)Harutyunyan, Khachatrian, Kale, Ver~Steeg, and
  Galstyan]{harutyunyanMultitaskLearningBenchmarking2019}
Hrayr Harutyunyan, Hrant Khachatrian, David~C. Kale, Greg Ver~Steeg, and Aram
  Galstyan.
\newblock Multitask learning and benchmarking with clinical time series data.
\newblock \emph{Scientific Data}, 6\penalty0 (1):\penalty0 96, December 2019.
\newblock ISSN 2052-4463.
\newblock \doi{10.1038/s41597-019-0103-9}.

\bibitem[Head et~al.(2021)Head, Kumar, Nahrstaedt, Louppe, and
  Shcherbatyi]{headScikitoptimizeScikitoptimize2021}
Tim Head, Manoj Kumar, Holger Nahrstaedt, Gilles Louppe, and Iaroslav
  Shcherbatyi.
\newblock Scikit-optimize/scikit-optimize.
\newblock Zenodo, October 2021.

\bibitem[Ho et~al.(2020)Ho, Jain, and
  Abbeel]{hoDenoisingDiffusionProbabilistic2020}
Jonathan Ho, Ajay Jain, and Pieter Abbeel.
\newblock Denoising {{Diffusion Probabilistic Models}}.
\newblock In \emph{Advances in {{Neural Information Processing Systems}}},
  volume~33, pp.\  6840--6851. Curran Associates, Inc., 2020.

\bibitem[Hochreiter \& Schmidhuber(1997)Hochreiter and
  Schmidhuber]{hochreiterLongShortTermMemory1997}
Sepp Hochreiter and J{\"u}rgen Schmidhuber.
\newblock Long {{Short-Term Memory}}.
\newblock Technical Report~9, TU Munich, 1997.

\bibitem[Huang et~al.(2021)Huang, Chen, Wang, Liu, and
  Liu]{huangInterpretableTemporalConvolutional2021}
Wei Huang, Yuwen Chen, Peng Wang, Xiang Liu, and Shuguang Liu.
\newblock An {{Interpretable Temporal Convolutional Network Model}} for {{Acute
  Kidney Injury Prediction}} in the {{Intensive Care Unit}}.
\newblock In \emph{2021 {{IEEE International Conference}} on {{Bioinformatics}}
  and {{Biomedicine}} ({{BIBM}})}, pp.\  3021--3028, December 2021.
\newblock \doi{10.1109/BIBM52615.2021.9669653}.

\bibitem[Hyland(2020)]{hylandEarlyPredictionCirculatory2020}
Stephanie~L Hyland.
\newblock Early prediction of circulatory failure in the intensive care unit
  using machine learning.
\newblock \emph{Nature MedIcIne}, 26:\penalty0 28, 2020.

\bibitem[Jarrett et~al.(2021)Jarrett, Bica, Ercole, Yoon, and
  Qian]{jarrettCLAIRVOYANCEPIPELINETOOLKIT2021}
Daniel Jarrett, Ioana Bica, Ari Ercole, Jinsung Yoon, and Zhaozhi Qian.
\newblock {{CLAIRVOYANCE}}: {{A PIPELINE TOOLKIT FOR MEDICAL TIME SERIES}}.
\newblock \emph{International Conference on Learning Representations}, pp.\
  ~32, 2021.

\bibitem[Jarrett et~al.(2022)Jarrett, Cebere, Liu, Curth, and {van der
  Schaar}]{jarrettHyperImputeGeneralizedIterative2022}
Daniel Jarrett, Bogdan Cebere, Tennison Liu, Alicia Curth, and Mihaela {van der
  Schaar}.
\newblock {{HyperImpute}}: {{Generalized Iterative Imputation}} with
  {{Automatic Model Selection}}, June 2022.

\bibitem[Jin et~al.(2023)Jin, Chen, Chen, Hu, Hu, and
  Zhang]{jinEstablishmentChineseCritical2023}
Senjun Jin, Lin Chen, Kun Chen, Chaozhou Hu, Sheng'an Hu, and Zhongheng Zhang.
\newblock Establishment of a {{Chinese}} critical care database from electronic
  healthcare records in a tertiary care medical center.
\newblock \emph{Scientific Data}, 10\penalty0 (1):\penalty0 49, January 2023.
\newblock ISSN 2052-4463.
\newblock \doi{10.1038/s41597-023-01952-3}.

\bibitem[Johnson et~al.(2017)Johnson, Pollard, and
  Mark]{johnsonReproducibilityCriticalCare2017}
Alistair E.~W. Johnson, Tom~J. Pollard, and Roger~G. Mark.
\newblock Reproducibility in critical care: A mortality prediction case study.
\newblock In \emph{Proceedings of the 2nd {{Machine Learning}} for {{Healthcare
  Conference}}}, pp.\  361--376. PMLR, November 2017.

\bibitem[Johnson et~al.(2023)Johnson, Bulgarelli, Shen, Gayles, Shammout,
  Horng, Pollard, Moody, Gow, Lehman, Celi, and
  Mark]{johnsonMIMICIVFreelyAccessible2023}
Alistair E.~W. Johnson, Lucas Bulgarelli, Lu~Shen, Alvin Gayles, Ayad Shammout,
  Steven Horng, Tom~J. Pollard, Benjamin Moody, Brian Gow, Li-wei~H. Lehman,
  Leo~A. Celi, and Roger~G. Mark.
\newblock {{MIMIC-IV}}, a freely accessible electronic health record dataset.
\newblock \emph{Scientific Data}, 10\penalty0 (1):\penalty0 1, January 2023.
\newblock ISSN 2052-4463.
\newblock \doi{10.1038/s41597-022-01899-x}.

\bibitem[Johnson et~al.(2016)Johnson, Pollard, Shen, Lehman, Feng, Ghassemi,
  Moody, Szolovits, Anthony~Celi, and
  Mark]{johnsonMIMICIIIFreelyAccessible2016}
Alistair~E.W. Johnson, Tom~J. Pollard, Lu~Shen, Li-wei~H. Lehman, Mengling
  Feng, Mohammad Ghassemi, Benjamin Moody, Peter Szolovits, Leo Anthony~Celi,
  and Roger~G. Mark.
\newblock {{MIMIC-III}}, a freely accessible critical care database.
\newblock \emph{Scientific Data}, 3\penalty0 (1):\penalty0 160035, December
  2016.
\newblock ISSN 2052-4463.
\newblock \doi{10.1038/sdata.2016.35}.

\bibitem[{KDIGO}(2012)]{kdigoKidneyDiseaseImproving2012}
{KDIGO}.
\newblock Kidney {{Disease}}: {{Improving Global Outcomes}} ({{KDIGO}}) {{Acute
  Kidney Injury Work Group}}: {{KDIGO}} clinical practice guideline for acute
  kidney injury.
\newblock \emph{Kidney International Supplements}, 2\penalty0 (1), March 2012.
\newblock ISSN 21571716.
\newblock \doi{10.1038/kisup.2012.2}.

\bibitem[Ke et~al.(2017)Ke, Meng, Finley, Wang, Chen, Ma, Ye, and
  Liu]{keLightGBMHighlyEfficient2017}
Guolin Ke, Qi~Meng, Thomas Finley, Taifeng Wang, Wei Chen, Weidong Ma, Qiwei
  Ye, and Tie-Yan Liu.
\newblock {{LightGBM}}: {{A Highly Efficient Gradient Boosting Decision Tree}}.
\newblock In \emph{Advances in {{Neural Information Processing Systems}}},
  volume~30. Curran Associates, Inc., 2017.

\bibitem[Kelly et~al.(2019)Kelly, Karthikesalingam, Suleyman, Corrado, and
  King]{kellyKeyChallengesDelivering2019}
Christopher~J. Kelly, Alan Karthikesalingam, Mustafa Suleyman, Greg Corrado,
  and Dominic King.
\newblock Key challenges for delivering clinical impact with artificial
  intelligence.
\newblock \emph{BMC Medicine}, 17\penalty0 (1):\penalty0 195, December 2019.
\newblock ISSN 1741-7015.
\newblock \doi{10.1186/s12916-019-1426-2}.

\bibitem[Kok et~al.(2020)Kok, Jahmunah, Oh, Zhou, Gururajan, Tao, Cheong,
  Gururajan, Molinari, and Acharya]{kokAutomatedPredictionSepsis2020}
Christopher Kok, V.~Jahmunah, Shu~Lih Oh, Xujuan Zhou, Raj Gururajan, Xiaohui
  Tao, Kang~Hao Cheong, Rashmi Gururajan, Filippo Molinari, and U.Rajendra
  Acharya.
\newblock Automated prediction of sepsis using temporal convolutional network.
\newblock \emph{Computers in Biology and Medicine}, 127:\penalty0 103957,
  December 2020.
\newblock ISSN 00104825.
\newblock \doi{10.1016/j.compbiomed.2020.103957}.

\bibitem[Kong et~al.(2021)Kong, Ping, Huang, Zhao, and
  Catanzaro]{kongDiffWaveVersatileDiffusion2021}
Zhifeng Kong, Wei Ping, Jiaji Huang, Kexin Zhao, and Bryan Catanzaro.
\newblock {{DiffWave}}: {{A Versatile Diffusion Model}} for {{Audio
  Synthesis}}, March 2021.

\bibitem[Koyner et~al.(2018)Koyner, Carey, Edelson, and
  Churpek]{koynerDevelopmentMachineLearning2018}
Jay~L. Koyner, Kyle~A. Carey, Dana~P. Edelson, and Matthew~M. Churpek.
\newblock The {{Development}} of a {{Machine Learning Inpatient Acute Kidney
  Injury Prediction Model}}*:.
\newblock \emph{Critical Care Medicine}, 46\penalty0 (7):\penalty0 1070--1077,
  July 2018.
\newblock ISSN 0090-3493.
\newblock \doi{10.1097/CCM.0000000000003123}.

\bibitem[Lauritsen et~al.(2020)Lauritsen, Kal{\o}r, Kongsgaard, Lauritsen,
  J{\o}rgensen, Lange, and Thiesson]{lauritsenEarlyDetectionSepsis2020}
Simon~Meyer Lauritsen, Mads~Ellersgaard Kal{\o}r, Emil~Lund Kongsgaard,
  Katrine~Meyer Lauritsen, Marianne~Johansson J{\o}rgensen, Jeppe Lange, and
  Bo~Thiesson.
\newblock Early detection of sepsis utilizing deep learning on electronic
  health record event sequences.
\newblock \emph{Artificial Intelligence in Medicine}, 104:\penalty0 101820,
  April 2020.
\newblock ISSN 09333657.
\newblock \doi{10.1016/j.artmed.2020.101820}.

\bibitem[Levy et~al.(2003)Levy, Fink, Marshall, Abraham, Angus, Cook, Cohen,
  Opal, Vincent, Ramsay, and {International Sepsis Definitions
  Conference}]{levy2001SCCMESICM2003}
Mitchell~M. Levy, Mitchell~P. Fink, John~C. Marshall, Edward Abraham, Derek
  Angus, Deborah Cook, Jonathan Cohen, Steven~M. Opal, Jean-Louis Vincent,
  Graham Ramsay, and {International Sepsis Definitions Conference}.
\newblock 2001 {{SCCM}}/{{ESICM}}/{{ACCP}}/{{ATS}}/{{SIS International Sepsis
  Definitions Conference}}.
\newblock \emph{Intensive Care Medicine}, 29\penalty0 (4):\penalty0 530--538,
  April 2003.
\newblock ISSN 0342-4642.
\newblock \doi{10.1007/s00134-003-1662-x}.

\bibitem[Lim et~al.(2021)Lim, Ar{\i}k, Loeff, and
  Pfister]{limTemporalFusionTransformers2021}
Bryan Lim, Sercan~{\"O}. Ar{\i}k, Nicolas Loeff, and Tomas Pfister.
\newblock Temporal {{Fusion Transformers}} for interpretable multi-horizon time
  series forecasting.
\newblock \emph{International Journal of Forecasting}, 37\penalty0
  (4):\penalty0 1748--1764, October 2021.
\newblock ISSN 0169-2070.
\newblock \doi{10.1016/j.ijforecast.2021.03.012}.

\bibitem[Lu et~al.(2022)Lu, Xu, Nguyen, Geng, Pfob, and
  {Sidey-Gibbons}]{luMachineLearningBased2022}
Sheng-Chieh Lu, Cai Xu, Chandler~H. Nguyen, Yimin Geng, Andr{\'e} Pfob, and
  Chris {Sidey-Gibbons}.
\newblock Machine {{Learning}}--{{Based Short-Term Mortality Prediction
  Models}} for {{Patients With Cancer Using Electronic Health Record Data}}:
  {{Systematic Review}} and {{Critical Appraisal}}.
\newblock \emph{JMIR Medical Informatics}, 10\penalty0 (3):\penalty0 e33182,
  March 2022.
\newblock \doi{10.2196/33182}.

\bibitem[Mandyam et~al.(2021)Mandyam, Yoo, Soules, Laudanski, and
  Engelhardt]{mandyamCOPECATCleaningOrganization2021}
Aishwarya Mandyam, Elizabeth~C. Yoo, Jeff Soules, Krzysztof Laudanski, and
  Barbara~E. Engelhardt.
\newblock {{COP-E-CAT}}: Cleaning and organization pipeline for {{EHR}}
  computational and analytic tasks.
\newblock In \emph{Proceedings of the 12th {{ACM Conference}} on
  {{Bioinformatics}}, {{Computational Biology}}, and {{Health Informatics}}},
  pp.\  1--9, Gainesville Florida, August 2021. ACM.
\newblock ISBN 978-1-4503-8450-6.
\newblock \doi{10.1145/3459930.3469536}.

\bibitem[Medic et~al.(2019)Medic, Klie{\ss}, Atallah, Weichert, Panda, Postma,
  and {El-Kerdi}]{medicEvidencebasedClinicalDecision2019}
Goran Medic, Melodi~Kosaner Klie{\ss}, Louis Atallah, Jochen Weichert, Saswat
  Panda, Maarten Postma, and Amer {El-Kerdi}.
\newblock Evidence-based {{Clinical Decision Support Systems}} for the
  prediction and detection of three disease states in critical care: {{A}}
  systematic literature review.
\newblock \emph{F1000Research}, 8:\penalty0 1728, 2019.
\newblock ISSN 2046-1402.
\newblock \doi{10.12688/f1000research.20498.2}.

\bibitem[Merath et~al.(2020)Merath, Hyer, Mehta, Farooq, Bagante, Sahara,
  Tsilimigras, Beal, Paredes, Wu, Ejaz, and
  Pawlik]{merathUseMachineLearning2020}
Katiuscha Merath, J.~Madison Hyer, Rittal Mehta, Ayesha Farooq, Fabio Bagante,
  Kota Sahara, Diamantis~I. Tsilimigras, Eliza Beal, Anghela~Z. Paredes, Lu~Wu,
  Aslam Ejaz, and Timothy~M. Pawlik.
\newblock Use of {{Machine Learning}} for {{Prediction}} of {{Patient Risk}} of
  {{Postoperative Complications After Liver}}, {{Pancreatic}}, and {{Colorectal
  Surgery}}.
\newblock \emph{Journal of Gastrointestinal Surgery}, 24\penalty0 (8):\penalty0
  1843--1851, August 2020.
\newblock ISSN 1091-255X, 1873-4626.
\newblock \doi{10.1007/s11605-019-04338-2}.

\bibitem[Moor et~al.(2019)Moor, Horn, Rieck, Roqueiro, and
  Borgwardt]{moorEarlyRecognitionSepsis2019}
Michael Moor, Max Horn, Bastian Rieck, Damian Roqueiro, and Karsten Borgwardt.
\newblock Early {{Recognition}} of {{Sepsis}} with {{Gaussian Process Temporal
  Convolutional Networks}} and {{Dynamic Time Warping}}.
\newblock In \emph{Proceedings of the 4th {{Machine Learning}} for {{Healthcare
  Conference}}}, pp.\  2--26. PMLR, October 2019.

\bibitem[Moor et~al.(2021{\natexlab{a}})Moor, Bennet, Plecko, Horn, Rieck,
  Meinshausen, B{\"u}hlmann, and Borgwardt]{moorPredictingSepsisMultisite2021}
Michael Moor, Nicolas Bennet, Drago Plecko, Max Horn, Bastian Rieck, Nicolai
  Meinshausen, Peter B{\"u}hlmann, and Karsten Borgwardt.
\newblock Predicting sepsis in multi-site, multi-national intensive care
  cohorts using deep learning, July 2021{\natexlab{a}}.

\bibitem[Moor et~al.(2021{\natexlab{b}})Moor, Rieck, Horn, Jutzeler, and
  Borgwardt]{moorEarlyPredictionSepsis2021}
Michael Moor, Bastian Rieck, Max Horn, Catherine~R. Jutzeler, and Karsten
  Borgwardt.
\newblock Early {{Prediction}} of {{Sepsis}} in the {{ICU Using Machine
  Learning}}: {{A Systematic Review}}.
\newblock \emph{Frontiers in Medicine}, 8, 2021{\natexlab{b}}.
\newblock ISSN 2296-858X.

\bibitem[Muralitharan et~al.(2021)Muralitharan, Nelson, Di, McGillion,
  Devereaux, Barr, and Petch]{muralitharanMachineLearningBased2021}
Sankavi Muralitharan, Walter Nelson, Shuang Di, Michael McGillion,
  Pj~Devereaux, Neil~Grant Barr, and Jeremy Petch.
\newblock Machine {{Learning}}--{{Based Early Warning Systems}} for {{Clinical
  Deterioration}}: {{Systematic Scoping Review}}.
\newblock \emph{Journal of Medical Internet Research}, 23\penalty0
  (2):\penalty0 e25187, February 2021.
\newblock ISSN 1438-8871.
\newblock \doi{10.2196/25187}.

\bibitem[{Nicki Skafte Detlefsen} et~al.(2022){Nicki Skafte Detlefsen}, {Jiri
  Borovec}, {Justus Schock}, {Ananya Harsh}, {Teddy Koker}, {Luca Di Liello},
  {Daniel Stancl}, {Changsheng Quan}, {Maxim Grechkin}, and {William
  Falcon}]{nickiskaftedetlefsenTorchMetricsMeasuringReproducibility2022}
{Nicki Skafte Detlefsen}, {Jiri Borovec}, {Justus Schock}, {Ananya Harsh},
  {Teddy Koker}, {Luca Di Liello}, {Daniel Stancl}, {Changsheng Quan}, {Maxim
  Grechkin}, and {William Falcon}.
\newblock {{TorchMetrics}} - measuring reproducibility in {{PyTorch}}, February
  2022.

\bibitem[Nikkinen et~al.(2022)Nikkinen, Kolehmainen, Aaltonen, J{\"a}ms{\"a},
  Alahuhta, and Vakkala]{nikkinenDevelopingSupervisedMachine2022}
Okke Nikkinen, Timo Kolehmainen, Toni Aaltonen, Elias J{\"a}ms{\"a}, Seppo
  Alahuhta, and Merja Vakkala.
\newblock Developing a supervised machine learning model for predicting
  perioperative acute kidney injury in arthroplasty patients.
\newblock \emph{Computers in Biology and Medicine}, 144:\penalty0 105351, May
  2022.
\newblock ISSN 00104825.
\newblock \doi{10.1016/j.compbiomed.2022.105351}.

\bibitem[Oliver et~al.(2023)Oliver, Allyn, Carencotte, Allou, and
  Ferdynus]{oliverIntroducingBlendedICUDataset2023}
Matthieu Oliver, J{\'e}r{\^o}me Allyn, R{\'e}mi Carencotte, Nicolas Allou, and
  Cyril Ferdynus.
\newblock Introducing the {{BlendedICU}} dataset, the first harmonized,
  international intensive care dataset.
\newblock \emph{Journal of Biomedical Informatics}, 146:\penalty0 104502,
  October 2023.
\newblock ISSN 1532-0464.
\newblock \doi{10.1016/j.jbi.2023.104502}.

\bibitem[Pan et~al.(2019)Pan, Du, Ngiam, Wang, Shum, and
  Feng]{panSelfCorrectingDeepLearning2019}
Ziyuan Pan, Hao Du, Kee~Yuan Ngiam, Fei Wang, Ping Shum, and Mengling Feng.
\newblock A {{Self-Correcting Deep Learning Approach}} to {{Predict Acute
  Conditions}} in {{Critical Care}}.
\newblock \emph{arXiv:1901.04364 [cs, stat]}, January 2019.

\bibitem[Pedregosa et~al.(2011)Pedregosa, Varoquaux, Gramfort, Michel, Thirion,
  Grisel, Blondel, Prettenhofer, Weiss, Dubourg, Vanderplas, Passos,
  Cournapeau, Brucher, Perrot, and Duchesnay]{scikit-learn}
F.~Pedregosa, G.~Varoquaux, A.~Gramfort, V.~Michel, B.~Thirion, O.~Grisel,
  M.~Blondel, P.~Prettenhofer, R.~Weiss, V.~Dubourg, J.~Vanderplas, A.~Passos,
  D.~Cournapeau, M.~Brucher, M.~Perrot, and E.~Duchesnay.
\newblock Scikit-learn: {{Machine}} learning in {{Python}}.
\newblock \emph{Journal of Machine Learning Research}, 12:\penalty0 2825--2830,
  2011.

\bibitem[Perotte et~al.(2015)Perotte, Ranganath, Hirsch, Blei, and
  Elhadad]{perotteRiskPredictionChronic2015}
Adler Perotte, Rajesh Ranganath, Jamie~S Hirsch, David Blei, and No{\'e}mie
  Elhadad.
\newblock Risk prediction for chronic kidney disease progression using
  heterogeneous electronic health record data and time series analysis.
\newblock \emph{Journal of the American Medical Informatics Association :
  JAMIA}, 22\penalty0 (4):\penalty0 872--880, July 2015.
\newblock ISSN 1067-5027.
\newblock \doi{10.1093/jamia/ocv024}.

\bibitem[Pirracchio et~al.(2015)Pirracchio, Petersen, Carone, Rigon, Chevret,
  and {van der Laan}]{pirracchioMortalityPredictionIntensive2015a}
Romain Pirracchio, Maya~L Petersen, Marco Carone, Matthieu~Resche Rigon, Sylvie
  Chevret, and Mark~J {van der Laan}.
\newblock Mortality prediction in intensive care units with the {{Super ICU
  Learner Algorithm}} ({{SICULA}}): {{A}} population-based study.
\newblock \emph{The Lancet Respiratory Medicine}, 3\penalty0 (1):\penalty0
  42--52, January 2015.
\newblock ISSN 22132600.
\newblock \doi{10.1016/S2213-2600(14)70239-5}.

\bibitem[Pollard et~al.(2018)Pollard, Johnson, Raffa, Celi, Mark, and
  Badawi]{pollardEICUCollaborativeResearch2018}
Tom~J. Pollard, Alistair E.~W. Johnson, Jesse~D. Raffa, Leo~A. Celi, Roger~G.
  Mark, and Omar Badawi.
\newblock The {{eICU Collaborative Research Database}}, a freely available
  multi-center database for critical care research.
\newblock \emph{Scientific Data}, 5\penalty0 (1):\penalty0 180178, December
  2018.
\newblock ISSN 2052-4463.
\newblock \doi{10.1038/sdata.2018.178}.

\bibitem[Purushotham et~al.(2018)Purushotham, Meng, Che, and
  Liu]{purushothamBenchmarkingDeepLearning2018}
Sanjay Purushotham, Chuizheng Meng, Zhengping Che, and Yan Liu.
\newblock Benchmarking deep learning models on large healthcare datasets.
\newblock \emph{Journal of Biomedical Informatics}, 83:\penalty0 112--134, July
  2018.
\newblock ISSN 15320464.
\newblock \doi{10.1016/j.jbi.2018.04.007}.

\bibitem[Rank et~al.(2020)Rank, Pfahringer, Kempfert, Stamm, K{\"u}hne,
  Schoenrath, Falk, Eickhoff, and
  Meyer]{rankDeeplearningbasedRealtimePrediction2020}
Nina Rank, Boris Pfahringer, J{\"o}rg Kempfert, Christof Stamm, Titus
  K{\"u}hne, Felix Schoenrath, Volkmar Falk, Carsten Eickhoff, and Alexander
  Meyer.
\newblock Deep-learning-based real-time prediction of acute kidney injury
  outperforms human predictive performance.
\newblock \emph{npj Digital Medicine}, 3\penalty0 (1):\penalty0 139, December
  2020.
\newblock ISSN 2398-6352.
\newblock \doi{10.1038/s41746-020-00346-8}.

\bibitem[Reyna et~al.(2019)Reyna, Josef, Seyedi, Jeter, Shashikumar,
  Brandon~Westover, Sharma, Nemati, and
  Clifford]{reynaEarlyPredictionSepsis2019}
Matthew~A Reyna, Chris Josef, Salman Seyedi, Russell Jeter, Supreeth~P
  Shashikumar, M~Brandon~Westover, Ashish Sharma, Shamim Nemati, and Gari~D
  Clifford.
\newblock Early {{Prediction}} of {{Sepsis}} from {{Clinical Data}}: The
  {{PhysioNet}}/{{Computing}} in {{Cardiology Challenge}} 2019.
\newblock In \emph{2019 {{Computing}} in {{Cardiology}} ({{CinC}})}, pp.\  Page
  1--Page 4, September 2019.
\newblock \doi{10.23919/CinC49843.2019.9005736}.

\bibitem[Rodemund et~al.(2023)Rodemund, Wernly, Jung, Cozowicz, and
  Kok{\"o}fer]{rodemundSalzburgIntensiveCare2023}
Niklas Rodemund, Bernhard Wernly, Christian Jung, Crispiana Cozowicz, and
  Andreas Kok{\"o}fer.
\newblock The {{Salzburg Intensive Care}} database ({{SICdb}}): An openly
  available critical care dataset.
\newblock \emph{Intensive Care Medicine}, April 2023.
\newblock ISSN 1432-1238.
\newblock \doi{10.1007/s00134-023-07046-3}.

\bibitem[Ronneberger et~al.(2015)Ronneberger, Fischer, and
  Brox]{ronnebergerUNetConvolutionalNetworks2015}
Olaf Ronneberger, Philipp Fischer, and Thomas Brox.
\newblock U-{{Net}}: {{Convolutional Networks}} for {{Biomedical Image
  Segmentation}}.
\newblock In Nassir Navab, Joachim Hornegger, William~M. Wells, and
  Alejandro~F. Frangi (eds.), \emph{Medical {{Image Computing}} and
  {{Computer-Assisted Intervention}} -- {{MICCAI}} 2015}, Lecture {{Notes}} in
  {{Computer Science}}, pp.\  234--241, Cham, 2015. Springer International
  Publishing.
\newblock ISBN 978-3-319-24574-4.
\newblock \doi{10.1007/978-3-319-24574-4_28}.

\bibitem[Sarwar et~al.(2023)Sarwar, Seifollahi, Chan, Zhang, Aksakalli, Hudson,
  Verspoor, and Cavedon]{sarwarSecondaryUseElectronic2023}
Tabinda Sarwar, Sattar Seifollahi, Jeffrey Chan, Xiuzhen Zhang, Vural
  Aksakalli, Irene Hudson, Karin Verspoor, and Lawrence Cavedon.
\newblock The {{Secondary Use}} of {{Electronic Health Records}} for {{Data
  Mining}}: {{Data Characteristics}} and {{Challenges}}.
\newblock \emph{ACM Computing Surveys}, 55\penalty0 (2):\penalty0 1--40, March
  2023.
\newblock ISSN 0360-0300, 1557-7341.
\newblock \doi{10.1145/3490234}.

\bibitem[Sauer et~al.(2022{\natexlab{a}})Sauer, Chen, Hyland, Girbes, Elbers,
  and Celi]{sauerLeveragingElectronicHealth2022}
Christopher~M Sauer, Li-Ching Chen, Stephanie~L Hyland, Armand Girbes, Paul
  Elbers, and Leo~A Celi.
\newblock Leveraging electronic health records for data science: Common
  pitfalls and how to avoid them.
\newblock \emph{The Lancet Digital Health}, 4\penalty0 (12):\penalty0
  e893--e898, December 2022{\natexlab{a}}.
\newblock ISSN 25897500.
\newblock \doi{10.1016/S2589-7500(22)00154-6}.

\bibitem[Sauer et~al.(2022{\natexlab{b}})Sauer, Dam, Celi, Faltys, {de la Hoz},
  Adhikari, Ziesemer, Girbes, Thoral, and
  Elbers]{sauerSystematicReviewComparison2022}
Christopher~M. Sauer, Tariq~A. Dam, Leo~A. Celi, Martin Faltys, Miguel A.~A.
  {de la Hoz}, Lasith Adhikari, Kirsten~A. Ziesemer, Armand Girbes, Patrick~J.
  Thoral, and Paul Elbers.
\newblock Systematic {{Review}} and {{Comparison}} of {{Publicly Available ICU
  Data Sets}}---{{A Decision Guide}} for {{Clinicians}} and {{Data
  Scientists}}.
\newblock \emph{Critical Care Medicine}, 50\penalty0 (6):\penalty0 e581--e588,
  June 2022{\natexlab{b}}.
\newblock ISSN 0090-3493.
\newblock \doi{10.1097/CCM.0000000000005517}.

\bibitem[Saveliev \& {van der Schaar}(2023)Saveliev and {van der
  Schaar}]{savelievTemporAIFacilitatingMachine2023}
Evgeny~S. Saveliev and Mihaela {van der Schaar}.
\newblock {{TemporAI}}: {{Facilitating Machine Learning Innovation}} in {{Time
  Domain Tasks}} for {{Medicine}}, January 2023.

\bibitem[Seymour et~al.(2016)Seymour, Liu, Iwashyna, Brunkhorst, Rea, Scherag,
  Rubenfeld, Kahn, {Shankar-Hari}, Singer, Deutschman, Escobar, and
  Angus]{seymourAssessmentClinicalCriteria2016}
Christopher~W. Seymour, Vincent~X. Liu, Theodore~J. Iwashyna, Frank~M.
  Brunkhorst, Thomas~D. Rea, Andr{\'e} Scherag, Gordon Rubenfeld, Jeremy~M.
  Kahn, Manu {Shankar-Hari}, Mervyn Singer, Clifford~S. Deutschman, Gabriel~J.
  Escobar, and Derek~C. Angus.
\newblock Assessment of {{Clinical Criteria}} for {{Sepsis}}: {{For}} the
  {{Third International Consensus Definitions}} for {{Sepsis}} and {{Septic
  Shock}} ({{Sepsis-3}}).
\newblock \emph{JAMA}, 315\penalty0 (8):\penalty0 762, February 2016.
\newblock ISSN 0098-7484.
\newblock \doi{10.1001/jama.2016.0288}.

\bibitem[Shamout et~al.(2021)Shamout, Zhu, and
  Clifton]{shamoutMachineLearningClinical2021}
Farah Shamout, Tingting Zhu, and David~A. Clifton.
\newblock Machine {{Learning}} for {{Clinical Outcome Prediction}}.
\newblock \emph{IEEE Reviews in Biomedical Engineering}, 14:\penalty0 116--126,
  2021.
\newblock ISSN 1937-3333, 1941-1189.
\newblock \doi{10.1109/RBME.2020.3007816}.

\bibitem[Sharma et~al.(2017)Sharma, Shukla, Tiwari, and
  Mishra]{sharmaMortalityPredictionICU2017}
Alok Sharma, Anupam Shukla, Ritu Tiwari, and Apoorva Mishra.
\newblock Mortality {{Prediction}} of {{ICU}} patients using {{Machine
  Leaning}}: {{A}} survey.
\newblock In \emph{Proceedings of the {{International Conference}} on
  {{Compute}} and {{Data Analysis}} - {{ICCDA}} '17}, pp.\  49--53, Lakeland,
  FL, USA, 2017. ACM Press.
\newblock ISBN 978-1-4503-5241-3.
\newblock \doi{10.1145/3093241.3093267}.

\bibitem[Sheikhalishahi et~al.(2020)Sheikhalishahi, Balaraman, and
  Osmani]{sheikhalishahiBenchmarkingMachineLearning2020}
Seyedmostafa Sheikhalishahi, Vevake Balaraman, and Venet Osmani.
\newblock Benchmarking machine learning models on multi-centre {{eICU}}
  critical care dataset.
\newblock \emph{PLOS ONE}, 15\penalty0 (7):\penalty0 e0235424, July 2020.
\newblock ISSN 1932-6203.
\newblock \doi{10.1371/journal.pone.0235424}.

\bibitem[Shillan et~al.(2019)Shillan, Sterne, Champneys, and
  Gibbison]{shillanUseMachineLearning2019}
Duncan Shillan, Jonathan A.~C. Sterne, Alan Champneys, and Ben Gibbison.
\newblock Use of machine learning to analyse routinely collected intensive care
  unit data: A systematic review.
\newblock \emph{Critical Care}, 23\penalty0 (1):\penalty0 284, December 2019.
\newblock ISSN 1364-8535.
\newblock \doi{10.1186/s13054-019-2564-9}.

\bibitem[Silva et~al.(2012)Silva, Moody, Scott, Celi, and
  Mark]{silvaPredictingInHospitalMortality2012}
Ikaro Silva, George Moody, Daniel~J. Scott, Leo~A. Celi, and Roger~G. Mark.
\newblock Predicting {{In-Hospital Mortality}} of {{ICU Patients}}: {{The
  PhysioNet}}/{{Computing}} in {{Cardiology Challenge}} 2012.
\newblock \emph{Computing in Cardiology}, 39:\penalty0 245--248, 2012.
\newblock ISSN 2325-8861.

\bibitem[Singer et~al.(2016)Singer, Deutschman, Seymour, {Shankar-Hari},
  Annane, Bauer, Bellomo, Bernard, Chiche, Coopersmith, Hotchkiss, Levy,
  Marshall, Martin, Opal, Rubenfeld, {van der Poll}, Vincent, and
  Angus]{singerThirdInternationalConsensus2016}
Mervyn Singer, Clifford~S. Deutschman, Christopher~Warren Seymour, Manu
  {Shankar-Hari}, Djillali Annane, Michael Bauer, Rinaldo Bellomo, Gordon~R.
  Bernard, Jean-Daniel Chiche, Craig~M. Coopersmith, Richard~S. Hotchkiss,
  Mitchell~M. Levy, John~C. Marshall, Greg~S. Martin, Steven~M. Opal, Gordon~D.
  Rubenfeld, Tom {van der Poll}, Jean-Louis Vincent, and Derek~C. Angus.
\newblock The {{Third International Consensus Definitions}} for {{Sepsis}} and
  {{Septic Shock}} ({{Sepsis-3}}).
\newblock \emph{JAMA}, 315\penalty0 (8):\penalty0 801--810, February 2016.
\newblock ISSN 0098-7484.
\newblock \doi{10.1001/jama.2016.0287}.

\bibitem[Snoek et~al.(2012)Snoek, Larochelle, and
  Adams]{snoekPracticalBayesianOptimization2012}
Jasper Snoek, Hugo Larochelle, and Ryan~P. Adams.
\newblock Practical {{Bayesian Optimization}} of {{Machine Learning
  Algorithms}}, August 2012.

\bibitem[Syed et~al.(2021)Syed, Syed, Sexton, Syeda, Garza, Zozus, Syed, Begum,
  Syed, Sanford, and Prior]{syedApplicationMachineLearning2021}
Mahanazuddin Syed, Shorabuddin Syed, Kevin Sexton, Hafsa~Bareen Syeda, Maryam
  Garza, Meredith Zozus, Farhanuddin Syed, Salma Begum, Abdullah~Usama Syed,
  Joseph Sanford, and Fred Prior.
\newblock Application of {{Machine Learning}} in {{Intensive Care Unit}}
  ({{ICU}}) {{Settings Using MIMIC Dataset}}: {{Systematic Review}}.
\newblock \emph{Informatics}, 8\penalty0 (1):\penalty0 16, March 2021.
\newblock ISSN 2227-9709.
\newblock \doi{10.3390/informatics8010016}.

\bibitem[Tang et~al.(2020)Tang, Davarmanesh, Song, Koutra, Sjoding, and
  Wiens]{tangDemocratizingEHRAnalyses2020}
Shengpu Tang, Parmida Davarmanesh, Yanmeng Song, Danai Koutra, Michael~W
  Sjoding, and Jenna Wiens.
\newblock Democratizing {{EHR}} analyses with {{FIDDLE}}: A flexible
  data-driven preprocessing pipeline for structured clinical data.
\newblock \emph{Journal of the American Medical Informatics Association},
  27\penalty0 (12):\penalty0 1921--1934, December 2020.
\newblock ISSN 1527-974X.
\newblock \doi{10.1093/jamia/ocaa139}.

\bibitem[Tashiro et~al.(2021)Tashiro, Song, Song, and
  Ermon]{tashiroCSDIConditionalScorebased2021a}
Yusuke Tashiro, Jiaming Song, Yang Song, and Stefano Ermon.
\newblock {{CSDI}}: {{Conditional Score-based Diffusion Models}} for
  {{Probabilistic Time Series Imputation}}.
\newblock In \emph{Advances in {{Neural Information Processing Systems}}},
  volume~34, pp.\  24804--24816. Curran Associates, Inc., 2021.

\bibitem[Thoral et~al.(2021)Thoral, Peppink, Driessen, Sijbrands, Kompanje,
  Kaplan, Bailey, Kesecioglu, Cecconi, Churpek, Clermont, {van der Schaar},
  Ercole, Girbes, and Elbers]{thoralSharingICUPatient2021}
Patrick~J. Thoral, Jan~M. Peppink, Ronald~H. Driessen, Eric J.~G. Sijbrands,
  Erwin J.~O. Kompanje, Lewis Kaplan, Heatherlee Bailey, Jozef Kesecioglu,
  Maurizio Cecconi, Matthew Churpek, Gilles Clermont, Mihaela {van der Schaar},
  Ari Ercole, Armand R.~J. Girbes, and Paul W.~G. Elbers.
\newblock Sharing {{ICU Patient Data Responsibly Under}} the {{Society}} of
  {{Critical Care Medicine}}/{{European Society}} of {{Intensive Care Medicine
  Joint Data Science Collaboration}}: {{The Amsterdam University Medical
  Centers Database}} ({{AmsterdamUMCdb}}) {{Example}}*.
\newblock \emph{Critical Care Medicine}, 49\penalty0 (6):\penalty0 e563--e577,
  June 2021.
\newblock ISSN 0090-3493.
\newblock \doi{10.1097/CCM.0000000000004916}.

\bibitem[Toma{\v s}ev et~al.(2019)Toma{\v s}ev, Glorot, Rae, Zielinski, Askham,
  Saraiva, Mottram, Meyer, Ravuri, Protsyuk, Connell, Hughes, Karthikesalingam,
  Cornebise, Montgomery, Rees, Laing, Baker, Peterson, Reeves, Hassabis, King,
  Suleyman, Back, Nielson, Ledsam, and
  Mohamed]{tomasevClinicallyApplicableApproach2019}
Nenad Toma{\v s}ev, Xavier Glorot, Jack~W. Rae, Michal Zielinski, Harry Askham,
  Andre Saraiva, Anne Mottram, Clemens Meyer, Suman Ravuri, Ivan Protsyuk,
  Alistair Connell, C{\'i}an~O. Hughes, Alan Karthikesalingam, Julien
  Cornebise, Hugh Montgomery, Geraint Rees, Chris Laing, Clifton~R. Baker,
  Kelly Peterson, Ruth Reeves, Demis Hassabis, Dominic King, Mustafa Suleyman,
  Trevor Back, Christopher Nielson, Joseph~R. Ledsam, and Shakir Mohamed.
\newblock A clinically applicable approach to continuous prediction of future
  acute kidney injury.
\newblock \emph{Nature}, 572\penalty0 (7767):\penalty0 116--119, August 2019.
\newblock ISSN 1476-4687.
\newblock \doi{10.1038/s41586-019-1390-1}.

\bibitem[{van de Water} \& Arnrich(2023){van de Water} and
  Arnrich]{vandewaterClosingGapsImputation2023}
Robin {van de Water} and Bert Arnrich.
\newblock Closing {{Gaps}}: {{An Imputation Analysis}} of {{ICU Vital Signs}}.
\newblock In \emph{Deep {{Generative Models}} for {{Health Workshop NeurIPS}}
  2023}, October 2023.

\bibitem[Vaswani et~al.(2017)Vaswani, Shazeer, Parmar, Uszkoreit, Jones, Gomez,
  Kaiser, and Polosukhin]{vaswaniAttentionAllYou2017}
Ashish Vaswani, Noam Shazeer, Niki Parmar, Jakob Uszkoreit, Llion Jones,
  Aidan~N. Gomez, Lukasz Kaiser, and Illia Polosukhin.
\newblock Attention {{Is All You Need}}.
\newblock \emph{arXiv:1706.03762 [cs]}, December 2017.

\bibitem[Virtanen et~al.(2020)Virtanen, Gommers, Oliphant, Haberland, Reddy,
  Cournapeau, Burovski, Peterson, Weckesser, Bright, {van der Walt}, Brett,
  Wilson, Millman, Mayorov, Nelson, Jones, Kern, Larson, Carey, Polat, Feng,
  Moore, VanderPlas, Laxalde, Perktold, Cimrman, Henriksen, Quintero, Harris,
  Archibald, Ribeiro, Pedregosa, {van Mulbregt}, and {SciPy 1.0
  Contributors}]{2020SciPy-NMeth}
Pauli Virtanen, Ralf Gommers, Travis~E. Oliphant, Matt Haberland, Tyler Reddy,
  David Cournapeau, Evgeni Burovski, Pearu Peterson, Warren Weckesser, Jonathan
  Bright, St{\'e}fan~J. {van der Walt}, Matthew Brett, Joshua Wilson, K.~Jarrod
  Millman, Nikolay Mayorov, Andrew R.~J. Nelson, Eric Jones, Robert Kern, Eric
  Larson, C~J Carey, {\.I}lhan Polat, Yu~Feng, Eric~W. Moore, Jake VanderPlas,
  Denis Laxalde, Josef Perktold, Robert Cimrman, Ian Henriksen, E.~A. Quintero,
  Charles~R. Harris, Anne~M. Archibald, Ant{\^o}nio~H. Ribeiro, Fabian
  Pedregosa, Paul {van Mulbregt}, and {SciPy 1.0 Contributors}.
\newblock {{SciPy}} 1.0: {{Fundamental}} algorithms for scientific computing in
  python.
\newblock \emph{Nature Methods}, 17:\penalty0 261--272, 2020.
\newblock \doi{10.1038/s41592-019-0686-2}.

\bibitem[Vohra(2016)]{vohraApacheParquet2016}
Deepak Vohra.
\newblock Apache {{Parquet}}.
\newblock In Deepak Vohra (ed.), \emph{Practical {{Hadoop Ecosystem}}: {{A
  Definitive Guide}} to {{Hadoop-Related Frameworks}} and {{Tools}}}, pp.\
  325--335. Apress, Berkeley, CA, 2016.
\newblock ISBN 978-1-4842-2199-0.
\newblock \doi{10.1007/978-1-4842-2199-0_8}.

\bibitem[Wang et~al.(2020{\natexlab{a}})Wang, Deng, Zhang, and
  Zhao]{wangRealTimePredictionAKI2020}
Hui Wang, Fuxing Deng, Buyao Zhang, and Shuangping Zhao.
\newblock Real-{{Time Prediction}} of {{AKI Among Middle-Aged}} and {{Older}}
  in {{ICU}}: {{A Retrospective}} and {{Machine Learning Study}}.
\newblock Preprint, In Review, August 2020{\natexlab{a}}.

\bibitem[Wang et~al.(2020{\natexlab{b}})Wang, McDermott, Chauhan, Ghassemi,
  Hughes, and Naumann]{wangMIMICExtractDataExtraction2020}
Shirly Wang, Matthew B.~A. McDermott, Geeticka Chauhan, Marzyeh Ghassemi,
  Michael~C. Hughes, and Tristan Naumann.
\newblock {{MIMIC-Extract}}: A data extraction, preprocessing, and
  representation pipeline for {{MIMIC-III}}.
\newblock In \emph{Proceedings of the {{ACM Conference}} on {{Health}},
  {{Inference}}, and {{Learning}}}, pp.\  222--235, Toronto Ontario Canada,
  April 2020{\natexlab{b}}. ACM.
\newblock ISBN 978-1-4503-7046-2.
\newblock \doi{10.1145/3368555.3384469}.

\bibitem[Wang et~al.(2022)Wang, Zhao, Callcut, and
  Petzold]{wangIntegratingPhysiologicalTime2022}
Yuqing Wang, Yun Zhao, Rachael Callcut, and Linda Petzold.
\newblock Integrating {{Physiological Time Series}} and {{Clinical Notes}} with
  {{Transformer}} for {{Early Prediction}} of {{Sepsis}}, March 2022.

\bibitem[Yang et~al.(2023)Yang, Wu, Jiang, Lin, Gao, Danek, and
  Sun]{yangPyHealthDeepLearning2023}
Chaoqi Yang, Zhenbang Wu, Patrick Jiang, Zhen Lin, Junyi Gao, Benjamin~P.
  Danek, and Jimeng Sun.
\newblock {{PyHealth}}: {{A Deep Learning Toolkit}} for {{Healthcare
  Applications}}.
\newblock In \emph{Proceedings of the 29th {{ACM SIGKDD Conference}} on
  {{Knowledge Discovery}} and {{Data Mining}}}, {{KDD}} '23, pp.\  5788--5789,
  New York, NY, USA, August 2023. Association for Computing Machinery.
\newblock ISBN 9798400701030.
\newblock \doi{10.1145/3580305.3599178}.

\bibitem[Y{\`e}che et~al.(2022)Y{\`e}che, Kuznetsova, Zimmermann, H{\"u}ser,
  Lyu, Faltys, and R{\"a}tsch]{yecheHiRIDICUBenchmarkComprehensiveMachine2022}
Hugo Y{\`e}che, Rita Kuznetsova, Marc Zimmermann, Matthias H{\"u}ser, Xinrui
  Lyu, Martin Faltys, and Gunnar R{\"a}tsch.
\newblock {{HiRID-ICU-Benchmark}} -- {{A Comprehensive Machine Learning
  Benchmark}} on {{High-resolution ICU Data}}.
\newblock \emph{arXiv:2111.08536 [cs]}, January 2022.

\bibitem[Yoo et~al.(2003)Yoo, Jette, and Grondona]{yooSLURMSimpleLinux2003}
Andy~B. Yoo, Morris~A. Jette, and Mark Grondona.
\newblock {{SLURM}}: {{Simple Linux Utility}} for {{Resource Management}}.
\newblock In Dror Feitelson, Larry Rudolph, and Uwe Schwiegelshohn (eds.),
  \emph{Job {{Scheduling Strategies}} for {{Parallel Processing}}}, Lecture
  {{Notes}} in {{Computer Science}}, pp.\  44--60, Berlin, Heidelberg, 2003.
  Springer.
\newblock ISBN 978-3-540-39727-4.
\newblock \doi{10.1007/10968987_3}.

\end{thebibliography}
\bibliographystyle{iclr2024_conference}
\appendix

\FloatBarrier
\newpage
{\large\textsc{Appendices: table of contents}}
\textsc{
\begin{itemize}
    \item \autoref{app:yaib_context}: YAIB's contribution in context
    \item \autoref{app:ext-results}: Extended results
    \item \autoref{appendix:datasets}: Datasets
    \item \autoref{appendix:outc-detail}: Outcome definitions
    \item \autoref{app:usage}: YAIB's usage and implementation
    \item \autoref{app:extend}: Extending YAIB
    \item \autoref{app:exp_setup}: Experimental setup and reproducibility
    \item \autoref{app:hyperparams}: Hyperparameters
    \item \autoref{app:checklist}: Machine Learning Reproducibility Checklist
\end{itemize}}
\section{Appendix: YAIB's contribution in context}
\label{app:yaib_context}
This Appendix provides an extensive description for the positioning of \acrshort{yaib} in the contemporary clinical ML research landscape. We particularly recommend it to those that are looking into creating their own solutions for clinical ML.
\subsection{Extensibility and reproducibility}
We designed \acrshort{yaib} to be as extensible as possible while retaining full reproducibility. This means easy support of new databases, clinical concepts, tasks, experiment configurations, preprocessing pipelines, imputation methods, models, and evaluation metrics. If changes are necessary, they need to be reproducible and easily shareable across research teams. 
If the user only requires a few default ICU tasks from a single i.i.d. dataset to test their new method, any existing ICU benchmarks could be sufficient. Users do not need to apply for access to multiple datasets and do not have to deal with the intricacies of the clinical task definition. As long as the integration of a new model is seamless, such simple frameworks are fit-for-purpose and abstract much of the complexity, allowing the user to only worry about one thing: their model. If multiple papers used the exact same benchmark, results are also directly comparable between papers (an “apples-to-apples” comparison).

However, we found that this setup tends to be too restrictive and thus unrealistic. Users often want to highlight a particular aspect of their model, prompting them to adapt to the default task. At other times, they want to show clinical impact and need to adapt the default task to make it more realistic. Given the lack of successful translation of prediction models into clinical practice, reviewers are also increasingly requesting external validation – sometimes with multiple endpoints – which is difficult to shoehorn into most existing solutions.
\acrshort{yaib} embraces the need to tweak experimental setups. Results will no longer be directly comparable between papers, but we argue that true apples-to-apples comparisons were inherently rare. Instead of forcing users into a rigid framework, it allows for adaptations but requires them to be done in a transparent manner. Absolute performance should be compared only within the same paper or among papers with the same task setup (see our examples in Tables 5 and 6).

To facilitate the transparency of adaptations, we rely on a sophisticated framework to define clinical concepts across multiple datasets (ricu). We have adapted and extended ricu to provide a standard workflow for \acrshort{yaib} to integrate new databases and define new clinical concepts. To date, it has been successfully used to bring 4/5 +1 ICU datasets into a common format (including our addition of the Salzburg Intensive Care Database, which is currently in quality control). This approach is flexible enough that we have not yet encountered significant restrictions in mapping admissions, demographics, vital signs, laboratory values, medication (including rates and durations), clinical scores, and outcomes at different time scales across datasets. The main restriction of ricu is that it is currently implemented in the R language only, but we provide guidance on how to access it via rpy2, and we are in the process of porting it to Python; this will make our pipeline even more accessible, especially to clinical researchers. Our cohort definition functionality provides helper functions to apply inclusion/exclusion criteria on top of ricu and report step-by-step attrition numbers. The cohorts can be used in a modular fashion with custom preprocessing steps, imputations, prediction models, and evaluation metrics, all using the exact same code across multiple datasets.

Even so, there will likely be situations where the user may be better off with a custom solution. We expect this to occur once their use case diverges significantly from standard supervised learning. For example, federated learning or reinforcement learning setups may require significantly different training and evaluation loops. These are not currently supported, but we consider this as future work. In any case, the user can still use our data processing, cohort generation, and possibly other parts of \acrshort{yaib} (e.g., by exchanging the default training module with a custom module). Authors using \acrshort{yaib} should, therefore, provide their code and a detailed list of the changes they have made to the repository; modern version control allows us to verify this against the original \acrshort{yaib} repository easily. 

The \acrshort{yaib} pipeline has helped us to produce reproducible results quickly and provides the required extensibility for our purposes. We are in touch with some researchers who have used \acrshort{yaib} to date and provided feedback, although mainly in an informal way. We refer to \citetalias{vandewaterClosingGapsImputation2023} as an example of the usability of \acrshort{yaib}. This work used YAIB as a bedrock for implementing imputation methods and are in the process of extending this to more methods and downstream tasks. For concrete examples and guidance for how to extend \acrshort{yaib} , we refer to Appendix D and the wiki documentation. 
\subsection{The choice of features}
We chose the 52 most common clinical features shared by all datasets. They were readily available in all benchmarked datasets, demonstrating \acrshort{yaib}'s adaptability. This is done because our work focuses on the interoperability of datasets and the opportunity for experiments with a.o. transfer learning and domain adaption. We believe there is the most value in providing a modular setup where the user can add or remove features to suit their needs better and, most importantly, do so reproducibly. 

Nevertheless, several medications for eICU and MIMIC-IV are readily available; the ricu package maintains a full list of the currently available native concepts which are available\footnote{\url{https://github.com/eth-mds/ricu/blob/main/inst/extdata/config/concept-dict.json}}. Complex concepts, dependent on several native concepts, such as SOFA scores, are additionally available. Each concept that is available in ricu can be readily used in \acrshort{yaib}.  Some medications that are already implemented, such as antibiotics and vasopressors, are used in the definition of the complex Sepsis endpoint. Therefore, we decided to leave those out to have the same features for each task. 

We note, additionally, that it is straightforward to implement new concepts in our pipeline; Appendix E.2 describes the addition of Potassium Chloride to the ICU harmonization package ricu. A similar process can be followed for adding new medications, which immediately improves the usability of \acrshort{yaib}. Moreover, we are actively working on integrating more features, including comorbidities and medications. We would like to note that many features are not available across all datasets; this does not mean they can not be valuable in clinical prediction tasks. 

Finally, we would like to point out that \acrshort{yaib}’s end-to-end pipeline is designed as a solid starting point for 1) clinicians looking for external validation to employ ML in practice, 2) dataset creators looking for a solid platform to facilitate widespread use, and 3) the ML community to contribute novel prediction models. They can use a mature and externally developed framework, which adds to the credibility of any experiment results. Adding new feature concepts for their datasets can also increase the adoption of their datasets. They are likely domain experts for their respective datasets, meaning fewer errors are made in this process. This process will improve the usability of \acrshort{yaib} as an end-to-end benchmarking tool and improve the confidence of health experts in clinical ML.
\subsection{Using YAIB in novel scientific work}
We acknowledge the importance of reproducible ML experiments. In this section, we describe how future work can transparently use \acrshort{yaib} as a platform for comparing their contributions. The authors ideally provide one or more open GitHub repositories so it is straightforward to check versioning; this includes:
\begin{enumerate}
    \item The concept dictionary in JSON format if they add new concepts. The main repository contains the current version of the concept dictionary of the vanilla \texttt{ricu}\footnote{\url{https://github.com/eth-mds/ricu/blob/main/inst/extdata/config/concept-dict.json}}.
    \item The repository that is used to generate cohorts if they introduce a new task. Ideally, this is forked from the \acrshort{yaib}-cohorts repository.
    \item The \texttt{preprocessing.py} file in case this has been changed.
    \item The \texttt{model.py} and \texttt{dataset.py} file that contains the definition for the model and dataset and dataloader (if adjusted).
    \item \texttt{model.gin} file that specificies the used hyperparameters and hyperparameter ranges.
    \item \texttt{wandb.yml} if Weights and Biases is used for running experiments with this model.
    \item Provide versions of \texttt{ricu}, \texttt{YAIB-cohorts}, and \texttt{YAIB} they have used as a base.
\end{enumerate}
If authors cover these aspects when presenting new work; one can easily reproduce their experiments even though they might not have used a "vanilla" implementation of YAIB. An additional benefit of providing these materials is that authors of future work can hereby participate in making YAIB more comprehensive.
\subsection{Extended related work}
Comparison to existing frameworks
We thank the reviewer for bringing up the preprint of TemporAI, which is still in early development at the time of writing. While we included an earlier work by the same group, Clairvoyance, in our related work, we have now updated the manuscript by adding this work in the related work section and to Table 1. We note that Pyhealth is already included in the related work section of the original manuscript. However, we elaborate on the differences between \acrshort{yaib} and both works below.
\subsubsection{Clairvoyance}
Clairvoyance~\citep{jarrettCLAIRVOYANCEPIPELINETOOLKIT2021} is "a Unified, End-to-End AutoML Pipeline for Medical Time Series". As such, it does not focus on ICUs or benchmarking but instead standardizes model learning (imputation and training), model evaluation, and model selection, focusing on the computational aspects of developing a model. Clairvoyance comes with some code to define a task for treatment effects estimation on MIMIC III data. However, this task is hard coded and lightly documented, primarily serving as a demo of Clairvoyance. The exemplary nature of this task is further exemplified by the fact that, at no point the authors mention possible confounders/colliders of the treatment effect and whether they are conceivably adjusted for by the covariates, rendering any causal interpretation moot. It is unclear how this task can be easily adapted or extended to other databases without significant amounts of custom code.

\textbf{Advantages of \acrshort{yaib} compared to Clairvoyance:} \acrshort{yaib} puts ICU data and tasks front and center. \acrshort{yaib} supports the whole workflow, from raw data to clinical concepts to well-defined cohorts. This approach greatly facilitates the transparent and reproducible preprocessing of (often messy) ICU data, which Clairvoyance does not cover. We strongly believe that unless tasks can be adapted easily and reproducibly, it will lead to inevitable ad-hoc adaptations of the task that often end up irreproducible. \acrshort{yaib}, therefore, improves on existing modeling frameworks by putting an equal emphasis on standardized data processing for meaningful model development. 

\subsubsection{TemporAI}
TemporAI~\citep{savelievTemporAIFacilitatingMachine2023} is a package that is currently in early development without a peer-reviewed publication associated with it. While it promises to provide: "prediction, causal inference, and time-to-event analysis, as well as common preprocessing utilities and model interpretability methods," it is unclear from current documentation how to use established datasets with this package or how to use relevant medical prediction tasks. 

\textbf{The advantages of \acrshort{yaib} compared to TemporAI} are similar to those between \acrshort{yaib} and Clairvoyance: \acrshort{yaib} puts ICU data and tasks front and center for both ML scientists and clinicians. \acrshort{yaib} supports the whole workflow, from raw data to clinical concepts to well-defined cohorts. This approach greatly facilitates the transparent and reproducible preprocessing of (often messy) ICU data, which TemporAI, similarly to Clairvoyance, does not cover. However, we would like to note that using TemporAI (or Clairvoyance) with the \acrshort{yaib} pipeline to create a different end-to-end pipeline is possible as it allows for "swapping out" components. We provide the functionality in our \texttt{\acrshort{yaib}-cohorts} repository to convert any cohort to a format compatible with Clairvoyance and TemporAI.

\subsubsection{PyHealth}
PyHealth~\citep{yangPyHealthDeepLearning2023} is "a comprehensive deep learning toolkit designed for both ML researchers and healthcare practitioners." PyHealth aims to support all EHR databases. It is thus similar in scope to our proposed framework. Unfortunately, upon closer inspection, PyHealth only supports a small subset of the information in MIMIC and eICU. While diagnoses and prescriptions are, in theory, included, they are processed as a simple bag of diagnosis codes or drug codes without information on strength/duration or semantic interpretation of what they represent (e.g., what is a vasopressor needed in calculating the SOFA score). Vital signs are not supported at all, presumably because PyHealth reads information from raw .csv files and may struggle to process large quantities of vital sign data. PyHealth further states that the datasets are independent of task definitions. This, unfortunately, appears to mean that they have to be implemented anew for each database, with custom dataset-specific code for the same task. Furthermore, all currently available ICU tasks in PyHealth use static data only and do not include any time series.

\textbf{Advantages of \acrshort{yaib} compared to PyHealth:} \acrshort{yaib} supports all databases within a common, principled interface (see the response on data harmonization above). Moreover, \acrshort{yaib} enables a single task definition that one can directly use for any included dataset. As far as they can work with time series data, \acrshort{yaib} can incorporate any model defined in PyHealth.

\section{Appendix: Extended Results}
This Appendix contains results that were left out of the main text.
\label{app:ext-results}
\begin{table}[ht]
\small
\centering
    \caption{\textit{Comparing the use of dynamic feature generation (FG) to the baseline of ICU mortality prediction, \acrshort{auroc} ($\uparrow$).} Note that an otherwise identical experiment setup was used to obtain results for the ``without feature generation'' results.}
    \begin{threeparttable}
    \begin{adjustbox}{center}
        \begin{tabularx}{\textwidth}{lc>{\color{darkgray!60}}cc>{\color{darkgray!60}}cc>{\color{darkgray!60}}cc>{\color{darkgray!60}}c}
            \toprule
            \addlinespace[0.2em]
             & \multicolumn{2}{c}{\textbf{\acrshort{aumc}}} & \multicolumn{2}{c}{\textbf{\acrshort{hirid}}} & \multicolumn{2}{c}{\textbf{\acrshort{eicu}}} & \multicolumn{2}{c}{\textbf{\acrshort{miiv}}}\\
            \addlinespace[0.1em]
             \cline{2-9}
             \addlinespace[0.1em]
                \textbf{Preprocessing}& w/ FG & w/o FG &  w/ FG & w/o FG &  w/ FG & w/o FG &  w/ FG & w/o FG \\
            \addlinespace[0.1em]
            \cmidrule{1-9}\morecmidrules\cmidrule{1-9} 
            LR & 83.7±0.6 & 82.2±0.5 & 84.0±0.3 & 81.8±0.7 & 84.8±0.2 & 81.1±0.1 & 86.1±0.1 & 80.2±0.3 \\
            LGBM & \textbf{84.5±0.6} & \textbf{83.5±0.4} & \textbf{84.5±0.3} & \textbf{82.5±0.6} & \textbf{85.7±0.2} & \textbf{83.5±0.3} & \textbf{87.7±0.2} & \textbf{85.9±0.1} \\
            \bottomrule
        \end{tabularx}
        \end{adjustbox}
        \begin{tablenotes}[flushleft, para]
            \footnotesize
        \end{tablenotes}
    \end{threeparttable}
    \label{tab:without_dynamic}
\end{table}

\begin{table}[h]
\small
    \caption{\textit{Comparing the use of dynamic feature generation (FG) to the baseline of ICU mortality prediction, \acrshort{auprc} ($\uparrow$).} Note that an otherwise identical experiment setup was used to obtain results for the ``without feature generation'' results.}
    \centering
    \begin{threeparttable}
    \begin{adjustbox}{center}
        \begin{tabularx}{\textwidth}{lc>{\color{darkgray!60}}cc>{\color{darkgray!60}}cc>{\color{darkgray!60}}cc>{\color{darkgray!60}}c}
            \toprule
            \addlinespace[0.2em]
             & \multicolumn{2}{c}{\textbf{\acrshort{aumc}}} & \multicolumn{2}{c}{\textbf{\acrshort{hirid}}} & \multicolumn{2}{c}{\textbf{\acrshort{eicu}}} & \multicolumn{2}{c}{\textbf{\acrshort{miiv}}}\\
            \addlinespace[0.1em]
             \cline{2-9}
             \addlinespace[0.1em]
                \textbf{Preprocessing}& w/ FG & w/o FG &  w/ FG & w/o FG &  w/ FG & w/o FG &  w/ FG & w/o FG \\
            \addlinespace[0.1em]
            \cmidrule{1-9}\morecmidrules\cmidrule{1-9} 
            LR & \textbf{52.9±1.2} & \textbf{50.8±1.2} & 36.9±1.1 & 33.1±0.8& 33.0±0.7 & 28.3±0.4 & 39.7±0.6 &32.1±0.6\\
            LGBM & \textbf{53.7±1.2} & \textbf{50.9±1.0} & \textbf{40.6±0.8} & \textbf{35.3±0.9} & \textbf{36.0±0.6} & \textbf{32.5±0.9} & \textbf{44.2±0.7} & \textbf{40.1±0.7}\\
            \bottomrule
        \end{tabularx}
        \end{adjustbox}
        \begin{tablenotes}[flushleft, para]
            \footnotesize
        \end{tablenotes}
    \end{threeparttable}
        \label{tab:without_dynamic_auprc}
\end{table}
\paragraph{Feature generation} We compare the use of feature generation for classical \acrshort{ml} models. The \preproc{} package provides the functionality of assembling different preprocessing steps to be supplied by the user (i.e., a recipe). We show the results in \autoref{tab:without_dynamic} and \ref{tab:without_dynamic_auprc}.
\setlength{\tabcolsep}{4pt}
\begin{table}[h]
\centering
    \caption{\textit{\acrshort{auroc} ($\uparrow$) performance comparison of including static data for ICU mortality prediction.}}
    \small
    \label{tab:without_static_auroc}
    \begin{threeparttable}
        \begin{tabularx}{\textwidth}{lX>{\leavevmode\color{darkgray!60}}XX>{\leavevmode\color{darkgray!60}}XX>{\leavevmode\color{darkgray!60}}XX>{\leavevmode\color{darkgray!60}}X}
            \toprule
            \addlinespace[0.2em]
             & \multicolumn{2}{c}{\textbf{\acrshort{aumc}}} & \multicolumn{2}{c}{\textbf{\acrshort{hirid}}} & \multicolumn{2}{c}{\textbf{\acrshort{eicu}}} & \multicolumn{2}{c}{\textbf{\acrshort{miiv}}}\\
            \addlinespace[0.1em]
             \cline{2-9}
            \addlinespace[0.2em]
                \textbf{Inclusion} & w/ static & w/o static & w/ static & w/o static & w/ static & w/o static & w/ static & w/o static \\
            \cmidrule{1-9}\morecmidrules\cmidrule{1-9} 
            LR & 83.7±0.6 & 82.9±0.6& 84.0±0.3 & 82.8±0.7 & 84.8±0.2 & 84.3±0.2 & 86.1±0.1 & 84.3±0.3 \\ 
            LGBM & \textbf{84.5±0.6} & \textbf{83.4±0.4} & \textbf{84.4±0.3} & \textbf{83.9±0.7} & \textbf{85.7±0.2} & 84.7±0.2 & 84.7±0.2 & 86.2±0.2 \\
                \addlinespace[0.5em]
            GRU & 83.7±0.7 & \textbf{83.4±0.6} & \textbf{84.3±0.7} & \textbf{84.0±0.8} & \textbf{85.9±0.2} & 85.7±0.2 & \textbf{87.4±0.2} & \textbf{86.9±0.3} \\
            LSTM & 83.7±0.7 & 82.9±0.6 & 84.0±0.7& \textbf{83.4±0.7} & 85.5±0.2 & 85.1±0.2 & 86.7±0.4 & 86.1±0.3 \\
            TCN & \textbf{84.0±0.6} & \textbf{83.5±0.6} & \textbf{84.6±0.7} & \textbf{83.9±0.8} & 85.4±0.2 & \textbf{86.4±0.3} & 87.1±0.3 & 86.4±0.3 \\
            TF & \textbf{84.1±0.2} & \textbf{83.7±0.4} & \textbf{84.9±0.7} & \textbf{84.4±0.7} & \textbf{85.9±0.2} & 85.7±0.2 & 86.9±0.3 & 86.5±0.3 \\
            \bottomrule
        \end{tabularx}
    \end{threeparttable}
\end{table}
\setlength{\tabcolsep}{4pt}
\begin{table}[h]
    \small
        \caption{\textit{\acrshort{auprc} ($\uparrow$) performance comparison of including static data for ICU Mortality Prediction.}}
        \centering
    \label{tab:without_static_auprc}
    \begin{threeparttable}
        \begin{tabularx}{\textwidth}{lc>{\color{darkgray!60}}cc>{\color{darkgray!60}}cc>{\color{darkgray!60}}cc>{\color{darkgray!60}}c}
        \toprule
            \addlinespace[0.2em]
             & \multicolumn{2}{c}{\textbf{\acrshort{aumc}}} & \multicolumn{2}{c}{\textbf{\acrshort{hirid}}} & \multicolumn{2}{c}{\textbf{\acrshort{eicu}}} & \multicolumn{2}{c}{\textbf{\acrshort{miiv}}}\\
            \addlinespace[0.1em]
             \cline{2-9}
            \addlinespace[0.2em]
                \textbf{Inclusion} & w/ static & w/o static & w/ static & w/o static & w/ static & w/o static & w/ static & w/o static \\
            \cmidrule{1-9}\morecmidrules\cmidrule{1-9}            
            LR & 52.9±1.2 & \textbf{51.7±1.2} & 36.9±1.1 & 34.0±1.1 & 33.0±0.7 & 32.2±0.6 & 39.7±0.6 & 36.9±0.6\\
            LGBM & \textbf{53.7±1.2} & 51.1±1.1 & \textbf{40.6±0.8} & \textbf{38.9±1.5} & \textbf{36.0±0.6} & 34.1±0.7 & \textbf{44.2±0.7} & \textbf{41.0±0.7}\\
                \addlinespace[0.3em]
            GRU & 53.1±1.5 & \textbf{52.9±1.2} & 37.6±1.2 & 37.3±1.1 & \textbf{36.1±0.9} & \textbf{35.3±0.8} & 42.4±0.6 & \textbf{41.5±0.7} \\
            LSTM & \textbf{53.6±1.4} & 51.0±1.0 & 37.8±1.0 & 36.2±1.4 & 35.7±0.8 & \textbf{34.6±0.}7 & 41.0±0.7 & 40.2±0.8\\
            TCN  & \textbf{54.2±1.4} & \textbf{52.7±1.0} & 39.2±1.3 & \textbf{37.5±1.5} & 34.3±0.6 & \textbf{35.4±0.8} & 41.4±0.8 & \textbf{40.8±0.7}\\
            TF & \textbf{54.4±1.1} & \textbf{52.7±0.9} & 39.3±1.5 & \textbf{38.4±1.5} & 34.7±0.8 & 34.5±0.7 & 42.2±0.3 & \textbf{41.3±0.8}\\
            \bottomrule 
        \end{tabularx}
    \end{threeparttable}
\end{table}



\begin{figure}
    \centering
    \includegraphics[width=0.5\textwidth]{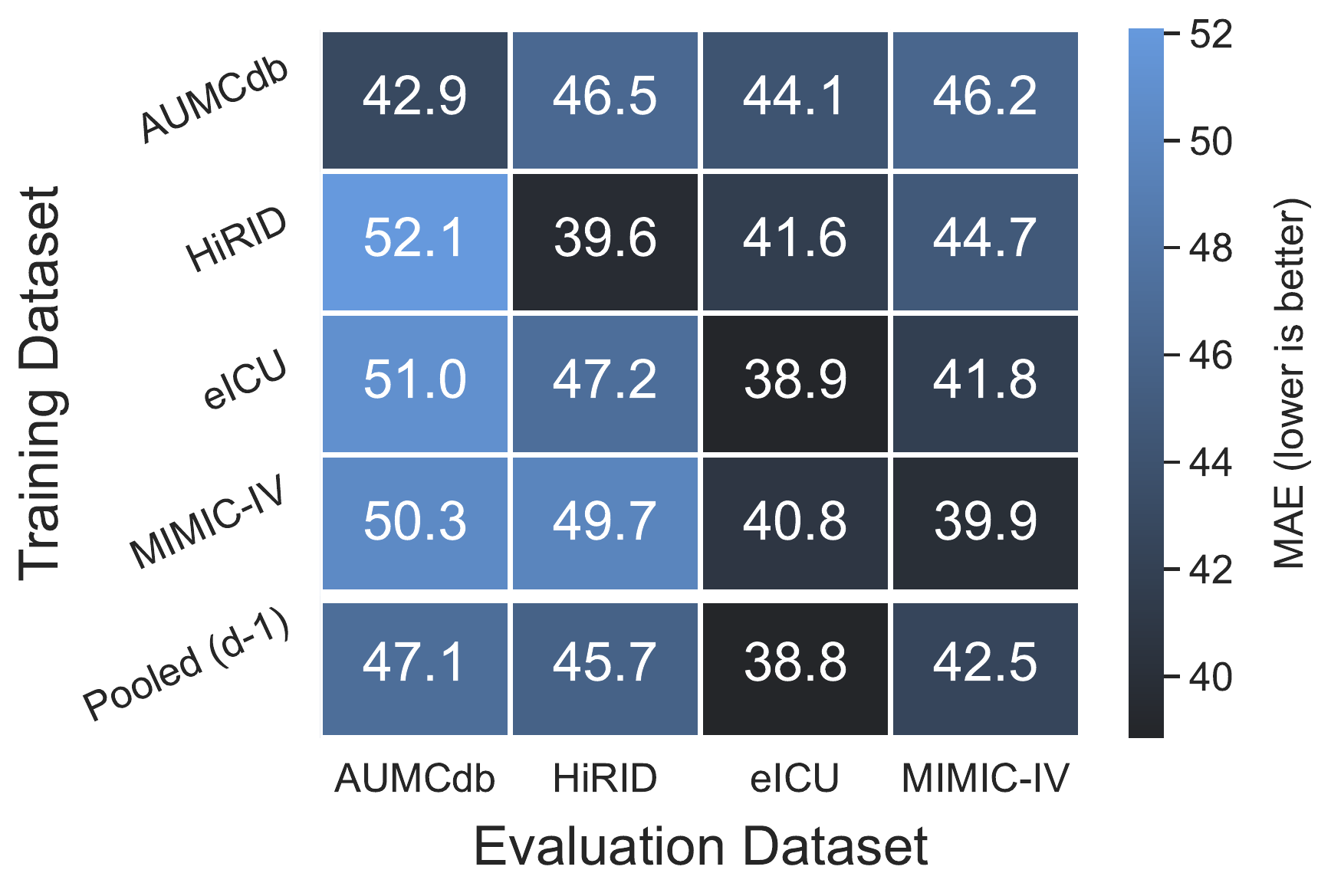}
    \caption{\textit{Performance in MAE of the Transformer model on \textit{\acrfull{los}}.}}
    \label{fig:ext_val_los}
\end{figure}

\paragraph{The impact of static features}
We leveraged this customizable preprocessing to perform training prediction models without static features (i.e., age, sex, height, and weight). The AUROC results are found in \autoref{tab:without_static_auroc} (the AUPRC in \autoref{tab:without_static_auprc}).
From these results, we can see that including the static data seems to result in better performance across all models and datasets. 
\begin{table}[h]
\small
        \caption{\textit{Baseline AUROC ($\uparrow$), AUPRC ($\uparrow$) performance of the included ML algorithms on the demo cohorts.}}
    \begin{threeparttable}
        \begin{tabularx}{\textwidth}{l>{\centering\arraybackslash}X>{\centering\arraybackslash\leavevmode\color{darkgray!60}}X>{\centering\arraybackslash}X>{\centering\arraybackslash\leavevmode\color{darkgray!60}}X} 
            \toprule
                         \addlinespace[0.2em]
            & \multicolumn{2}{c}{\textbf{\acrshort{eicu} Demo}} & \multicolumn{2}{c}{\textbf{\acrshort{miiii} Demo}}\\
            \addlinespace[0.1em]
             \cline{2-5}
                \addlinespace[0.1em]
                \textbf{Algorithm} & AUROC & AUPRC & AUROC & AUPRC \\
                \cmidrule{1-5}\morecmidrules\cmidrule{1-5}             
            \textbf{ICU Mortality}\\
            \addlinespace[0.2em]
            LR & 67.7±0.9 & 15.4±1.3 & 52.9±3.2 & \textbf{37.6±2.3} \\
            LGBM & \textbf{72.9±1.1} & \textbf{21.4±1.7} & \textbf{59.0±2.1} & 33.0±2.4 \\
                \addlinespace[0.3em]
            GRU & 71.5±1.3 & \textbf{20.8±1.5} & 50.5±3.6 & 33.9±3.5 \\
            LSTM & \textbf{71.3±1.6} & 18.9±1.9 & 55.5±2.3 & 34.2±3.3 \\
            TCN & 69.2±1.7 & 18.0±1.5 & 55.1±4.0 & \textbf{38.8±3.6} \\
            TF & 71.6±1.3 & 18.7±1.7 & 53.3±3.5 & \textbf{36.8±3.1} \\
            \cmidrule{1-5}\morecmidrules\cmidrule{1-5}            
            \textbf{AKI}\\
            LR & 61.3±0.6 & 16.8±0.4 & 52.9±2.3 & 16.8±2.0 \\
            LGBM & \textbf{72.6±0.3} & \textbf{23.8±0.4} & \textbf{60.7±1.7} & \textbf{22.4±1.8} \\
                \addlinespace[0.3em]
            GRU & 63.0±0.8 & 17.3±0.6 & 50.5±3.3 & 17.6±1.5 \\
            LSTM & 61.9±0.9 & 16.2±0.6 & 53.5±2.5 & 15.4±1.3 \\
            TCN & 64.5±1.0 & 17.6±0.6 & 53.7±2.8 & 19.6±2.0 \\
            TF & 70.1±0.5 & 21.7±0.7 & 53.9±2.0 & 15.3±1.0 \\
            \cmidrule{1-5}\morecmidrules\cmidrule{1-5}          
            \textbf{Sepsis}\\
            LR & 63.8±0.9 & 3.7±0.3 & * & * \\
            LGBM & 53.5±1.5 & 2.8±0.2 & * & * \\
                \addlinespace[0.3em]
            GRU & 64.7±1.3 & 4.1±0.3 & * & * \\
            LSTM & 65.3±1.3 & 4.3±0.5 & * & * \\
            TCN & 66.4±1.1 & 4.2±0.3 & * & * \\
            TF & \textbf{68.4±1.1} & \textbf{5.8±0.6} & * & * \\
            \bottomrule
        \end{tabularx}
        \begin{tablenotes}[flushleft, para]
            \footnotesize 
            \item * Our sepsis definition resulted in just one sepsis case for the \acrshort{miiii} demo dataset. As a result, we could not use the 5-fold cross-validation approach to train a model reliably.
        \end{tablenotes}
    \end{threeparttable}
        \label{tab:demo_baselines}
\end{table}
\begin{table}[h]
    \center
            \small
     \caption{\textit{Baseline performance on the regression tasks.} Results are reported in \acrlong{mae} ($\downarrow$).}
        \begin{tabularx}{\textwidth}{l>{\centering\arraybackslash}X>{\centering\arraybackslash}X>{\hsize=.20\hsize}X>{\centering\arraybackslash\leavevmode}X>{\centering\arraybackslash\leavevmode}X}
            \toprule \addlinespace[0.2em]  
            \textbf{} & \multicolumn{2}{c}{\textbf{\Acl{kf}}}& & \multicolumn{2}{c}{\textbf{Length of Stay}}\\
            \cmidrule{2-3}\cmidrule{5-6}
            & \textbf{\acrshort{eicu} Demo} & \textbf{\acrshort{miiii} Demo} & & \textbf{\acrshort{eicu} Demo} & \textbf{\acrshort{miiii} Demo} \\
            \cmidrule{2-3}\cmidrule{5-6}\morecmidrules\cmidrule{1-6}
            EN & \textbf{0.30±0.00} & \textbf{0.33±0.03} & & 38.5±0.2 & 52.1±1.3\\
            LGBM & 0.87±0.01 & 0.86±0.04 & & \textbf{37.7±0.2} & \textbf{50.4±1.1}\\
            \addlinespace[0.3em]
            GRU & 0.54±0.02 & 3.23±0.19 & & 39.7±0.6 & 54.7±1.0\\
            LSTM & 0.51±0.02 & 2.95±0.18 & & 39.1±0.5 & 54.8±1.2 \\
            TCN & 0.46±0.02 & 2.98±0.19 & & 38.4±0.6 & 56.8±1.2\\
            TF & 0.60±0.03 & 3.22±0.18 & & 38.9±1.2 & 57.1±1.2 \\
            \bottomrule
        \end{tabularx}
    \label{tab:reg_baselines_demo}
\end{table}
\paragraph{Demo datasets} We offer out-of-the-box (i.e., executable straight after downloading the repository) experiment definitions with five tasks defined on two demo datasets: \acrshort{miiii} demo and \acrshort{eicu} demo. The results can be seen in \autoref{tab:demo_baselines} and \ref{tab:reg_baselines_demo}. The traditional ml models perform better, most likely explained by the low number of samples. The kidney function task highlights the large difference in performance especially.
\paragraph{External validation (extended)}  The \acrfull{los} results for ICU mortality prediction can be found in \autoref{fig:ext_val_los}. 
\paragraph{Fine-tuning (extended)} The AUPRC results for our experiment with transfer learning can be found below and show a similar trend to \autoref{fig:fine-tuning}.
\begin{figure}
    \centering
    \includegraphics[width=0.5\textwidth]{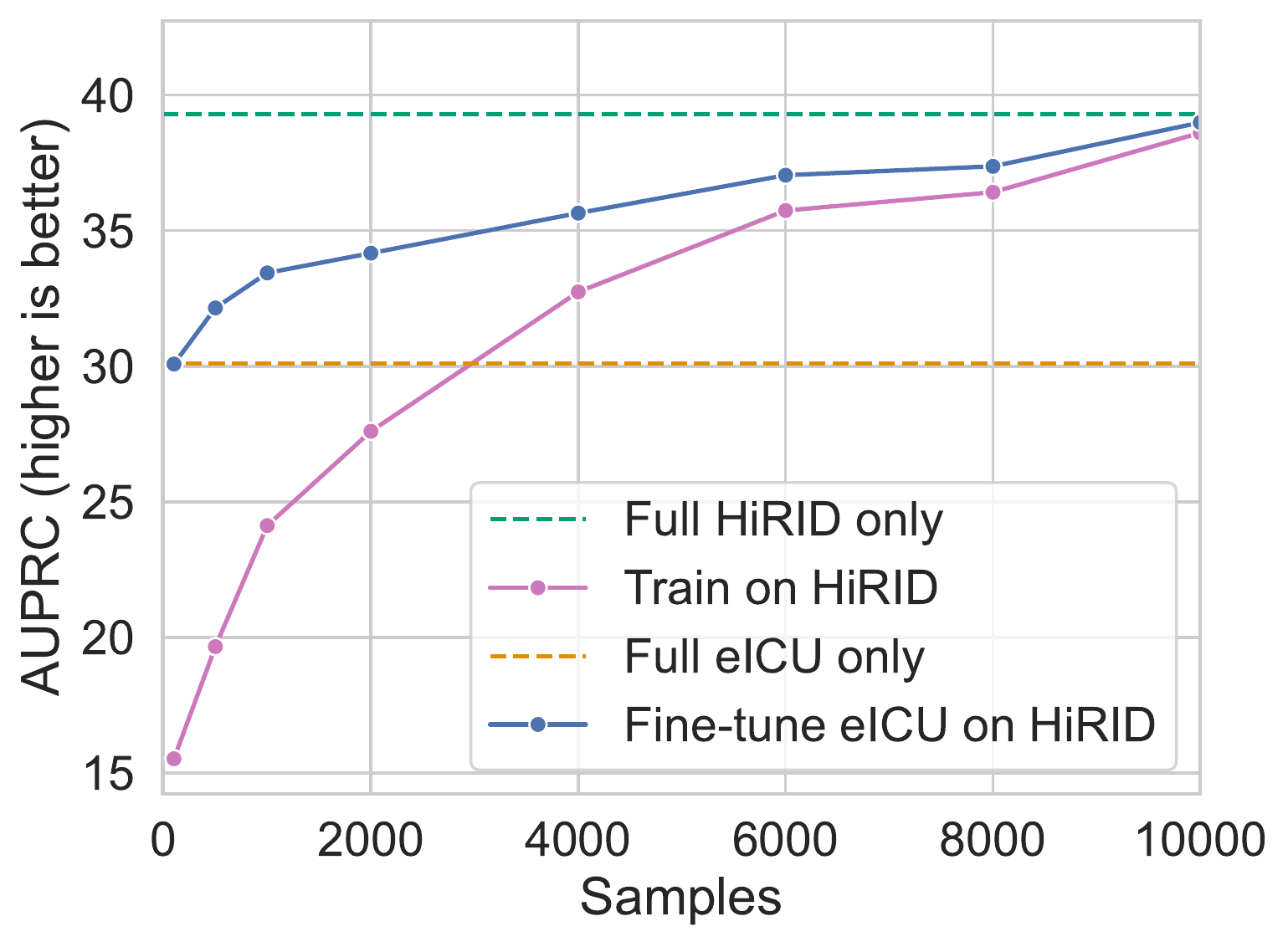}
    \caption{\textit{AUPRC for fine-tuning an eICU GRU model for ICU mortality prediction on HiRID.}}
    \label{fig:fine-tuning-auprcl}
\end{figure}
\FloatBarrier
\section{Appendix: Datasets}
This Appendix contains detailed description of the datasets and the preprocessing methodology.
\label{appendix:datasets}
\subsection{Database characteristics}
The \acrfull{mimic}-III dataset is the most commonly used dataset used for ML in ICU settings; \cite{syedApplicationMachineLearning2021} found 61 eligible studies that used a form of the MIMIC dataset.
It was collected in the USA at the Beth Isreal Deaconess Medical Center~\citep{johnsonMIMICIIIFreelyAccessible2016}. 
The newer MIMIC-IV includes several improvements, among which newer patient records and a revised structure including regular hospital information~\citep{johnsonMIMICIVFreelyAccessible2023}.
 The \acrfull{eicu}~\citep{pollardEICUCollaborativeResearch2018} is an effort to collect the first sizable (200,000 admissions) multi-center dataset. 
It was collected using Philips ICU monitoring systems in the USA at 208 participating hospitals.
The \acrfull{hirid} dataset was collected at Bern University Hospital, Switzerland, and has incorporated more observations than the aforementioned datasets~\citep{hylandEarlyPredictionCirculatory2020}.
The \acrfull{aumc} is the most recently released ICU dataset~\citep{thoralSharingICUPatient2021}. 
Collected in the Netherlands, it has a temporal resolution of up to 1 minute and has prioritized patient de-identification.
Note that there is no benchmark software for this dataset yet.
Each dataset we are using has undergone de-identification procedures, and we have not tried to re-identify the people involved, as per the user agreement for each dataset.
\autoref{tab:dataset_comparison_extended} shows some key characteristics of each dataset. A more comprehensive overview of ICU datasets can be found in the work of \cite{sauerSystematicReviewComparison2022}. 
\begin{table}[h]
\small
\caption{\textit{Supplemental details of openly accessible ICU datasets.} Note that accessing each dataset requires completing a credentialing procedure.}
\label{tab:dataset_comparison_extended}
\centering
\begin{tabular}{|m{0.23\textwidth}>{\centering\arraybackslash}m{0.25\textwidth}>{\centering\arraybackslash} m{0.14\textwidth}>{\centering\arraybackslash} m{0.14\textwidth}>{\centering\arraybackslash}m{0.14\textwidth}|}
\toprule
\textbf{Dataset}                 & \textbf{\acrshort{miiii} / IV} & \textbf{\acrshort{eicu}}  & \textbf{\acrshort{hirid}} & \textbf{\acrshort{aumc}}  \\  
\hline
\textbf{Stays*}              & 40k (0.1k)** / 73k       & 201k  (2k)      & 34k          & 23k    \\
\hline
\textbf{Version} & v1.4 / v2.2       & v2.0     & v1.1.1           & v1.0.2 \\
\hline
\textbf{Frequency (time-series)} & 1 hour    & 5 minutes & 2 / 5 minutes & up to 1 minute \\ 
\hline
\textbf{Origin}                 & USA       & USA       & Switzerland   & Netherlands    \\ 
\hline
\textbf{Originally published} &
2015~\citep{johnsonMIMICIIIFreelyAccessible2016} / 2020~\citep{johnsonMIMICIVFreelyAccessible2023}& 2017~\citep{pollardEICUCollaborativeResearch2018}& 2020~\citep{hylandEarlyPredictionCirculatory2020}& 2019~\citep{thoralSharingICUPatient2021}\\ 
\hline
\textbf{License} & A (C) / A & A (C) & A & B\\
\hline
\textbf{Repository link} &\href{https://physionet.org/content/mimiciii/}{$\rho$} \href{https://physionet.org/content/mimiciii-demo}{($\rho$)}/ \href{https://physionet.org/content/mimiciv/}{$\rho$} &\href{https://physionet.org/content/eicu-crd/}{$\rho$} \href{https://physionet.org/content/eicu-crd-demo}{($\rho$)} & \href{https://physionet.org/content/hirid/}{$\rho$} & \href{https://amsterdammedicaldatascience.nl/amsterdamumcdb/}{$\alpha$}\\
\bottomrule
\end{tabular}
\begin{tablenotes}[flushleft] 
    \item Note that accessing each full dataset requires completing a credentialing procedure. 
    \item *: Stays were taken and rounded from the latest available versions of the databases as of the time of writing. 
    \item **: The brackets () indicate characteristics of the demo (freely accessible) version of the dataset
    \item A: \href{https://physionet.org/content/hirid/view-license/1.1.1/}{PhysioNet Contributor Review Health Data License 1.5.0}
    \item B: \href{https://cdn-links.lww.com/permalink/ccm/g/ccm_49_6_2021_01_18_thoral_ccmed-d-20-02227_sdc3.pdf}{Access Request Form and End User License Agreement for AmsterdamUMCdb 
    1.6}
    \item C: \href{https://opendatacommons.org/licenses/odbl/1-0/}{Open Data Commons Open Database License v1.0}
    \item $\rho$: Physionet
    \item $\alpha$: Amsterdam Medical Data Science
\end{tablenotes}

\end{table}

The authors of \acrshort{miiii} and \acrshort{eicu} have made small selected datasets available for the purpose of experimentation. These datasets are also publicly available on Physionet.
We support the publicly accessible "demo" datasets provided for \acrshort{eicu}\footnote{\url{https://physionet.org/content/eicu-crd-demo}} and \acrshort{miiii}\footnote{\url{https://physionet.org/content/mimiciii-demo}}. 
In accordance with the demo dataset license (Open Data Commons Open Database License v1.0, see \autoref{tab:dataset_comparison_extended}, License C), it is permitted to adapt and share the data. Still, we recommend the user to complete a human subject research training to make sure the usage of the dataset does not violate the usage proposal.
They contain respectively 2,500 (\acrshort{eicu}) and 100 stays (\acrshort{miiii}) before exclusion.
For the purposes testing and validating \acrshort{yaib}, we have created demo-cohorts, \textit{extracted solely from these datasets}, for each of our supported tasks.
Usage of the task cohorts and dataset is only permitted in accordance with the above license.

\subsection{Exclusion criteria}

\begin{figure}
\centering
\includegraphics[width=\textwidth]{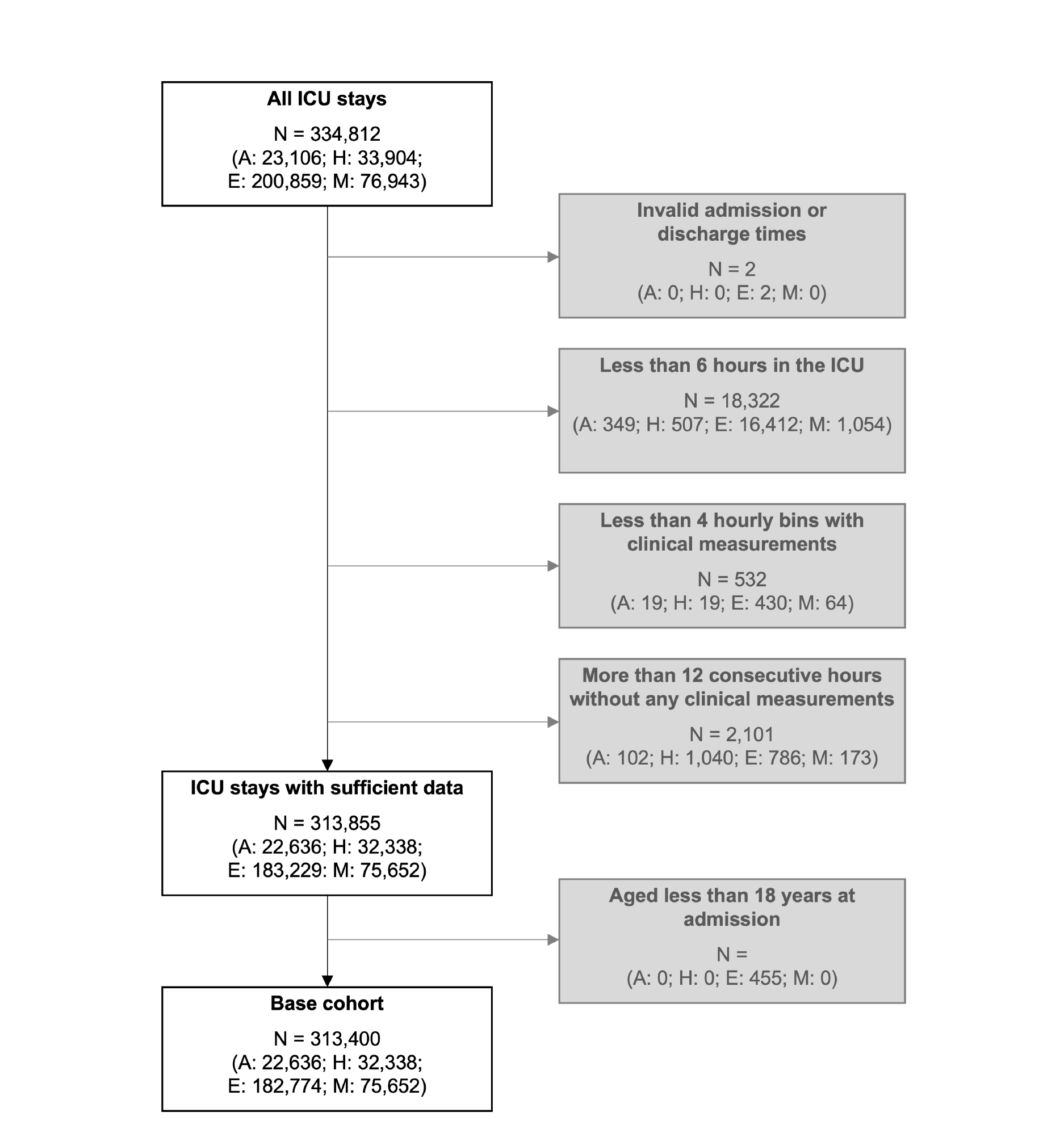}
\caption{\textit{Exclusion criteria applied to the base cohort}. \textbf{N}: Total amount of cases. \textbf{A}: \acrshort{aumc}, \textbf{H}: \acrshort{hirid}, \textbf{E}: \acrshort{eicu}, \textbf{M}: \acrshort{miiv}}
\label{fig:attrition-base}
\end{figure}

We included all available ICU stays of adult patients in our analysis. For each stay, we applied the following exclusion criteria to ensure sufficient data volume and quality: remove any stays with \textbf{1)} an invalid admission or discharge time defined as a missing value or negative calculated length of stay, \textbf{2)} less than six hours spent in the ICU, \textbf{3)} less than four separate hours across the entire stay where at least one feature was measured, \textbf{4)} any time interval of $\geq$12 consecutive hours throughout the stay during which no feature was measured. \autoref{fig:attrition-base} details the number of stays overall and by dataset excluded this way.

\begin{table}
    \small
    \caption{\textit{Characteristics of 1) the included datasets (above) and 2) the task cohorts (below).}}
    \label{tab:characteristics}
    \centering
    \begin{threeparttable}
        \begin{tabularx}{\textwidth}{Xcccc}
            \toprule
             \textbf{General characteristics} & \textbf{\acrshort{aumc}} & \textbf{\acrshort{hirid}} & \textbf{\acrshort{eicu}} & \textbf{\acrshort{miiv}} \\
            \midrule
            \addlinespace[0.3em]
            \textbf{Version} & 1.02 & 1.1.1 & 2.0 & 2.0 \\
            \addlinespace[0.3em]
            \textbf{Number of patients} & 19,790 & -\footnotemark[1] & 160,816 & 53,090 \\
            \textbf{Number of ICU stays} & 22,636 & 32,338 & 182,774 & 75,652  \\
            \addlinespace[0.6em]
            \textbf{Age at admission} (years) & 65 [55, 75]\footnotemark[2] & 65 [55, 75] & 65 [53, 76] & 65 [53, 76]  \\ 
            \textbf{Female} & 7,699 (35) & 11,542 (36) & 83,940 (46) & 33,499 (44) \\
            \multicolumn{5}{l}{\textbf{Race}}\\
            \hspace{1em}Asian & - & - & 3,008 (3) & 2,225 (3) 	\\
            \hspace{1em}Black & - & - & 19,867 (11) & 8,223 (12) \\
            \hspace{1em}White & - & - & 140,938 (78) & 51,575 (76) \\
            \hspace{1em}Other & - & - & 16,978 (9) & 5,514 (8)	\\
            \hspace{1em}Unknown & & & 1,983 & 8,115 \\
            \addlinespace[0.5em]
            \multicolumn{5}{l}{\textbf{Admission type}}\\
            \hspace{1em}Medical & 4,131 (21) & - & 134,532 (79) & 49,217 (65) \\
            \hspace{1em}Surgical & 14,007 (72) & - & 31,909 (19) & 25,674 (34)\\
            \hspace{1em}Other & 1,225 (6) & - & 4,702 (3) & 761 (1)	\\
            \hspace{1em}Unknown & 1,069 & - & 11,631 & 0 \\
            \addlinespace[0.5em]
            \textbf{Hospital length of stay} (days) & - & - & 6 [3, 10] & 7 [4, 13] \\
            \midrule
            \textbf{Task cohorts} & \textbf{\acrshort{aumc}} & \textbf{\acrshort{hirid}} & \textbf{\acrshort{eicu}} & \textbf{\acrshort{miiv}} \\
            \midrule
            \multicolumn{5}{l}{\textbf{ICU mortality}} \\
            \hspace{1em}Number of included stays & 10,535 & 12,859 & 113,382 & 52,045 \\
            \hspace{1em}Died & 1,660 (15.8) & 1,097 (8.2) & 6,253 (5.5) & 3,779 (7.3) \\
            \multicolumn{5}{l}{\textbf{Onset of acute kidney injury}}\\
            \hspace{1em}Number of included stays & 20,290 & 31,772 & 164,791 & 66,032\\
            \hspace{1em}KDIGO* $\geq$ 1 & 3,776 (18.6) & 7,383 (23.2) & 62,535 (37.9) & 27,509 (41.7) \\
            \multicolumn{5}{l}{\textbf{Onset of Sepsis}}\\
            \hspace{1em}Number of included stays & 18,184 & 29,894 & 123,864 & 67,056 \\
            \hspace{1em}Sepsis-3 criteria & 764 (4.2) & 1,986 (6.6) & 5,835 (4.7) & 3,730 (5.6) \\
            \multicolumn{5}{l}{\textbf{Kidney function (creatinine)}}\\
            \hspace{1em}Number of included stays & 8,003 & 7,499 & 69,117 & 35,657 \\
            \hspace{1em}Creatinine value & 0.97 [0.70, 1.61] & 0.92 [0.67, 1.50] & 1.00 [0.71, 1.68] &  1.00 [0.70, 1.60] \\
            \multicolumn{5}{l}{\textbf{ICU remaining length of stay}}\\
            \hspace{1em}Number of included stays & 22,636 & 32,338 & 182,774 & 75,652 \\
            \hspace{1em}ICU length of stay (hours) & 24 [19, 77] & 24 [19, 50] & 42 [23, 76] & 48 [26, 89] \\
            \bottomrule
        \end{tabularx}
        \begin{tablenotes}[flushleft]\setlength\labelsep{0pt} 
            \item \footnotemark[1] \acrshort{hirid} only provides stay-level identifiers.
            \item \footnotemark[2] Since \acrshort{aumc} only includes age groups, we calculated the median of the group midpoints.
            \item * KDIGO, Kidney Disease Improving Global Outcomes~\citep{kdigoKidneyDiseaseImproving2012}.
            \item \textbf{Numeric} variables are summarized by \textit{median [IQR]}.
            \item \textbf{Categorical} variables are summarized by \textit{incidence (\%)}. 
        \end{tablenotes}
    \end{threeparttable}
\end{table}

\begin{sidewaysfigure}
\centering
\includegraphics[width=0.9\textwidth]{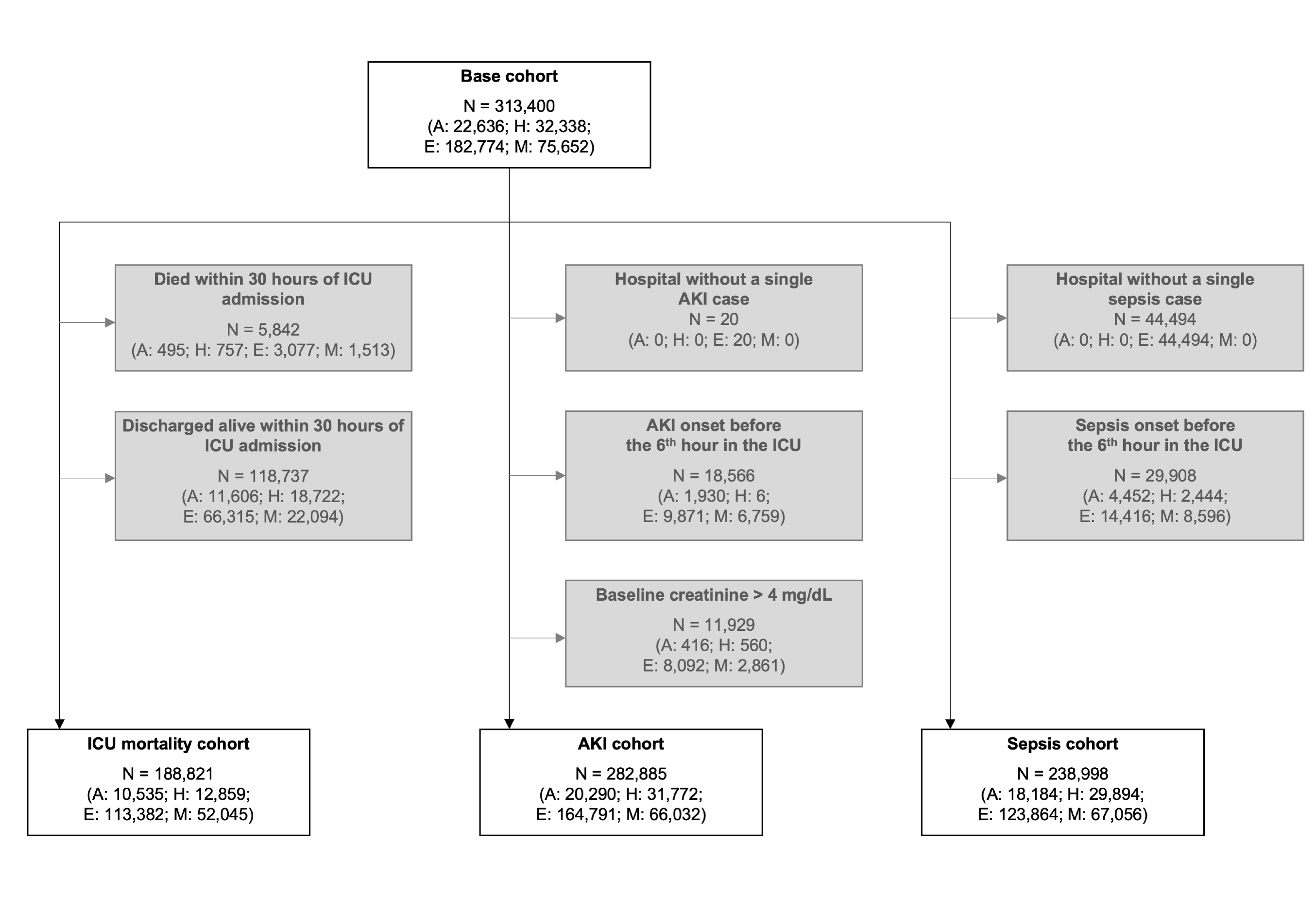}
\caption[short]{Additional exclusion criteria applied for the classification tasks.}
\label{fig:attrition-classification}
\end{sidewaysfigure}
\begin{figure}[ht]
\centering
\includegraphics[width=\textwidth]{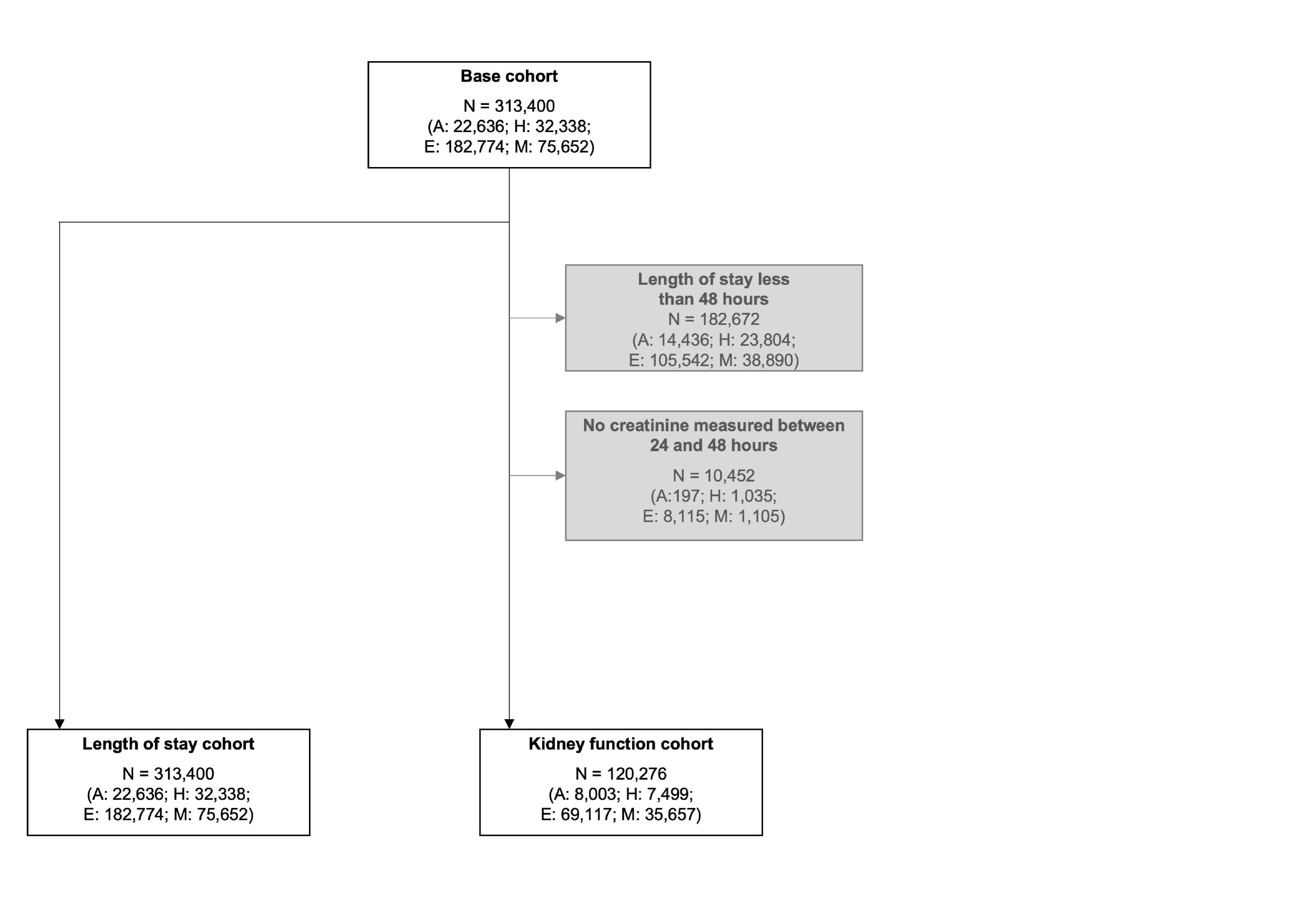}
\caption[short]{Additional exclusion criteria applied for the regression tasks.}
\label{fig:attrition-regression}
\end{figure}
Additional exclusion criteria were applied based on the individual tasks; the details can be found schematically in \autoref{fig:attrition-classification} and \ref{fig:attrition-regression}. For ICU mortality, we excluded all patients with a length of stay of fewer than 30 hours (either due to death or discharge). A minimum length of 30 hours was chosen to exclude any patients that were about to die (the sickest patients) or be discharged (the healthiest patients) at the time of prediction at 24 hours. For creatinine (\acrlong{kf}), we excluded all patients with a length of stay of fewer than 48 hours or without a creatinine measurement between 24 and 48 hours (which was the outcome of interest). For AKI and sepsis, we excluded any stays where disease onset was outside the ICU or within the first six hours of the ICU stay. To account for differences in data recording across hospitals in eICU, we further excluded hospitals that did not have a single patient with AKI or sepsis to exclude hospitals with an insufficient recording of features necessary to define the outcome. Finally, for the AKI task, we excluded stays where the baseline creatinine, defined as the last creatinine measurement prior to ICU (if exists) or the earliest measurement in the ICU, was $\gt$4 mg/dL to exclude patients with preexisting renal insufficiency. For a numerical overview, please consult \autoref{tab:characteristics}.

\subsection{Preprocessing}
A total of 52 features were used for model training (\autoref{tab:supp-features}), 4 of which were static and 48 that were dynamic. These features were selected as they are available across all datasets for most patients. Dynamic features primarily include vital signs (7 variables) and laboratory tests (39 variables), with two more variables that measure input (fraction of inspired oxygen) and output (urine). All variables were extracted via the \href{https://cran.r-project.org/web/packages/ricu/index.html}{\texttt{ricu}} R package (version 0.5.3). The \texttt{ricu} name for each package is shown in \autoref{tab:supp-features}. The exact definition for each feature and how it was extracted from the individual databases can be found in the concept configuration file of the package’s GitHub repository (commit \href{https://github.com/eth-mds/ricu/blob/09902bdc57a1f2f720a924d32f5e053ce2ce7f97/inst/extdata/config/concept-dict.json}{885bd0c}). We also provided cohort definition code for this work, which can be run in both R and Python, in a github repository. 

\begin{longtable}[c]{p{0.58\textwidth}>{\ttfamily}p{0.15\textwidth}>{\itshape}p{0.22\textwidth}}
    \caption{\textit{Clinical concepts used as input to the prediction models.}}\label{tab:supp-features}\\
    \toprule
     Feature & \texttt{ricu} & unit  \\
    \hline
    \addlinespace[0.3em]
    \endfirsthead
    
    \caption{\textit{Clinical concepts used as input to the prediction models (continued)}}\\
    \toprule
     Feature & \texttt{ricu} & unit  \\
    \hline
    \addlinespace[0.3em]
    \endhead
    
    \textbf{Static} &  &  \\
    Age at hospital admission & age & Years \\
    Female sex & sex & - \\
    Patient height & height & cm \\
    Patient weight & weight & kg \\
    & & \\
    \textbf{Time-varying} &  &  \\
    Blood pressure (systolic) & sbp & mmHg \\
    Blood pressure (diastolic) & dbp & mmHg \\
    Heart rate & hr & beats/minute \\
    Mean arterial pressure & map & mmHg \\
    Oxygen saturation & o2sat & \% \\
    Respiratory rate & resp & breaths/minute \\
    Temperature & temp & $^{\circ}$C \\
    Albumin & alb & g/dL \\
    Alkaline phosphatase & alp & IU/L \\
    Alanine aminotransferase & alt & IU/L \\
    Aspartate aminotransferase & ast & IU/L \\
    Base excess & be & mmol/L \\
    Bicarbonate & bicar & mmol/L \\
    Bilirubin (total) & bili & mg/dL \\
    Bilirubin (direct) & bili\_dir & mg/dL \\
    Band form neutrophils & bnd & \% \\
    Blood urea nitrogen & bun & mg/dL \\
    Calcium & ca & mg/dL \\
    Calcium ionized & cai & mmol/L \\
    Creatinine & crea & mg/dL \\
    Creatinine kinase & ck & IU/L \\
    Creatinine kinase MB & ckmb & ng/mL \\
    Chloride & cl & mmol/L \\
    CO$^2$ partial pressure & pco2 & mmHg \\
    C-reactive protein & crp & mg/L \\
    Fibrinogen & fgn & mg/dL \\
    Glucose & glu & mg/dL \\
    Haemoglobin & hgb & g/dL \\
    International normalised ratio (INR) & inr\_pt & - \\
    Lactate & lact & mmol/L \\
    Lymphocytes & lymph & \% \\
    Mean cell haemoglobin & mch & pg \\
    Mean corpuscular haemoglobin concentration & mchc & \% \\
    Mean corpuscular volume & mcv & fL \\
    Methaemoglobin & methb & \% \\
    Magnesium & mg & mg/dL \\
    Neutrophils & neut & \% \\
    O$^2$ partial pressure & po2 & mmHg \\
    Partial thromboplastin time & ptt & sec \\
    pH of blood & ph & - \\
    Phosphate & phos & mg/dL \\
    Platelets & plt & 1,000 / $\mu$L \\
    Potassium & k & mmol/L \\
    Sodium & na & mmol/L \\
    Troponin T & tnt & ng/mL \\
    White blood cells & wbc & 1,000 / $\mu$L \\
    Fraction of inspired oxygen & fio2 & \% \\
    Urine output & urine & mL \\
    \bottomrule
\end{longtable}
\paragraph*{Additional features}
Furthermore, we consulted clinical experts to identify which features might be missing from our prediction setup. Several clinical features are currently missing from this setup, which could potentially improve prediction performance: \textit{Glasgow coma scale score, Intubation, Ventilator settings, Renal replacement therapy, and Vasopressors}. We expect to be able to integrate more concepts as we collaborate with authors of datasets to make them available.

\section{Appendix: Outcome definitions}
\label{appendix:outc-detail}
The outcome definitions per task for each dataset are detailed in this Appendix.
\subsection{ICU mortality}
ICU mortality was defined as death while in the ICU. This was generally ascertained via the recorded discharge status or discharge destination. Note that our definition of ICU mortality differs from the definition of \verb|death| in the \texttt{ricu} R package, which describes hospital mortality that is unavailable for some included datasets.

\textbf{AUMCdb} Death was inferred from the \verb|destination| column of the \verb|admissions| table. A destination of ``Overleden'' (Dutch for ``passed away'') was treated as a death in the ICU. Since the date of death was recorded outside of the ICU and may therefore be imprecise, the recorded ICU discharge date was used as a more precise proxy for the time of death. 

\textbf{HiRID} Death was inferred from the column \verb|discharge_status| in table \verb|general|. The status of ``dead'' was treated as a death in the ICU. Time of death was inferred as the last measurement of IDs \verb|110| (mean arterial blood pressure) or \verb|200| (heart rate) in column \verb|variableid| of table \verb|observations|.

\textbf{eICU} Death was inferred from the column \verb|unitdischargestatus| in table \verb|patient|. The status of ``Expired'' was treated as a death in the ICU. The recorded ICU discharge date was used as a proxy for the time of death.

\textbf{MIMIC IV} Death was inferred from the column \verb|hospital_expire_flag| in table \verb|admissions|. Since MIMIC IV only records a joint ICU/hospital expiration flag, ward transfers were analyzed to ascertain the location of death. If the last ward was the ICU, the death was considered ICU mortality.

\begin{table}[h]
    \caption{Staging of AKI according to KDIGO~\citep{kdigoKidneyDiseaseImproving2012}}
    \centering
    \label{tab:kdigo}
    \begin{threeparttable}
        \begin{tabularx}{\textwidth}{l>{\centering\arraybackslash}X>{\centering\arraybackslash}X}
            \toprule
            \textbf{Stage} & \textbf{Serum creatinine} & \textbf{Urine output}  \\
            \midrule
            1 & \makecell{1.5–1.9 times baseline \\\\ OR \\\\ $\geq$0.3 mg/dl ($\geq$26.5 $\mu$mol/l) increase \\ within 48 hours} & $\lt$0.5 ml/kg/h for 6–12 hours \\
            \midrule
            2 & \makecell{2.0–2.9 times baseline } & $\lt$0.5 ml/kg/h for $\geq$12 hours \\
            \midrule
            3 & \makecell{3.0 times baseline (prior 7 days) \\\\ OR \\\\ Increase in serum creatinine to \\ $\geq$4.0 mg/dl ($\geq$353.6 $\mu$mol/l) \\ within 48 hours \\\\ OR \\\\ Initiation of renal replacement therapy} & \makecell{$\lt$0.3 ml/kg/h for $\geq$24 hours \\\\ OR \\\\ Anuria for $\geq$12 hours}\\
            \bottomrule
        \end{tabularx}
        \begin{tablenotes}[flushleft, para]
            \footnotesize
            \item \leavevmode\kern-\scriptspace\kern-\labelsep AKI, acute kidney injury; KDIGO, Kidney Disease Improving Global Outcomes.
        \end{tablenotes}
    \end{threeparttable}
\end{table}

\subsection{Acute kidney injury}
AKI was defined as KDIGO stage $\geq$1, either due to an increase in serum creatinine or low urine output (\autoref{tab:kdigo})~\citep{kdigoKidneyDiseaseImproving2012}. Baseline creatinine was defined as the lowest creatinine measurement over the last 7 days. Urine rate was calculated as the amount of urine output in ml divided by the number of hours since the last urine output measurement (for a max gap of 24h), except for HiRID, in which urine rate was recorded directly. The earliest urine output was divided by 1. The rate per kg was calculated based on the admission weight. If weight was missing, a weight of 75 kg was assumed instead.

\textbf{AUMCdb} Creatinine was defined via the standard \texttt{ricu} concept of serum creatinine as IDs \verb|6836|, \verb|9941|, or \verb|14216| in column \verb|itemid| of table \verb|numericitems|. Urine output was defined as IDs \verb|8794|, \verb|8796|, \verb|8798|, \verb|8800|, \verb|8803| in column \verb|itemid| of table \verb|numericitems| (note that this includes more items than those included in the standard \texttt{ricu} concept of urine output). 

\textbf{HiRID} Creatinine was defined via the standard \texttt{ricu} concept of serum creatinine as ID \verb|20000600| in column \verb|variableid| of table \verb|observations|. Urine rate was defined as ID \verb|10020000| in column \verb|variableid| of table \verb|observations|. 

\textbf{eICU} Creatinine was defined via the standard \texttt{ricu} concept of serum creatinine as IDs ``creatinine'' in column \verb|labname| of table \verb|lab|. Urine output was defined via the standard \texttt{ricu} concept of urine output as IDs ``Urine'' and ``URINE CATHETER'' in column \verb|celllabel| of table \verb|intakeoutput|.

\textbf{MIMIC IV} Creatinine was defined via the standard \texttt{ricu} concept of serum creatinine as ID \verb|50912| in column \verb|itemid| of table \verb|labevents|. Urine output was defined via the standard \texttt{ricu} concept of urine output as IDs \verb|226557|, \verb|226558|, \verb|226559|, \verb|226560|, \verb|226561|, \verb| 226563|, \verb|226564|, \verb|226565|, \verb|226566|, \verb|226567|, \verb|226584|, \verb|227510| in column \verb|itemid| of table \verb|outputevents|.

\begin{table}[h]
\center
\caption{\textit{Comparing \acrshort{miiv} sepsis cohorts according to three different definitions.}}
\begin{threeparttable}
\small
\renewcommand{\arraystretch}{1.1}
\begin{tabular}{lccc}
\toprule
\addlinespace[0.2em]
\textbf{Cohort}                    & \textbf{\cite{seymourAssessmentClinicalCriteria2016}*} & \textbf{\cite{moorPredictingSepsisMultisite2021}} & \textbf{\cite{calvertComputationalApproachEarly2016}} \\
\addlinespace[0.2em]
\hline
\multicolumn{1}{l}{\textbf{Stays}} & 67,056 & 53,642 & 65,901\\
\multicolumn{1}{l}{\textbf{Prevalence}} & 3,730 (5.6\%) & 8,919 (16.7\%) & 2,406 (3.7\%)\\
\bottomrule
\end{tabular}
\begin{tablenotes}[flushleft]
    \item *Our default definition.
\end{tablenotes}
\end{threeparttable}
\label{tab:sepsis_definition}
\end{table}
\subsection{Sepsis}
The onset of sepsis was defined using the Sepsis-3 criteria~\citep{singerThirdInternationalConsensus2016}, which defines sepsis as organ dysfunction due to infection. Following guidance from the original authors of Sepsis-3 \citep{seymourAssessmentClinicalCriteria2016}, organ dysfunction was defined as an increase in SOFA score $\geq$2 points compared to the lowest value over the last 24 hours. Suspicion of infection was defined as the simultaneous use of antibiotics and culture of body fluids. The time of sepsis onset was defined as the first time of organ dysfunction within 48 hours before and 24 hours after suspicion of infection. Time of suspicion was defined as the earlier antibiotic initiation or culture request. Antibiotics and culture were considered concomitant if the culture was requested $\leq$24 hours after antibiotic initiation or if antibiotics were started $\leq$72 hours after the culture was sent to the lab. Where available, antibiotic treatment was inferred from administration records; otherwise, we used prescription data. To exclude prophylactic antibiotics, we required that antibiotics were administered continuously for $\geq$3 days~\citep{reynaEarlyPredictionSepsis2019}. Antibiotic treatment was considered continuous if an antibiotic was administered once every 24 hours for 3 days (or until death) or was prescribed for the entire time spent in the ICU. \acrshort{hirid} and \acrshort{eicu} did not contain microbiological information. For these datasets, we followed \cite{moorPredictingSepsisMultisite2021} and defined suspicion of infection through antibiotics alone. Note, however, that the sepsis prevalence in our study was considerably lower than theirs, which was as high as 37\% in \acrshort{hirid}. We suspect this is because they did not require treatment for $\geq$3 days. For comparison and to contextualize the results of our experiments, we have benchmarked other sepsis definitions. See \autoref{tab:sepsis_definition} for the number of stays and incidence for each cohort in this experiment performed on \acrshort{miiv}.

\textbf{\acrshort{aumc}} The SOFA score, microbiological cultures, and antibiotic treatment were defined via the standard \texttt{ricu} concepts \verb|sofa|, \verb|abx|, and \verb|samp| (see the \href{https://github.com/eth-mds/ricu/blob/09902bdc57a1f2f720a924d32f5e053ce2ce7f97/inst/extdata/config/concept-dict.json}{\texttt{ricu}} package for more details). 

\textbf{\acrshort{hirid}} The SOFA score and antibiotic treatment were defined via the standard \texttt{ricu} concept \verb|sofa| and \verb|abx| (see the \href{https://github.com/eth-mds/ricu/blob/09902bdc57a1f2f720a924d32f5e053ce2ce7f97/inst/extdata/config/concept-dict.json}{\texttt{ricu}} package for more details). No microbiology data were available in HiRID. 

\textbf{\acrshort{eicu}} The SOFA score and antibiotic treatment were defined via the standard \texttt{ricu} concept \verb|sofa| and \verb|abx| (see the \href{https://github.com/eth-mds/ricu/blob/09902bdc57a1f2f720a924d32f5e053ce2ce7f97/inst/extdata/config/concept-dict.json}{\texttt{ricu}} package for more details). Microbiology data in eICU was not reliable \citep{moorPredictingSepsisMultisite2021} and therefore omitted.

\textbf{\acrshort{miiv}} The SOFA score and microbiological cultures were defined via the standard \texttt{ricu} concepts \verb|sofa| and \verb|samp| (see the \href{https://github.com/eth-mds/ricu/blob/09902bdc57a1f2f720a924d32f5e053ce2ce7f97/inst/extdata/config/concept-dict.json}{\texttt{ricu}} package for more details). Antibiotics were defined based on table \verb|inputevents|. This differs from the standard \texttt{ricu} \verb|abx| concept, which also considers the \verb|prescriptions| table. 

\subsection{Kidney function}

The median creatinine level over the course of the second day of a patient’s ICU stay was defined as the target of interest. Given the available datasets, this was chosen as the most suitable proxy of kidney function.

\textbf{AUMCdb} Creatinine was defined via the standard \texttt{ricu} concept of serum creatinine as IDs \verb|6836|, \verb|9941|, or \verb|14216| in column \verb|itemid| of table \verb|numericitems|. 

\textbf{HiRID} Creatinine was defined via the standard \texttt{ricu} concept of serum creatinine as ID \verb|20000600| in column \verb|variableid| of table \verb|observations|. 

\textbf{eICU} Creatinine was defined via the standard \texttt{ricu} concept of serum creatinine as IDs ``creatinine'' in column \verb|labname| of table \verb|lab|. 

\textbf{MIMIC IV} Creatinine was defined via the standard \texttt{ricu} concept of serum creatinine as ID \verb|50912| in column \verb|itemid| of table \verb|labevents|. 

\subsection{Remaining length of stay}

At each hour, the remaining length of stay in the ICU was calculated in hours until discharge. A maximum forecasting window of 7 days was chosen, as forecasts beyond this interval were judged extremely difficult and of lesser clinical relevance. Arguably, an even shorter window of 3 days could be chosen to support short-term capacity planning. This is left for future investigation.

\textbf{AUMCdb} Length of stay was calculated via the standard  \texttt{ricu} concept using the columns \verb|admittedat| and \verb|dischargedat| of table \verb|admissions|.

\textbf{HiRID} Length of stay was calculated via the standard  \texttt{ricu} concept using the column \verb|admissiontime| of table \verb|general| as well as the last observation in table \verb|observations|.

\textbf{eICU} Length of stay was calculated via the standard  \texttt{ricu} concept using the columsn \verb|unitadmitoffset| and \verb|unitdischargeoffset| of table \verb|patient|.

\textbf{MIMIC IV} Length of stay was calculated via the standard  \texttt{ricu} concept using the columsn \verb|intime| and \verb|outtime| of table \verb|icustays|.

\section{Appendix: YAIB's usage and implementation}
\label{app:usage}
We describe the most important practical aspects of using \acrshort{yaib} in research in this Appendix. Please note that there is a \textit{wiki}, dedicated to usage and development, available at: \url{https://github.com/rvandewater/YAIB/wiki}.
\label{appendix:yaib_example}
\subsection{Use of existing code repositories}
\label{app:existing_code}
As mentioned in the main text, we used parts of the HiRID-Benchmark code and heavily modified and extended the code to support our extra features and extensibility, mentioned in \autoref{tab:benchmark_comparison}. HiRID-Benchmark is available at \href{https://github.com/ratschlab/HIRID-ICU-Benchmark}{GitHub} and makes use of an \href{https://github.com/ratschlab/HIRID-ICU-Benchmark/blob/master/LICENSE}{MIT license}, as does our code repository.

\subsection{Adding a data source}
\label{subsec:new_dataset}
Adding a new dataset type, to use within \acrshort{yaib}, can be easily done by providing it in a \texttt{.gin} task definition file, see \autoref{code:preprocessing-definition}. Note, however, that any datasets formatted in the default way do not require any changes to be used by \acrshort{yaib}. By default, we have chosen to work with the Apache parquet~\citep{vohraApacheParquet2016} file format, which is a modern, open-source column-oriented format that does not require a lot of storage due to efficient data compression\footnote{\url{https://parquet.apache.org/}}. We separate the data into three separate files: \texttt{DYNAMIC}, \texttt{STATIC}, and \texttt{OUTCOME}; this is defined for dynamic variables (that change during the stay), constant parameters, and the prediction task label respectively. Our 
cohort definition code produces the files in exactly this format. Furthermore, we see the concept of \texttt{roles} with the definition of the \texttt{vars} dictionary. These roles are assigned as defined in \preproc{} 
, the preprocessing package developed alongside \acrshort{yaib}. The \texttt{GROUP} variable defines which internal dataset variable should be used to ``group by'' for, e.g., aggregating patient vital signs. The \texttt{SEQUENCE} variable defines the sequential dimension of the dataset (in the common case, this would be time). The other keys in this dictionary define the feature columns and outcome variables for prediction.
\begin{lstlisting}[frame=single, style=pycharm, language=Python, caption={\textit{Example preprocessing pipeline structure.} 
%See \url{https://github.com/rvandewater/YAIB/blob/development/icu_benchmarks/data/preprocessor.py} for the full file.
}, label=code:preprocessing-definition, columns=fullflexible, basicstyle=\ttfamily\tiny]
@gin.configurable("base_classification_preprocessor")
class DefaultClassificationPreprocessor(Preprocessor):
    def __init__(self, generate_features: bool = True, scaling: bool = True, use_static_features: bool = True):
        """
        Args:
            generate_features: Generate features for dynamic data.
            scaling: Scaling of dynamic and static data.
            use_static_features: Use static features.
        Returns:
            Preprocessed data.
        """


    def apply(self, data, vars):
        """
        Args:
            data: Train, validation and test data dictionary. Further divided in static, dynamic, and outcome.
            vars: Variables for static, dynamic, outcome.
        Returns:
            Preprocessed data.
        """
        ...
        return data

    def _process_static(self, data, vars):
        ...
        return data

    def _process_dynamic(self, data, vars):
        ...
        return data

    def _dynamic_feature_generation(self, data, dynamic_vars):
        ...
        return data

\end{lstlisting}
\subsection{Creating a preprocessing pipeline}
Our preprocessing pipeline is set up to be as general as possible and allows for custom implementations, defined as subclass from the \texttt{Preprocessor} class and passed as a command-line argument. For our tasks, we have defined a default preprocessing pipeline for both classification and regression tasks. \autoref{code:task-definition} shows the class structure of the default classification preprocessor. In the private methods of this class, \preproc{} is used to apply feature generation steps (which differ for ml and dl models). The abstract \texttt{Preprocessor} has two functions that need to be implemented: \texttt{\_\_init\_\_()} (which initializes the preprocessor and configures the settings) and \texttt{apply(data)} (which returns the preprocessed data dictionary of features and labels for each of the train, validate, and test splits)

\subsection{Example: training a mortality prediction LSTM model}
\begin{figure}[h]
    \centering
    \includegraphics[width=0.7\textwidth]{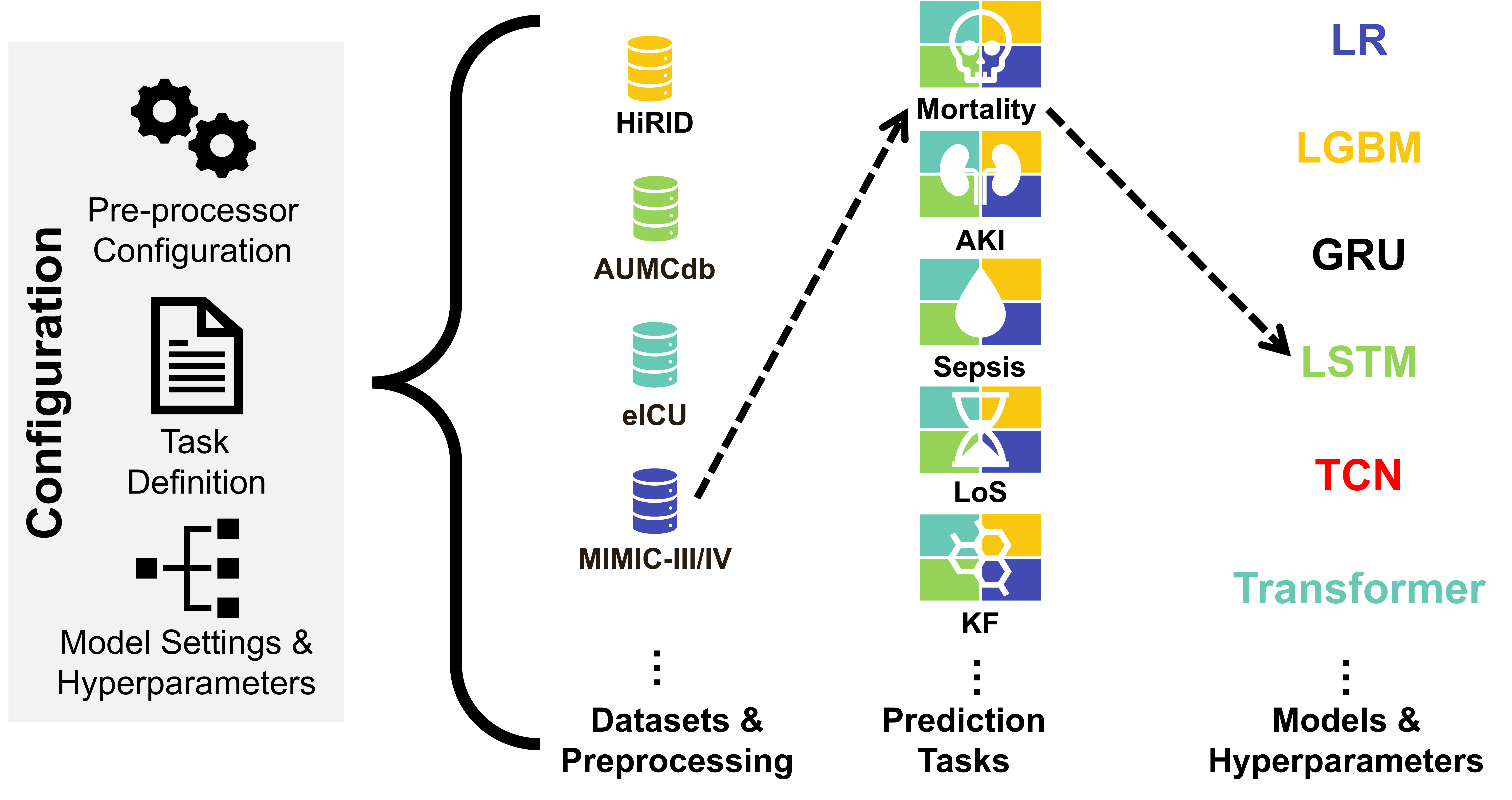}
    \caption{\textit{Experiment definition schematic.} The fundamental experiment configuration of the benchmark contains three basic elements, \textbf{1)} the dataset, \textbf{2)} the prediction task, and \textbf{3)} the model and (list of) hyperparameters. Each element can be combined in different ways. Additionally, we provide an interface for extending each element (Datasets \& Preprocessing, Prediction Tasks, Models \& Hyperparameters) in the process. Provided is an example of an experiment configuration: predicting Sepsis on MIMIC~\citep{thoralSharingICUPatient2021} with an LSTM~\citep{hochreiterLongShortTermMemory1997} model.}
    \label{fig:yaib_schematic}
\end{figure}
We demonstrate the complete process of training a \acrfull{lstm} model to predict sepsis on the MIMIC-III demo dataset with \acrshort{yaib}. This is shown schematically in \autoref{fig:yaib_schematic}.
\begin{lstlisting}[frame=single, float=h, style=pycharm, language=Python, caption={\textit{Example configuration for a task definition configuration.} We identify 4 main categories: the prediction type (classification or regression), the deep learning loss, the dataset variables, and preprocessing settings, and the cross-validation settings. Note that the dataset is fully configurable here.}, label=code:task-definition, columns=fullflexible, basicstyle=\ttfamily\tiny]
# COMMON IMPORTS
include "configs/tasks/common/Imports.gin"

# MODE SETTINGS
Run.mode = "Classification"
NUM_CLASSES = 2 # Binary classification
HORIZON = 24
train_common.weight = "balanced"

# DEEP LEARNING
DLPredictionWrapper.loss = @cross_entropy
# DATASET AND PREPROCESSING
preprocess.file_names = {
    "DYNAMIC": "dyn.parquet",
    "OUTCOME": "outc.parquet",
    "STATIC": "sta.parquet",
}
vars = {
    "GROUP": "stay_id",
    "LABEL": "label",
    "SEQUENCE": "time",
    "DYNAMIC": ["alb", "alp", "alt", "ast", "be", "bicar", "bili", "bili_dir", "bnd", "bun", "ca", "cai", "ck", "ckmb", "cl",
        "crea", "crp", "dbp", "fgn", "fio2", "glu", "hgb", "hr", "inr_pt", "k", "lact", "lymph", "map", "mch", "mchc", "mcv",
        "methb", "mg", "na", "neut", "o2sat", "pco2", "ph", "phos", "plt", "po2", "ptt", "resp", "sbp", "temp", "tnt", 
        "urine", "wbc"],
    "STATIC": ["age", "sex", "height", "weight"],
}

# SELECTING PREPROCESSOR
preprocess.preprocessor = @base_classification_preprocessor
preprocess.vars = %vars
preprocess.use_static = True

# SELECTING DATASET
PredictionDataset.vars = %vars

# CROSS VALIDATION
execute_repeated_cv.cv_repetitions = 5
execute_repeated_cv.cv_folds = 5
\end{lstlisting}
 In \autoref{code:task-definition}, the basic task setup for mortality prediction after 24 hours is shown. We define the dataset files, by default split into 3 parquet files with the corresponding names. The listing describes three dataset components: dynamic data, outcome definitions, and static data. Below, one sees the variables and different \textit{``roles''} assigned to concrete strings (see Appendix \ref{subsec:new_dataset} for detail). In this listing, we also pass the vars to the preprocessing and dataset class. Finally, we see the definition of the cross-validation folds and iterations. 

\begin{lstlisting}[frame=single, float=h, style=pycharm, language=Python, caption={\textit{Example configuration for hyperparameters of LSTM.} Tunable hyperparameter ranges can be floating points, integers, and categorical values.}, columns=fullflexible, label=code:lstm-gin, basicstyle=\ttfamily\tiny]
import gin.torch.external_configurables
import icu_benchmarks.models.wrappers
import icu_benchmarks.models.encoders

default_preprocessor.generate_features = False

# Train params
train_common.model = @DLWrapper()

DLWrapper.encoder = @LSTMNet()
DLWrapper.optimizer_fn = @Adam
DLWrapper.train.epochs = 1000
DLWrapper.train.batch_size = 64
DLWrapper.train.patience = 10
DLWrapper.train.min_delta = 1e-4

# Optimizer params
optimizer/hyperparameter.class_to_tune = @Adam
optimizer/hyperparameter.weight_decay = 1e-6
optimizer/hyperparameter.lr = (1e-5, 3e-4)

# Encoder params
model/hyperparameter.class_to_tune = @LSTMNet
model/hyperparameter.num_classes = %NUM_CLASSES
model/hyperparameter.hidden_dim = (32, 256, "log-uniform", 2)
model/hyperparameter.layer_dim = (1, 3)

# Hyperparamter tuning
tune_hyperparameters.scopes = ["model", "optimizer"]
tune_hyperparameters.n_initial_points = 5
tune_hyperparameters.n_calls = 30
tune_hyperparameters.folds_to_tune_on = 2
\end{lstlisting}
\begin{lstlisting}[frame=single, language=bash,float=h, caption={\textit{Running YAIB.} We train the LSTM model on the MIMIC-III demo dataset for the Mortality24 task.}, label=code:run-benchmark, columns=fullflexible, basicstyle=\ttfamily\tiny]
#!
(~\textcolor{blue}{icu-benchmarks} ~) train \
    -d demo_data/mortality24/mimic_demo \
    -n mimic_demo \
    -t BinaryClassification \
    -tn Mortality24 \
    --log-dir ../yaib_logs/ \
    -m LSTM \
    -gc \
    -lc \
    -s 2222 \
    -l ../yaib_logs/ \
    --tune
    
\end{lstlisting}
In \autoref{code:lstm-gin} we see the configuration for the \acrshort{lstm}. We first define the generating features from dynamic data (relevant for traditional ml). Then we bind the LSTM model with a gin flag. After this, the hyperparameters are specified. The optimizer and encoder parameters are then specified. Note that we can specify ranges of hyperparameters to be tuned by the hyperparameter optimizer. Settings for this can be found in the bottom cluster of code. 
\autoref{code:run-benchmark} shows how to train our LSTM model on the \texttt{mimic\_demo} dataset (included in our repository). 
\FloatBarrier

\subsection{Options}
\begin{table}[h]
\small
    \centering
        \caption{\textit{Options for the \texttt{train} command.}}
    \begin{tabularx}{\textwidth}{>{\ttfamily\hsize=.50\hsize}X>{\centering\arraybackslash\hsize=.20\hsize}X>{\centering\arraybackslash}X|}
        \toprule
        \textbf{Flag} & \textbf{Required} & \textbf{Description} \\
            \midrule \midrule
    --reproducible & No & Make torch reproducible. (default: True)\\
    \hline
    -hp, --hyperparams & No & Hyperparameters for model.\\
    \hline
    --tune & No & Find best hyperparameters. (default: False)\\
    \hline
    --checkpoint & No & Use previous checkpoint.\\
    \bottomrule
\end{tabularx}
\label{tab:yaib_train_args}
\end{table}
\begin{table}[h]
\small
    \centering
        \caption{\textit{Options for the \texttt{evaluate} command.}}
    \begin{tabularx}{\textwidth}{>{\ttfamily\hsize=.50\hsize}X>{\centering\arraybackslash\hsize=.20\hsize}X>{\centering\arraybackslash}X|}
        \toprule
        \textbf{Flag} & \textbf{Required} & \textbf{Description} \\
            \midrule \midrule
        -sn --source-name & Yes & Name of the source dataset. \\ 
        \hline
        --source-dir & Yes & Directory containing gin and model weights. \\
        \bottomrule
    \end{tabularx}
\label{tab:yaib_evaluate_args}
\end{table}
\begin{table}[h]
    \centering
    \small
    \caption{\textit{General arguments for the use of \acrshort{yaib}.}}
    \begin{tabularx}{\textwidth}{>{\ttfamily\hsize=.50\hsize}X>{\centering\arraybackslash\hsize=.20\hsize}X>{\centering\arraybackslash}X|}
    \toprule
      \textbf{Flag} & \textbf{Required} & \textbf{Description} \\
    \midrule \midrule
      -d, --data-dir & Yes & Path to the parquet data directory. \\
      \hline
      -n, --name & Yes & Name of the (target) dataset.\\
      \hline
       -t, --task & Yes & Name of the task gin.\\
      \hline
       -m, --model & No & Name of the model gin )Default.\\
      \hline
      -tn, --task-name & No & Name of the task, used for naming experiments.\\
      \hline
      -e, --experiment & No & Name of the experiment gin.\\
      \hline
      -l, --log-dir & No & Log directory with model weights.\\
      \hline
      -s, --seed & No & Random seed for processing, tuning, and training.\\
      \hline
      -v, --verbose & No & Whether to use verbose logging. (default: True)\\
      \hline
      --cpu  & No & Set to use CPU. (default: False)\\
      \hline
      -db, --debug & No & Set to load less data. (default: False)\\
      \hline
      -lc, --load\_cache & No & Set to load generated data cache. (default: False)\\
      \hline
      -gc, --generate\_cache& No &Set to generate data cache. (default: False)\\
      \hline
      -p, --preprocessor & No & Load custom preprocessor from file.\\
      \hline
      -pl, --plot & No & Generate common plots. (default: False)\\
      \hline
      -wd, --wandb-sweep &  No & Activates \acrlong{wandb} hyper parameter sweep. (default: False)\\
      \hline
      -imp, --pretrained-imputation & No &  Path to pre trained imputation model.\\
      \bottomrule
    \end{tabularx}
    \label{tab:yaib_general_args}
\end{table}

We specify the command line options that YAIB provides for benchmarking prediction tasks. One can use the \texttt{icu-benchmarks} command with either \texttt{train} or \texttt{evaluate}. \autoref{tab:yaib_train_args} shows the arguments that can be used with \texttt{train}. \autoref{tab:yaib_evaluate_args} contains the options to use with \texttt{evaluate} (i.e., for evaluation with a pre-trained model). 
In \autoref{tab:yaib_general_args}, the general options for \acrshort{yaib} are shown. 
\FloatBarrier

\newpage
\section{Appendix: Extending YAIB}
\label{app:extend}

\acrshort{yaib} is built to adapt to your needs with as little effort as possible. \acrshort{yaib} allows you to change any part of your pipeline with ease: add a new dataset, define additional clinical concepts, adapt a task, implement imputation algorithms, use additional models, or evaluate custom metrics. This section provides examples on \acrshort{yaib} may be extended with respect to each of the above.

\begin{lstlisting}[frame=single, float=h, style=pycharm, language=json, caption={\textit{ID and table configuration for the Salzburg Intensive Care Database Database in JSON.}}, columns=fullflexible, label=code:config-salzburg, basicstyle=\ttfamily\tiny]
{
    "name": "sic",
    "id_cfg": {
      "patient": {
        "id": "patientid",
        "position": 1,
        "start": "firstadmission",
        "end": "offsetofdeath",
        "table": "cases"
      },
      "icustay": {
        "id": "caseid",
        "position": 2,
        "start": "offsetafterfirstadmission",
        "end": "timeofstay",
        "table": "cases"
      }
    },
    "tables": {
      "cases": {
        "files": "cases.csv.gz",
        "defaults": {
          "index_var": "offsetafterfirstadmission",
          "time_vars": ["offsetafterfirstadmission", "offsetofdeath"]
        },
        "cols": {
          "caseid": {
            "name": "CaseID",
            "spec": "col_integer"
          },
          "patientid": {
            "name": "PatientID",
            "spec": "col_integer"
          },
          "admissionyear": {
            "name": "AdmissionYear",
            "spec": "col_integer"
          },
          ...
        }
      },
      "d_references": {
        ...
      },
      ...
    }
}
\end{lstlisting}

\subsection{Add a new dataset: Salzburg Intensive Care Database (SICdb)}

\acrshort{sicdb} is a recently published open source ICU dataset \citep{rodemundSalzburgIntensiveCare2023}. The dataset includes ~27,000 admissions to the ICU at University Hospital Salzburg (Austria) from 2013 to 2021. \acrshort{sicdb} provides highly granular data with up to minute level resolution. To make \acrshort{sicdb} available within \acrshort{yaib}, it must be interfaced to \texttt{ricu}. This ensures that \texttt{ricu} knows how to access and extract data from the dataset. Interfacing a new dataset requires three key steps: 

\begin{enumerate}
    \item Define the table structure and possible ID types as a JSON configuration.
    \item Define how all ID systems and their origin times relate to each other.   
\end{enumerate}

\paragraph*{Define ID types and table structure}

First, \texttt{ricu} needs to know what tables and columns exist within the data source. This is specified via the JSON configuration file partially shown in \autoref{code:config-salzburg}. Tables are defined under the \texttt{tables} element. The definition of each table contains the name of its source file, usually provided in \texttt{.csv} or \texttt{.csv.gz} format, and a list of all columns and their data types. In addition, default roles can be defined for certain columns in the table. These usually include the time index (if present), all other time columns, and a column that is considered to contain the value of interest. 

In addition to the available tables, \texttt{ricu} also expects information about the main ID types used in the dataset. Each piece of information in ICU datasets is usually linked to a certain unit of observation, most commonly the patient (\texttt{patient}), the hospital admission (\texttt{hospadm}), or the specific ICU stay (\texttt{icustay}). By knowing what IDs a piece of information is measured for, \texttt{ricu} is able to temporally relate all information within the dataset. For example, \texttt{labevents} in MIMIC IV are recorded for hospital admissions, whereas \texttt{chartevents} are recorded for ICU stays. Defining how these two ID systems relate to each other allows them to be mapped to a common time scale (e.g., time since ICU admission). At the same time, knowing that an ICU stay is detailed in \texttt{icustays} in MIMIC and in \texttt{cases} in \acrshort{sicdb} allows to define the same semantic reference point in both databases.

The two ID types available in \acrshort{sicdb} are the \texttt{patient} and the \texttt{icustay}. There is no separate demarcation of the \texttt{hospadm}. The observation time for a \texttt{patient} ranges from their first observed admission to their death (if it occurred). \texttt{icustay}s range from the current admission to the ICU until the end of the stay. 

\paragraph*{Calculate origin times for each ID type}

After \texttt{ricu} has been told which ID systems exist in the dataset, it also needs to know when they start and end. Much of this process is automated. For \acrshort{sicdb}, all origin times are already provided in a format suitable for use with \texttt{ricu}, and thus the default behavior is appropriate for them. However, minor adjustments are often necessary. For example, discharge times in \acrshort{sicdb} are not provided as absolute times but in seconds since the start of admission. Such adjustments can be made on a case-by-case basis by subtyping the respective functions (in this case \texttt{id\_win\_helper}) and overwriting the default behavior (\autoref{code:salzburg-helper}). Since \acrshort{sicdb} provides time in seconds since admission, \texttt{ricu} must further be told to work with relative times in seconds. This can be conveniently achieved through existing helper functions (\autoref{code:salzburg-helper}). 

Following the steps above makes \acrshort{sicdb} available and fully usable within \texttt{ricu}. While some further helper functions may be necessary to enable optional functionality such as automatic determination of measurement units (\acrshort{sicdb} stores units in a separate reference table that needs to be merged at runtime), these are not essential to the main functionality of \texttt{ricu}. Note that the above only interfaces \acrshort{sicdb}. It does \textit{not} automatically map all existing clinical concepts for \acrshort{sicdb}. Defining a clinical concept for \acrshort{sicdb} still requires manual mapping of the concept to \acrshort{sicdb} data items, for example via the appropriate measurement IDs (see also the next session on adding a clinical concept). We do not expect that this process can ever be fully avoided (unless it was already performed prior, for example by mapping to and providing the data in OMOP format). However, we found that the framework and helper functions that \texttt{ricu} provides greatly simplify this process. Additional information on \texttt{ricu} and its design principles can be found in \cite{bennettRicuInterfaceIntensive2023}.

\begin{lstlisting}[frame=single, float=h, style=pycharm, language=R, caption={\textit{Helper functions that calculate the origin and temporal relation between events in the Salzburg Intensive Care Database.}}, columns=fullflexible, label=code:salzburg-helper, basicstyle=\ttfamily\tiny]

id_win_helper.sic_env <- function(x) {
  # return a mapping between two ID systems (e.g., ICU stay ID and patient ID), 
  # including relative start and end times
  cfg <- sort(as_id_cfg(x), decreasing = TRUE)
  
  ids <- field(cfg, "id")
  sta <- field(cfg, "start")
  end <- field(cfg, "end")
  
  tbl <- as_src_tbl(x, unique(field(cfg, "table")))
  
  mis <- setdiff(sta, colnames(tbl))
  
  res <- load_src(tbl, cols = c(ids, intersect(sta, colnames(tbl)), end))
  
  assert_that(length(mis) == 1L)
  res[, firstadmission := 0L]
  
  res <- res[, c(sta, end) := lapply(.SD, s_as_mins), .SDcols = c(sta, end)]
  res[, timeofstay := offsetafterfirstadmission + timeofstay] # convert to absolute discharge time
  
  res <- setcolorder(res, c(ids, sta, end))
  res <- rename_cols(res, c(ids, paste0(ids, "_start"),
                            paste0(ids, "_end")), by_ref = TRUE)
  
  as_id_tbl(res, ids[2L], by_ref = TRUE)
}

load_difftime.sic_tbl <- function(x, rows, cols = colnames(x),
                                   id_hint = id_vars(x),
                                   time_vars = ricu::time_vars(x), ...) {
  # Load time differences in SICdb by treating each time variable as 
  # the relative time since admission in seconds
  load_as_relative_time(x, {{ rows }}, cols, id_hint, time_vars, s_as_mins)
}

\end{lstlisting}

\subsection{Define a clinical concept: Potassium chloride}

\acrshort{yaib} comes with a wide range of pre-defined clinical concepts. However, it is likely that new tasks and applications require additional variables. An area of particular interest in this respect are medications. Using the exemplar of Potassium Chloride (a fluid frequently administered in the ICU), we demonstrate how new drugs can be added to \acrshort{yaib} with minimal effort. For many concepts, all that is required is a JSON dict that describes the correct measurement IDs within each dataset (\autoref{code:kcl}). The JSON snippet can then be appended to the existing \texttt{ricu} concept file (for additions that should become part of the main package) or added to the search path of \texttt{load\_dictionary} via the \texttt{cfg\_dirs} parameter. The definition of complicated transformations such as calculation of hourly rates is also supported by existing helper functions. If custom calculations are necessary, they can be provided as user-defined functions via the \texttt{callback} element.

\begin{lstlisting}[frame=single,float=h, style=pycharm, language=json, caption={\textit{JSON definition for rate of potassium chloride across included datasets.}}, columns=fullflexible, label=code:kcl, basicstyle=\ttfamily\tiny]

"kcl_dur": {
    "description": "potassium chloride duration",
    "category": "medications",
    "aggregate": "max",
    "sources": {
      "aumc": [
        {
          "ids": 9001,
          "table": "drugitems",
          "sub_var": "itemid",
          "stop_var": "stop",
          "grp_var": "orderid",
          "callback": "aumc_dur"
        }
      ],
      "eicu": [
        {
          "regex": "^potassium chloride",
          "table": "infusiondrug",
          "sub_var": "drugname",
          "callback": "eicu_duration(gap_length = hours(5L))",
          "class": "rgx_itm"
        }
      ],
      "hirid": [
        {
          "ids": 1000396,
          "table": "pharma",
          "sub_var": "pharmaid",
          "grp_var": "infusionid",
          "callback": "hirid_duration"
        }
      ],
      "miiv": [
        {
          "ids": [225166, 227522],
          "table": "inputevents",
          "sub_var": "itemid",
          "stop_var": "endtime",
          "grp_var": "linkorderid",
          "callback": "mimic_dur_inmv"
        }
      ]
    }
  }
\end{lstlisting}

\subsection{Adapt a task: KDIGO stage}

Cohorts created in \acrshort{yaib} are built to be easily adaptable. For example, predicting ordinal KDIGO stage rather than binary presence/absence of AKI is as simple as changing the outcome variable from \texttt{aki} to \texttt{kdigo}. More substantial changes to the task are also supported. If the general setup (i.e., covariates, exclusion criteria, etc.) remains the same but a novel outcome should be predicted, the outcome should be created as a clinical concept in \texttt{ricu} which can then simply be used as the outcome variable. If covariates, exclusion criteria, or prediction times change, existing helper functions can be utilized. See the existing task definitions for further examples.

\subsection{Adding a new prediction model: random forest, RNN, and Temporal Fusion Transformer}
We allow prediction models to be easily added and integrated into a \acrfull{pl}~\citep{falcon2019pytorch} module. This incorporates advanced logging and debugging capabilities, as well as built-in parallelism. Our interface derives from the PL \texttt{BaseModule}\footnote{\url{https://lightning.ai/docs/pytorch/stable/common/lightning_module.html}}.

For standard Scikit-Learn type \acrshort{ml} models (e.g., \acrfull{lgbm}~\citep{keLightGBMHighlyEfficient2017}), one can implement the \texttt{MLWrapper}, incorporating the model steps in the process. Note that this class is also derived from the \acrshort{pl}\texttt{BaseModule}; this leads to minimal code overhead. See \autoref{code:ml-model-definition} for details for the implementation of a random forest model. The only needed code here is the hyperparameter configuration and the initialization of the superclass.
\begin{lstlisting}[frame=single, style=pycharm, float=h, caption={\textit{Example ML model definition.} We create a Random Forest classifier that implements the default \acrshort{yaib} prediction model interface}, label=code:ml-model-definition, columns=fullflexible, basicstyle=\ttfamily\tiny]
@gin.configurable
class RFClassifier(MLWrapper):
    _supported_run_modes = [RunMode.classification]

    def __init__(self, *args, **kwargs):
        self.model = self.set_model_args(ensemble.RandomForestClassifier, *args, **kwargs)
        super().__init__(*args, **kwargs)
\end{lstlisting}
\begin{lstlisting}[frame=single, style=pycharm, float=h, caption={\textit{Example DL model definition.} We use the inbuilt Torch RNN layers to built a Recurrent Neural Network. }, label=code:dl-model-definition, columns=fullflexible, basicstyle=\ttfamily\tiny]
@gin.configurable
class RNNet(DLPredictionWrapper):
    """Torch standard RNN model"""
    
    _supported_run_modes = [RunMode.classification, RunMode.regression]

    def __init__(self, input_size, hidden_dim, layer_dim, num_classes, *args, **kwargs):
        super().__init__(
            input_size=input_size, hidden_dim=hidden_dim, layer_dim=layer_dim, num_classes=num_classes, *args, **kwargs
        )
        self.hidden_dim = hidden_dim
        self.layer_dim = layer_dim
        self.rnn = nn.RNN(input_size[2], hidden_dim, layer_dim, batch_first=True)
        self.logit = nn.Linear(hidden_dim, num_classes)

    def init_hidden(self, x):
        h0 = x.new_zeros(self.layer_dim, x.size(0), self.hidden_dim)
        return h0

    def forward(self, x):
        h0 = self.init_hidden(x)
        out, hn = self.rnn(x, h0)
        pred = self.logit(out)
        return pred
\end{lstlisting}

The definition of \acrshort{dl} models can be done by creating a subclass from the  \texttt{DLPredictionWrapper}; this inherits the standard methods needed for training DL learning models. Again, our implementation using \acrshort{pl} significantly reduces the code overhead and complexity. See \autoref{code:dl-model-definition} for the example of a simple RNN model. We can then create a gin configuration file for this model such as that in \autoref{code:tft-model-gin} to specify default parameters and hyperparameter ranges for hp-tuning.

More advanced, or state-of-the-art, models are also easily implemented. One of \acrshort{yaib}'s users has implemented a Temporal Fusion Transformer architecture~\citep{limTemporalFusionTransformers2021}. This model provides good performance on multi-horizon time series forecasting, as well as interpretable insights. Specifically, \cite{limTemporalFusionTransformers2021} describe the TFT as follows: \textit{"(1) examining the importance of each input variable in prediction, (2) visualizing persistent temporal patterns, and (3) identifying any regimes or events that lead to significant changes in temporal dynamics"}. See \autoref{code:tft-model-definition} for implementation details. This implementation is based upon the NVIDIA PyTorch implementation\footnote{\url{https://github.com/NVIDIA/DeepLearningExamples/tree/master/PyTorch/Forecasting/TFT}}.
\begin{lstlisting}[frame=single, style=pycharm, float=h, caption={\textit{Temporal Fusion Transformer model definition.} Note that this implementation is similar to existing code and can use existing methods and the original dataloader of \acrshort{yaib} which avoids code duplication.}, label=code:tft-model-definition, columns=fullflexible, basicstyle=\ttfamily\tiny]
class TFT(DLPredictionWrapper):
    """ 
    Implementation of Temporal Fusion Transformer, https://arxiv.org/abs/1912.09363.
    """


    _supported_run_modes = [RunMode.classification, RunMode.regression]
    def __init__(
        self,
        num_classes, # Classes for multiclass classification
        encoder_length,  # Determines interval to use for prediction
        hidden, # Amount of hidden layers
        dropout, # Dropout layers
        n_heads, # Attention heads
        dropout_att, 
        example_length,  # Determines interval to predict
        quantiles=[0.1, 0.5, 0.9],  # quantiles to produce
        static_categorical_inp_size=[2],  # Number of categories
        temporal_known_categorical_inp_size=[],
        temporal_observed_categorical_inp_size=[48],  # Number of categorical observed variables
        static_continuous_inp_size=3,  # Number of static continuous variables
        temporal_known_continuous_inp_size=0,
        temporal_observed_continuous_inp_size=48,
        temporal_target_size=1,  # Number of target variables
        **kwargs,
    ):
    
        #derived variables
        num_static_vars = len(static_categorical_inp_size) + static_continuous_inp_size
        num_future_vars = len(temporal_known_categorical_inp_size) + temporal_known_continuous_inp_size
        num_historic_vars = sum([num_future_vars, temporal_observed_continuous_inp_size, temporal_target_size,      len(temporal_observed_categorical_inp_size),])
        
        super().__init__(num_classes=num_classes, encoder_length=encoder_length, hidden=hidden,
                 n_heads=n_heads, dropout_att=dropout_att, example_length=example_length, quantiles=quantiles,                 num_static_vars=num_static_vars, num_future_vars=num_future_vars, num_historic_vars=num_historic_vars, *args,static_categorical_inp_size=1, temporal_known_categorical_inp_size=0, temporal_observed_categorical_inp_size=48, static_continuous_inp_size=3, temporal_known_continuous_inp_size=0, temporal_observed_continuous_inp_size=48, temporal_target_size=1, **kwargs)

        
        
        self.encoder_length = encoder_length # Determines from how distant past we want to use data from

        self.embedding = LazyEmbedding(static_categorical_inp_size, temporal_known_categorical_inp_size, temporal_observed_categorical_inp_size, static_continuous_inp_size, temporal_known_continuous_inp_size, temporal_observed_continuous_inp_size, temporal_target_size,hidden) # embeddings for all variables
        
        self.static_encoder = StaticCovariateEncoder(num_static_vars, hidden, dropout)  # encoding for static variables
        self.TFTback = TFTBack(encoder_length, num_historic_vars, hidden, dropout, num_future_vars, n_heads, dropout_att, example_length,quantiles)
        self.logit = nn.Linear(len(quantiles), num_classes) # Linear layer to output to the number of classes and allow modification by predictionwrapper.
\end{lstlisting}
To train and optimize this model from a choice of hyperparameters, we need to specify a GIN file to bind the parameters, see \autoref{code:tft-model-gin}. Note that we can use modifiers for the optimizer (e.g, Adam optimizer) and ranges that we can specify in rounded brackets "()". Square brackets, "[]",  result in a random choice where the variable is uniformly sampled. 
\begin{lstlisting}[frame=single, style=pycharm, float=h, caption={\textit{Temporal Fusion Transformer parameter configuration.} }, label=code:tft-model-gin, columns=fullflexible, basicstyle=\ttfamily\tiny]
# Hyperparameters for TFT model.

# Common settings for DL models
include "configs/prediction_models/common/DLCommon.gin"

# Optimizer params
train_common.model = @TFT

optimizer/hyperparameter.class_to_tune = @Adam
optimizer/hyperparameter.weight_decay = 1e-6
optimizer/hyperparameter.lr = (1e-5, 3e-4)

# Encoder params
model/hyperparameter.class_to_tune = @TFT
model/hyperparameter.encoder_length = 24
model/hyperparameter.hidden = 256
model/hyperparameter.num_classes = %NUM_CLASSES
model/hyperparameter.dropout = (0.0, 0.4)
model/hyperparameter.dropout_att = (0.0, 0.4)
model/hyperparameter.n_heads =4
model/hyperparameter.example_length=25
\end{lstlisting}

\subsection{Adding a new imputation method: CSDI}
We have added a range of imputation methods to \acrshort{yaib}, including interfaces to existing imputation libraries~\citep{duPyPOTSPythonToolbox2023, jarrettHyperImputeGeneralizedIterative2022}. Here, we describe the addition of a recently introduced method that uses conditional score-based diffusion models conditioned on observed data: the \ac{csdi}\citep{tashiroCSDIConditionalScorebased2021a}. To make the process of implementing these models easier, we have created the \texttt{ImputationWrapper} class that extends the pre-existing \texttt{DLWrapper} (itself a subclass of the \texttt{LightningModule} of Pytorch-lightning) with extra functionality. 

The \acrshort{csdi} model is a diffusion model that follows the general architecture of conditional diffusion models\citep{hoDenoisingDiffusionProbabilistic2020}; It introduces noise into a subset of time series data used as conditional observations to later denoise the data and predict accurate values for the imputation targets. \ac{csdi} is based on a U-Net architecture\citep{ronnebergerUNetConvolutionalNetworks2015} including residual connections.

\begin{figure}[h]
\centering
\includegraphics[width=0.8\textwidth]{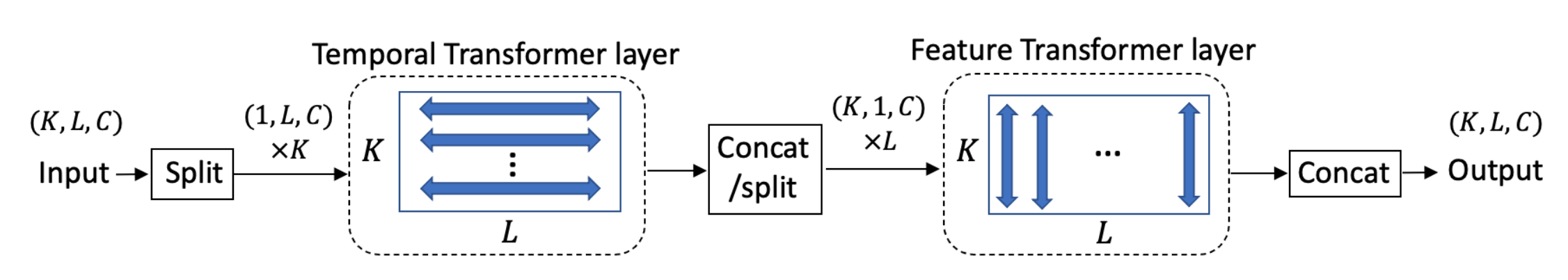}
\caption{The attention mechanism of \ac{csdi} adapted from \cite{tashiroCSDIConditionalScorebased2021a}.}
\label{fig:csdi_attention}
\end{figure}

\cite{tashiroCSDIConditionalScorebased2021a} included two additional features into their model, which are inspired by DiffWave~\citep{kongDiffWaveVersatileDiffusion2021}: an attention mechanism and the ability to input side information. The attention mechanism uses transformer layers, as shown in \autoref{fig:csdi_attention}. An input with K features, L length, and C channels is reshaped first to apply temporal attention and later reshaped again to apply feature attention. The second additional feature allows side information to be used as input to the model by a categorical feature embedding \citep{tashiroCSDIConditionalScorebased2021a}.

See \autoref{code:csdi-init} for the most important implementation code: the model initialization. We note that of this code, very little has been adapted from the original code repository\footnote{\url{https://github.com/ermongroup/CSDI/tree/main}} included in the original publication~\citep{tashiroCSDIConditionalScorebased2021a}.

\begin{lstlisting}[frame=single, float=h, style=pycharm, language=Python, caption={\textit{Implementing the CSDI architecture in YAIB}. Note that %\href{https://github.com/rvandewater/YAIB/blob/development/icu_benchmarks/imputation/csdi.py}{our implementation}
our implementation is very similar to the \href{https://github.com/ermongroup/CSDI/tree/main}{original github repository}, which demonstrates the flexibility of implementing new models in YAIB.}, columns=fullflexible, label=code:csdi-init, basicstyle=\ttfamily\tiny]
{
    def __init__(
        self, input_size, time_step_embedding_size, feature_embedding_size, unconditional, target_strategy, num_diffusion_steps, diffusion_step_embedding_dim, n_attention_heads, num_residual_layers, noise_schedule, beta_start, beta_end, n_samples, conv_channels, *args, **kwargs,
    ):
        super().__init__(...) 
        self.target_dim = input_size[2]
        self.n_samples = n_samples

        self.emb_time_dim = time_step_embedding_size
        self.emb_feature_dim = feature_embedding_size
        self.is_unconditional = unconditional
        self.target_strategy = target_strategy

        self.emb_total_dim = self.emb_time_dim + self.emb_feature_dim
        if not self.is_unconditional:
            self.emb_total_dim += 1  # for conditional mask
        self.embed_layer = nn.Embedding(num_embeddings=self.target_dim, embedding_dim=self.emb_feature_dim)

        input_dim = 1 if self.is_unconditional else 2
        self.diffmodel = diff_CSDI(
            conv_channels,
            num_diffusion_steps,
            diffusion_step_embedding_dim,
            self.emb_total_dim,
            n_attention_heads,
            num_residual_layers,
            input_dim,
        )

        # parameters for diffusion models
        self.num_steps = num_diffusion_steps
        if noise_schedule == "quad":
            self.beta = np.linspace(beta_start**0.5, beta_end**0.5, self.num_steps) ** 2
        elif noise_schedule == "linear":
            self.beta = np.linspace(beta_start, beta_end, self.num_steps)

        self.alpha_hat = 1 - self.beta
        self.alpha = np.cumprod(self.alpha_hat)
        self.alpha_torch = torch.tensor(self.alpha).float().unsqueeze(1).unsqueeze(1)

}
\end{lstlisting}

\subsection{Adding an evaluation metric: Jensen Shannon Divergence and Binary Fairness}
\label{app:ext-eval}
We support adding multiple types of evaluation metrics for benchmarking \acrshort{dl} or \acrshort{ml} models. We additionally support three common metric libraries: TorchMetrics~\citep{nickiskaftedetlefsenTorchMetricsMeasuringReproducibility2022}, Ignite~\citep{pytorch-ignite}, and Scikit-Learn~\citep{scikit-learn}. Adding a metric is a straightforward procedure. We added the Jensen Shannon Divergence (JSD) with the help of the SciPy library~\citep{2020SciPy-NMeth}. See \autoref{code:metric-jsd} for details.
\begin{lstlisting}[frame=single, float=h, style=pycharm, language=Python, caption={\textit{Implementing JSD using SciPy in YAIB.} In this case we used the Ignite interface, but users can also choose to extend from the TorchMetric or SK-Learn interface. }, columns=fullflexible, label=code:metric-jsd, basicstyle=\ttfamily\tiny]
class JSD(EpochMetric):
    def __init__(self, output_transform: Callable = lambda x: x, check_compute_fn: bool = False) -> None:
        super(JSD, self).__init__(lambda x, y: JSD_fn(x, y), output_transform=output_transform, check_compute_fn=check_compute_fn)

        def JSD_fn(y_preds: torch.Tensor, y_targets: torch.Tensor):
            return jensenshannon(abs(y_preds).flatten(), abs(y_targets).flatten()) ** 2
\end{lstlisting}

One can then add the metric to be evaluated for a particular, see \autoref{code:metric-constants}.

In order to asses fairness within ML prediction, a common metric to check is group fairness. This is computed through the ratio between positivity rates and true positives rates for different groups. Two types of these metrics are demographic parity~\citep{caldersBuildingClassifiersIndependency2009} and equal opportunity ratio~\citep{hardtEqualityOpportunitySupervised2016}. 
The TorchMetrics~\citep{nickiskaftedetlefsenTorchMetricsMeasuringReproducibility2022} library includes the group fairness module interface which we can adapt for use in \acrshort{yaib}. 
We use a wrapper, that extends the TorchMetrics implementation and extracts a "group tensor" that indicates to which group the sample belongs. See \autoref{code:metric-fairness} to achieve the desired result.
\begin{lstlisting}[frame=single, float=h, style=pycharm, language=Python, caption={\textit{Adding a wrapper for Group fairness metric.}}, columns=fullflexible, label=code:metric-fairness, basicstyle=\ttfamily\tiny]
class BinaryFairnessWrapper(BinaryFairness):
    """
        This class is a wrapper for the BinaryFairness metric from TorchMetrics.
    """
    group_name = None
    def __init__(self, group_name = "sex", *args, **kwargs) -> None:
        self.group_name = group_name
        super().__init__(*args, **kwargs)
    def update(self, preds, target, data, feature_names) -> None:
        """" Standard metric update function"""
        groups = data[:, :, feature_names.index(self.group_name)]
        group_per_id = groups[:, 0]
        return super().update(preds=preds.cpu(),
                              target=target.cpu(),
                              groups=group_per_id.long().cpu())
\end{lstlisting}
After this, we pass the data and feature names in the training step, as the function requires information about the assigned groups of the dataset; here, we track demographic parity by default, i.e., we use "sex" as the group name. In our pipeline, this has been one-hot encoded to 0 or 1. Then, adding this to the constants file, as seen in \autoref{code:metric-constants}, automatically calculates this metric during the training process.

\begin{lstlisting}[frame=single, float=h, style=pycharm, language=Python, caption={\textit{Metrics recorded for binary classification.} Adding the metrics to this dictionary results in automatic logging. See the whole file for more details.}, columns=fullflexible, label=code:metric-constants, basicstyle=\ttfamily\tiny]
class DLMetrics:
    BINARY_CLASSIFICATION = {
        "AUC": AUROC(task="binary"),
        "PR" : AveragePrecision(task="binary"),
        "F1": F1Score(task="binary", num_classes=2),
        "Calibration_Error": CalibrationError(task="binary",n_bins=10)
        "Calibration_Curve": CalibrationCurve,
        "PR_Curve": PrecisionRecallCurve,
        "RO_Curve": RocCurve,
        "JSD": JSD,
        "Binary_Fairness": BinaryFairnessWrapper(num_groups=2, task='demographic_parity', group_name="sex")
        }
\end{lstlisting}

\newpage
\section{Appendix: Experimental setup and reproducibility}
\label{app:exp_setup}
To reproduce the results obtained in this paper, we have detailed our methodology in this Appendix. 
\subsection{Infrastructure and Hardware}
We used a high-performance computing cluster to perform our experiments
No data was transferred to any external parties in this process. This computing cluster is run with renewable energy and can be considered climate-neutral. The cluster is running the SLURM~\citep{yooSLURMSimpleLinux2003} management tool on Ubuntu 20.04.6 LTS. with a number of Nvidia A100, A40, RTX 2080TI, and RTX Titan GPUs. The traditional machine learning algorithms were trained with Intel Xeon Platinum 8160 and AMD EPYC 7643 CPU resources. 
\begin{table}[h]
    \caption{\textit{Average estimated duration of training tasks}}
    \small
    \label{tab:model-duration}
    \centering
    \begin{tabularx}{\textwidth}{l>{\centering\arraybackslash}X>{\centering\arraybackslash}X>{\centering\arraybackslash}X}
        \toprule
         & \textbf{Once per stay classification} & \textbf{Hourly classification} & \textbf{Hourly regression} \\
        \midrule
        Machine Learning Models & 3 minutes & 10 minutes & 10 minutes\\
        Deep Learning Models & 30 minutes & 60 minutes & 90 minutes\\
        \bottomrule
    \end{tabularx}
\end{table}
\subsection{Libraries}
A full list of libraries is 
available in the \acrshort{yaib} repository; please use the Conda environment manager to install these. We have aimed to use the most recent library versions (while maintaining compatibility) to improve efficiency and reduce errors. At time of writing, the most recent version of Python that supports the used libraries was used: \texttt{Python 3.10}. The most important libraries are: \texttt{Gin-config 0.5.0}, \texttt{Pytorch 2.0} (in combination with Cuda 11.8.0), \texttt{Pytorch-lightning 2.0}, \texttt{Scikit-learn 1.2.2}, \texttt{Lightgbm 3.3.5}, \texttt{Pandas 2.0.0}.
\subsection{Complexity}
\begin{figure}
    \includegraphics[width=0.7\textwidth, angle=90]{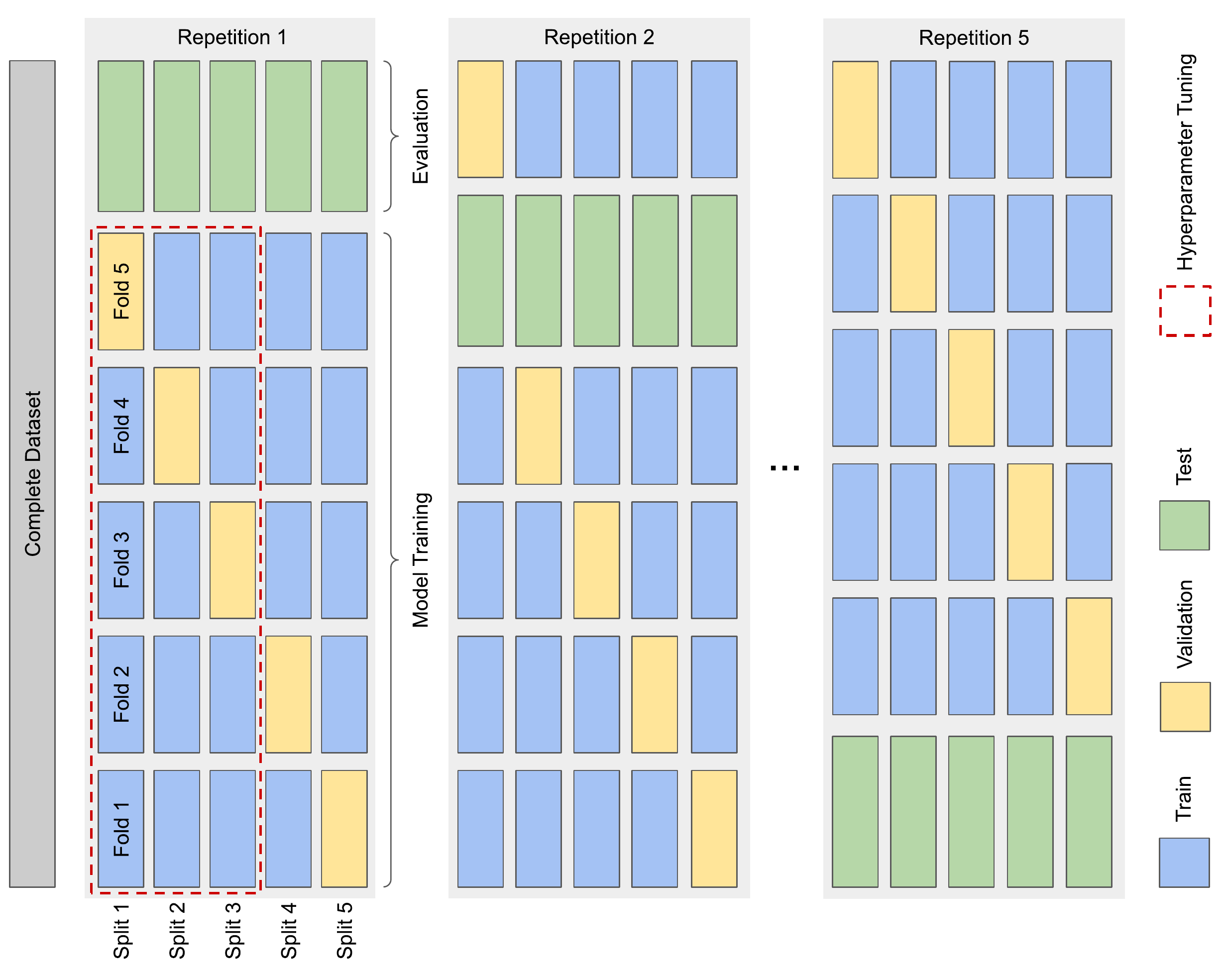}
    \caption[skip=0pt]{\textit{Schematic overview of 5 times repeated 5-fold cross-validation.} This example uses only the first three splits of the first repetition for hyperparameter tuning. The repetitions and amount of folds and the folds to tune on can be easily adjusted (see \autoref{code:lstm-gin}).}
    \label{fig:cv}
\end{figure}

As mentioned in \cite{yecheHiRIDICUBenchmarkComprehensiveMachine2022}, the transformer memory complexity concerning the sequence length is quadratic.   
With our hardware, training deep learning models overall takes less than 3 hours. Training ml models takes less than 1 hour. See \autoref{tab:model-duration} for average training durations. Note that there was large variability between the cohorts due to the size difference of datasets (i.e., \acrlong{eicu} contains almost eight times the amount of stays as \acrlong{aumc}). The algorithmic complexity is specified in the implementation details of the Scikit-learn~\citep{scikit-learn} (\acrfull{lr} and \acrfull{en}), LightGBM~\citep{keLightGBMHighlyEfficient2017} (\acrfull{lgbm}), and PyTorch library~\citep{falcon2019pytorch} (\acrfull{lstm}, \acrfull{rnn}, \acrfull{gru}). 
Furthermore,  \cite{baiEmpiricalEvaluationGeneric2018a} (\acrfull{tcn}) and \citep{yecheHiRIDICUBenchmarkComprehensiveMachine2022} (transformer) are implementations are also detailed in their respective works.
\autoref{fig:cv} shows our repeated cross-validation training method. Note that in order to obtain our final results, we do 5 repetitions of 5 fold cross-validation with an excluded test set.

\subsection{Reproduciblity}
For a detailed description to reproduce our experiments, we refer to the 
Paper reproducibility file (included in the repository). We have followed the standards specified by official ML reproducibility guidelines by Papers with Code\footnote{\url{https://github.com/paperswithcode/releasing-research-code}}.

\subsection{External validation}
All external validation validation models have been trained with an 80/20 train/val split to use as much of the dataset as possible. The test splits are the same as used for the same dataset experiments (diagonal in \autoref{fig:ext_validation})
For the external validation pooled results (d-1), we used subsets of 10,000 stays for each dataset to simulate a setting where the datasets have a similar sample size. We wanted to ensure the size of the datasets, which differs significantly between datasets, had no undue influence on the training.
\subsection{Fine-tuning}
For our fine-tuning experiment, shown in \autoref{fig:fine-tuning} we used an ADAM optimizer with a starting learning rate of 0.00001 and an exponential learning rate scheduler to reduce learning rates gradually. The rest of the hyperparameters are exactly as in the original source models.
\newpage

\section{Hyperparameters}
\label{app:hyperparams}
Here, we detail the tuning setup and hyperparameters used in our experiments. 
\subsection{Tuning approach}

In \autoref{tab:model-hparams-dl} and \ref{tab:model-hparams-ml}, we specify the hyperparameters used for hyperparameter-tuning for the baseline experiments for deep learning and machine learning models, respectively.  
We incorporated different sampling methods for hyperparameter selection. 
Hyperparameters were chosen to be mostly identical to \cite{yecheHiRIDICUBenchmarkComprehensiveMachine2022}, to improve comparability and for reproducibility reasons. However, we have chosen to allow for continuous ranges of hyperparameters in some cases, to improve the performance and functionality of \acrshort{yaib}.
\begin{table}[h]
        \caption{\textit{Model hyperparameters, default values for all models (above), and the distributions for the DL models (below), considered during Bayesian hyperparameter optimization.}}
    \small
    \label{tab:model-hparams-dl}
    \centering
    \begin{tabularx}{\textwidth}{lX>{\centering\arraybackslash}X}
        \toprule
         \textbf{Category} & \textbf{Parameter} & \textbf{Value} \\
        \midrule
        \multirow{3}{*}{Hyperparameter search} 
        & Initial points & 5 \\
        & Calls & 30  \\ 
        & Folds to tune on & 2 \\
        \midrule
        \multirow{5}{*}{General Parameters}
         & Epochs & 1000 \\
         & Min delta & 1E-4 \\
         & Patience & 10 \\
         & Batch size & 64 \\
         & Weight & Balanced \\
         \midrule
         \multirow{2}{*}{Loss}
         & Regression task & Mean Squared Error \\
         & Classification task & Cross-entropy  \\
         \midrule
         \multirow{3}{*}{Optimizer Parameters}
         & Optimizer & Adam  \\
         & Weight decay & 1e-6  \\  
         & Learning rate & Uniform([1E-5, 3E-4]) \\
        \midrule \midrule
        \textbf{Model} & \textbf{Parameter} & \textbf{Value} \\
        \midrule
        \multirow{2}{*}{LSTM} & Hidden dimension & Log-uniform([32, 128]) \\
        & Hidden dimension & RandomInt(1, 3) \\
        & Dropout probability & Uniform([0.0, 0.4]) \\
        \midrule
        \multirow{2}{*}{GRU} & Hidden dimension & Log-uniform([32, 128]) \\
        & Number of layers & RandomInt(1, 3) \\
        \midrule
        \multirow{4}{*}{TCN} 
        & Hidden dimension & Log-uniform([32, 128]) \\
        & Number of layers & Log-uniform([32, 256]) \\
        & Kernel size & Log-uniform([2, 32]) \\
        & Horizon & 24 \\
        \midrule
        \multirow{8}{*}{Transformer} & Hidden dimension & Log-Uniform([32, 128]) \\
        & Number of layers & RandomInt(1, 10) \\
        & Number of heads & Log-uniform(1, 8) \\
        & Depth & Uniform([1,3]) \\
        & Kernel size & Log-uniform([2, 32])\\
        & Dropout probability & Uniform([0.0, 0.4]) \\
        & Dropout attention & Uniform([0, 0.4])\\
        & L1 regularization & 0.0 \\
        & Hidden multiplication & 2 \\
        \bottomrule
    \end{tabularx}
\end{table}
\begin{table}[h]
        \caption{\textit{Model hyperparameters, default values for all models (above), and distributions for the ML models (below), considered during Bayesian hyperparameter optimization.}}
    \small
    \label{tab:model-hparams-ml}
    \centering
    \begin{tabularx}{\textwidth}{lX>{\centering\arraybackslash}X}
        \toprule
         \textbf{Category} & \textbf{Parameter} & \textbf{Value} \\
        \midrule
        \multirow{3}{*}{Hyperparameter search} 
        & Initial points & 10 \\
        & Calls & 50  \\ 
        & Folds to tune on & 3 \\
        \midrule
        \multirow{2}{*}{General Parameters}
         & Patience & 10 \\
         & Jobs & 8 \\
         \midrule
         \multirow{2}{*}{Loss}
         & Regression task & Logloss \\
         & Classification task & Cross-entropy  \\
        \midrule \midrule
        \textbf{Model} & \textbf{Parameter} & \textbf{Value} \\
        \midrule
        \multirow{5}{*}{LR} & C & Log-uniform([1E-3, 1E1]) \\
        & Penalty & Choice(l1, l2, elasticnet) \\
        & L1 Ratio & Uniform([0.0, 1.0])\\
        & Solver & saga\\
        & Max iterations & 100000 \\
        \midrule
        \multirow{7}{*}{EN} & Alpha & Log-uniform([1E-2, 1E1]) \\
        & Tol & Log-uniform([1E-5, 1E-1]) \\
        & Hidden dimension & RandomInt(1, 3) \\
        & Dropout probability & Uniform([0.0, 0.4]) \\
        & L1 Ratio & Uniform([0.0, 1.0])\\
        & Solver & saga\\
        & Max iterations & 10000 \\
        \midrule
        \multirow{7}{*}{LGBM} & Column sample & Uniform([0.33, 1.0]) \\
        & Sub sample & Uniform([0.33, 1.0]) \\
        & Leaves & Log-uniform([8, 128])\\
        & Max depth & RandomInt(3, 7) \\
        & Hidden dimension & RandomInt(1, 3) \\
        & Estimators & 10000 \\
        & Min child samples & 1000 \\
        & Subsample frequency & 1 \\
        \bottomrule
    \end{tabularx}
\end{table}
Log-uniform means that the parameters are sampled according to the reciprocal distribution: 
\begin{equation}
    f(x;a,b)={\frac {1}{x[\log _{e}(b)-\log _{e}(a)]}}\quad {\text{ for }}a\leq x\leq b{\text{ and }}a>0.
\end{equation}
Uniform means that the parameters are sampled according to the uniform distribution:
\begin{equation}
    f(x)={\begin{cases}{\frac {1}{b-a}}&\mathrm {for} \ a\leq x\leq b,\\[8pt]0&\mathrm {for} \ x<a\ \mathrm {or} \ x>b\end{cases}}
\end{equation}

\FloatBarrier

\subsection{Deep Learning Models}
\label{subapp:dl}
We detail the hyperparameters that have been chosen using our Bayesian hyperparameter optimization approach. 
\newpage
\paragraph{\acrfull{gru}} The range of hyperparameters considered for the \acrshort{gru} model are found in \autoref{tab:gru-hyperparams}.
\begin{table}[h]
\caption{\textit{Chosen hyperparameters for \acrfull{gru}.}}
\small
\begin{tabularx}{\textwidth}{l>{\arraybackslash\centering}X>{\arraybackslash\centering}X>{\centering\arraybackslash}X}
\toprule
\textbf{Dataset}   & \textbf{Learning Rate} & \textbf{Layer Dimension} & \textbf{Hidden Dimension} \\
\midrule \midrule
\textbf{Mortality} &               &                 &                  \\
AUMC      & 3.00E-04      & 3               & 48               \\
HiRID     & 2.37E-04      & 2               & 52               \\
eICU      & 3.00E-04      & 1               & 135              \\
MIMIC-IV  & 1.43E-04      & 2               & 77               \\
\midrule \midrule
\textbf{AKI}       &               &                 &                  \\
AUMC      & 2.81E-04      & 3               & 256              \\
HiRID     & 2.06E-04      & 3               & 115              \\
eICU      & 1.43E-04      & 3               & 240              \\
MIMIC-IV  & 2.82E-04      & 3               & 139              \\
\midrule \midrule
\textbf{Sepsis}    &               &                 &                  \\
AUMC      & 2.28E-04      & 2               & 77               \\
HiRID     & 3.00E-04      & 3               & 59               \\
eICU      & 8.51E-05      & 2               & 77               \\
MIMIC-IV  & 2.39E-04      & 3               & 52               \\
\midrule \midrule
\textbf{KF} &          &   &     \\
AUMC       & 3.00E-04 & 3 & 93  \\
HiRID      & 1.11E-04 & 3 & 196 \\
eICU       & 6.35E-05 & 3 & 196 \\
MIMIC-IV   & 1.11E-04 & 1 & 148 \\
\midrule \midrule
\textbf{LoS}       &               &                 &                  \\
AUMC      & 5.45E-05      & 3               & 158              \\
HiRID     & 1.00E-05      & 1               & 117              \\
eICU      & 1.03E-05      & 3               & 254              \\
MIMIC-IV  & 1.57E-05      & 2               & 237             \\
\bottomrule
\end{tabularx}
\label{tab:gru-hyperparams}
\end{table}
\newpage

\paragraph{\acrfull{lstm}} The range of hyperparameters considered for the \acrshort{lstm} model are found in \autoref{tab:lstm-hyperparams}.
\begin{table}[h]
\small
\caption{\textit{Chosen hyperparameters for \acrfull{lstm}.}}
\begin{tabularx}{\textwidth}{l>{\arraybackslash\centering}X>{\arraybackslash\centering}X>{\centering\arraybackslash}X}
\toprule
\textbf{Dataset}   & \textbf{Learning Rate} & \textbf{Layer Dimension} & \textbf{Hidden Dimension} \\
\midrule \midrule
\textbf{Mortality} &               &                 &                  \\
AUMC      & 1.87E-04      & 1               & 145              \\
HiRID     & 1.54E-04      & 2               & 256              \\
eICU      & 3.00E-04      & 3               & 149              \\
MIMIC-IV  & 3.00E-04      & 2               & 185              \\
\midrule \midrule
\textbf{AKI}       &               &                 &                  \\
AUMC      & 2.62E-04      & 3               & 57               \\
HiRID     & 3.00E-04      & 3               & 54               \\
eICU      & 3.00E-04      & 3               & 70               \\
MIMIC-IV  & 3.00E-04      & 3               & 256              \\
\midrule \midrule
\textbf{Sepsis}    &               &                 &                  \\
AUMC      & 2.10E-04      & 1               & 153              \\
HiRID     & 2.48E-04      & 1               & 139              \\
eICU      & 1.12E-04      & 2               & 40               \\
MIMIC-IV  & 2.46E-04      & 1               & 161              \\
\midrule \midrule
\textbf{KF} &          &   &     \\
AUMC       & 2.79E-04 & 2 & 81  \\
HiRID      & 3.00E-04 & 1 & 256 \\
eICU       & 1.75E-04 & 3 & 33  \\
MIMIC-IV   & 2.49E-04 & 1 & 256\\
\midrule \midrule
\textbf{LoS}       &               &                 &                  \\
AUMC      & 3.24E-05      & 1               & 62               \\
HiRID     & 6.65E-05      & 3               & 255              \\
eICU      & 2.86E-05      & 3               & 215              \\
MIMIC-IV  & 1.80E-05      & 3               & 253        \\
\bottomrule
\end{tabularx}
\label{tab:lstm-hyperparams}
\end{table}
\newpage
\paragraph{\acrfull{tcn}} The range of hyperparameters considered for the \acrshort{tcn} model are found in \autoref{tab:tcn-hyperparams}.
\begin{table}[h]
\small
\caption{\textit{Chosen hyperparameters for \acrfull{tcn}.}}
\begin{tabularx}{\textwidth}{l>{\arraybackslash\centering}X>{\arraybackslash\centering}X>{\arraybackslash\centering}X>{\centering\arraybackslash}X}
\toprule
\textbf{Dataset}   & \textbf{Learning Rate} & \textbf{Dropout}  & \textbf{Kernel} & \textbf{Number of Channels} \\
\midrule \midrule
\textbf{Mortality} &               &          &        &                  \\
AUMC      & 5.84E-05      & 2.71E-01 & 6      & 92               \\
HiRID     & 2.14E-04      & 1.12E-01 & 6      & 80               \\
eICU      & 2.35E-05      & 1.15E-02 & 23     & 100              \\
MIMIC-IV  & 1.81E-05      & 3.50E-01 & 3      & 130              \\
\midrule \midrule
\textbf{AKI}       &               &          &        &                  \\
AUMC      & 3.00E-04      & 4.00E-01 & 3      & 144              \\
HiRID     & 2.56E-04      & 2.56E-01 & 12     & 168              \\
eICU      & 3.00E-04      & 1.23E-01 & 3      & 81               \\
MIMIC-IV  & 2.98E-04      & 7.61E-02 & 3      & 249              \\
\midrule \midrule
\textbf{Sepsis}    &               &          &        &                  \\
AUMC      & 3.00E-04      & 0.00E+00 & 2      & 32               \\
HiRID     & 3.00E-04      & 0.00E+00 & 2      & 256              \\
eICU      & 1.93E-04      & 3.98E-01 & 2      & 61               \\
MIMIC-IV  & 2.23E-04      & 1.06E-01 & 4      & 78               \\
\midrule \midrule
\textbf{KF} &          &          &    &     \\
AUMC       & 2.56E-04 & 2.78E-01 & 6  & 169 \\
HiRID      & 1.75E-04 & 2.33E-01 & 3  & 34  \\
eICU       & 2.15E-04 & 1.92E-01 & 5  & 138 \\
MIMIC-IV   & 1.88E-05 & 2.15E-01 & 2  & 33 \\
\midrule \midrule
\textbf{LoS}       &               &          &        &                  \\
AUMC      & 3.00E-04      & 1.92E-01 & 2      & 32               \\
HiRID     & 2.74E-04      & 1.71E-01 & 29     & 43               \\
eICU      & 3.00E-04      & 2.91E-01 & 10     & 32               \\
MIMIC-IV  & 1.57E-04      & 1.67E-01 & 12     & 44               \\    
\bottomrule
\end{tabularx}
\label{tab:tcn-hyperparams}
\end{table}
\newpage
\paragraph{Transformer} The range of hyperparameters considered for the transformer model are found in \autoref{tab:transformer-hyperparams}.
\begin{table}[h]
\small
\caption{\textit{Chosen hyperparameters for transformer.}}
\begin{tabularx}{\textwidth}{l>{\arraybackslash\centering}X>{\arraybackslash\centering}X>{\arraybackslash\centering}X>{\centering\arraybackslash}X>{\centering\arraybackslash}X}
\toprule
\textbf{Dataset}   & \textbf{Learning Rate} & \textbf{Dropout}  & \textbf{Heads} &\textbf{ Hidden Dimension} & \textbf{Depth} \\
\midrule \midrule
\textbf{Mortality} &               &          &       &                  &       \\
AUMC      & 1.29E-04      & 1.32E-01 & 2     & 95               & 2     \\
HiRID     & 1.58E-04      & 0.00E+00 & 1     & 247              & 1     \\
eICU      & 1.00E-05      & 4.00E-01 & 3     & 256              & 1     \\
MIMIC-IV  & 6.18E-05      & 1.65E-01 & 1     & 48               & 3     \\
\midrule \midrule
\textbf{AKI}       &               &          &       &                  &       \\
AUMC      & 1.18E-04      & 9.75E-03 & 8     & 52               & 3     \\
HiRID     & 3.00E-04      & 1.50E-01 & 1     & 154              & 2     \\
eICU      & 1.28E-04      & 1.33E-01 & 2     & 96               & 2     \\
MIMIC-IV  & 1.22E-04      & 4.13E-02 & 2     & 72               & 3     \\
\midrule \midrule
\textbf{Sepsis}    &               &          &       &                  &       \\
AUMC      & 2.61E-04      & 4.39E-02 & 1     & 32               & 3     \\
HiRID     & 3.00E-04      & 0.00E+00 & 1     & 32               & 1     \\
eICU      & 2.76E-05      & 1.05E-02 & 2     & 211              & 2     \\
MIMIC-IV  & 3.53E-05      & 3.67E-01 & 1     & 98               & 3     \\
\midrule \midrule
\textbf{KF} &          &          &   &     &   \\
AUMC       & 1.95E-04 & 4.56E-02 & 1 & 51  & 2 \\
HiRID      & 2.48E-04 & 9.34E-02 & 7 & 160 & 3 \\
eICU       & 2.62E-04 & 2.82E-02 & 1 & 52  & 1 \\
MIMIC-IV   & 1.53E-04 & 8.62E-02 & 5 & 160 & 2 \\
\midrule \midrule
\textbf{LoS}       &               &          &       &                  &       \\
AUMC      & 1.13E-04      & 4.11E-02 & 3     & 76               & 2     \\
HiRID     & 1.88E-05      & 1.88E-05 & 4     & 102              & 1     \\
eICU      & 3.96E-05      & 5.67E-02 & 3     & 172              & 2     \\
MIMIC-IV  & 1.13E-04      & 4.11E-02 & 3     & 76               & 2    \\
\bottomrule
\end{tabularx}
\label{tab:transformer-hyperparams}

\end{table}
\newpage
\subsection{Machine Learning Models}
\label{subapp:ml}
\paragraph{\Acf{lr}}The range of hyperparameters considered for the \acrshort{lr} model are found in \autoref{tab:lr-hyperparams}.
\begin{table}[h]
\small
\caption{\textit{Chosen hyperparameters for \acrfull{lr}}.}
\begin{tabularx}{\textwidth}{l>{\arraybackslash\centering}X>{\arraybackslash\centering}X>{\centering\arraybackslash}X}
\toprule
\textbf{Dataset}   & \textbf{C}        & \textbf{Penalty}    & \textbf{L1 Ratio} \\
\midrule \midrule
\textbf{Mortality} &          &            &  \\
AUMC      & 3.63E-02 & elasticnet & 1.00E+00 \\
HiRID     & 3.45E-02 & l2         & 6.63E-02 \\
eICU      & 2.78E-02 & elasticnet & 1.00E+00 \\
MIMIC-IV  & 2.05E-01 & elasticnet & 1.00E+00 \\
\midrule \midrule
\textbf{AKI}       &          &            &          \\
AUMC      & 1.77E-02 & l1         & 1.00E+00 \\
HiRID     & 1.00E+01 & l1         & 4.16E-01 \\
eICU      & 2.52E-02 & l1         & 5.99E-01 \\
MIMIC-IV  & 1.28E-01 & l1         & 2.98E-01 \\
\midrule \midrule
\textbf{Sepsis}    &          &            &          \\
AUMC      & 4.87E-02 & l1         & 2.21E-01 \\
HiRID     & 2.59E-03 & l2         & 3.74E-01 \\
eICU      & 1.98E-03 & elasticnet & 6.20E-01 \\
MIMIC-IV  & 2.20E-03 & l1         & 1.81E-02 \\
\bottomrule
\end{tabularx}
\label{tab:lr-hyperparams}
\end{table}
\paragraph{\Acf{en}}The range of hyperparameters considered for the \acrshort{en} model are found in \autoref{tab:en-hyperparams}.
\begin{table}[h]
\small
\caption{\textit{Chosen hyperparameters for \acrfull{en}}.}
\begin{tabularx}{\textwidth}{l>{\arraybackslash\centering}X>{\arraybackslash\centering}X>{\centering\arraybackslash}X}
\toprule
\textbf{Dataset}   & \textbf{Alpha}        & \textbf{Tol}    & \textbf{L1 Ratio} \\
\midrule \midrule
\textbf{\acrshort{kf}} &          &          &          \\
AUMC       & 1.04E-02 & 3.60E-02 & 1.22E-03 \\
HiRID      & 1.04E-02 & 3.60E-02 & 1.22E-03 \\
eICU       & 1.05E-02 & 9.57E-04 & 8.09E-03 \\
MIMIC-IV   & 1.00E-02 & 2.45E-03 & 0.00E+00 \\
\midrule \midrule
\textbf{LoS}        &          &          &          \\
AUMC       & 1.00E-02 & 1.67E-05 & 7.66E-02 \\
HiRID      & 1.00E-02 & 1.02E-05 & 2.40E-02 \\
eICU       & 1.00E-02 & 4.82E-02 & 0.00E+00 \\
MIMIC-IV   & 1.00E-02 & 7.42E-02 & 0.00E+00 \\

\bottomrule
\end{tabularx}
\label{tab:en-hyperparams}
\end{table}
\newpage
\paragraph{\acrfull{lgbm}} The range of hyperparameters considered for the \acrshort{lgbm} model are found in \autoref{tab:lgbm-hyperparams}.
\begin{table}[h]
\small
\caption{\textit{Chosen hyperparameters for \acrfull{lgbm}.}}
\begin{tabularx}{\textwidth}{l>{\arraybackslash\centering}X>{\arraybackslash\centering}X>{\centering\arraybackslash}X>{\centering\arraybackslash}X}
\toprule
\textbf{Dataset}   & \textbf{Depth} & \textbf{Column Sample} & \textbf{Leaves} & \textbf{Subsample} \\
\midrule \midrule
\textbf{Mortality} &       &               &        &           \\
AUMC      & 6     & 9.89E-01      & 117    & 9.90E-01  \\
HiRID     & 5     & 1.00E+00      & 8      & 1.00E+00  \\
eICU      & 7     & 5.41E-01      & 128    & 1.00E+00  \\
MIMIC-IV  & 7     & 1.00E+00      & 28     & 1.00E+00  \\
\midrule \midrule
\textbf{AKI}       &       &               &        &           \\
AUMC      & 7     & 1.00E+00      & 110    & 8.69E-01  \\
HiRID     & 7     & 1.00E+00      & 79     & 8.69E-01  \\
eICU      & 7     & 9.97E-01      & 117    & 8.35E-01  \\
MIMIC-IV  & 7     & 1.00E+00      & 128    & 1.00E+00  \\
\midrule \midrule
\textbf{Sepsis}    &       &               &        &           \\
AUMC      & 7     & 1.00E+00      & 128    & 7.83E-01  \\
HiRID     & 4     & 6.18E-01      & 17     & 8.22E-01  \\
eICU      & 7     & 8.97E-01      & 52     & 6.04E-01  \\
MIMIC-IV  & 7     & 1.00E+00      & 128    & 5.84E-01  \\
\midrule \midrule
\textbf{KF}       &       &               &        &           \\
AUMC       & 7 & 7.01E-01 & 15  & 9.99E-01 \\
HiRID      & 3 & 9.58E-01 & 16  & 9.95E-01 \\
eICU       & 7 & 1.00E+00 & 47  & 1.00E+00 \\
MIMIC-IV   & 7 & 9.30E-01 & 33  & 9.96E-01 \\
\midrule \midrule
\textbf{LoS}       &       &               &        &           \\
AUMC       & 7 & 3.30E-01 & 83  & 1.00E+00 \\
HiRID      & 6 & 3.30E-01 & 49  & 5.01E-01 \\
eICU       & 7 & 3.30E-01 & 51  & 3.30E-01 \\
MIMIC-IV   & 7 & 3.30E-01 & 128 & 3.30E-01 \\
\bottomrule
\end{tabularx}
\label{tab:lgbm-hyperparams}
\end{table}
\newcommand{\answerYes}[2][]{\textcolor{blue}{ [Yes] #1 }\textit{#2}}
\newcommand{\answerNo}[2][]{\textcolor{orange}{[No] #1}\textit{#2}}
\newcommand{\answerNA}[2][]{\textcolor{gray}{[N/A] #1}\textit{#2}}
\newcommand{\answerTODO}[2][]{\textcolor{red}{\bf [TODO]}\textit{#2}}

\section{Machine Learning Reproducibility Checklist}
\label{app:checklist}
This checklist (version 2.0) was found to be the most recent version at the time of writing. It can be found at: \url{www.cs.mcgill.ca/~jpineau/ReproducibilityChecklist-v2.0.pdf}.
For all models and algorithms presented, check if you include:
\begin{itemize}
\item A clear description of the mathematical setting, algorithm, and/or model. \answerYes{See \autoref{app:exp_setup} for an overview of the used infrastructure, libraries, complexity, reproducibility. See \autoref{app:hyperparams} for the hyperparameter tuning setup.}
\item A clear explanation of any assumptions. \answerYes{We make limited assumptions about the data we use. Mainly, we assume stays are i.i.d. as is usual in clinical machine learning.}
\item An analysis of the complexity (time, space, sample size) of any algorithm. \answerYes{See \autoref{app:exp_setup}.}
\end{itemize}
For any theoretical claim, check if you include:
\begin{itemize}
\item A clear statement of the claim. \answerNA{We did not make theoretical claims.}
\item A complete proof of the claim.\answerNA{}
\end{itemize}
For all datasets used, check if you include:
\begin{itemize}
\item The relevant statistics, such as number of examples. \answerYes{See \autoref{appendix:datasets} and \autoref{appendix:outc-detail} for details of the data.}
\item The details of train / validation / test splits. \answerYes{See \autoref{subsec:exp_settings}.}
\item An explanation of any data that were excluded, and all pre-processing step. \answerYes{See \autoref{subsec:exp_settings},  \autoref{appendix:datasets}, and \autoref{appendix:outc-detail}.}
\item A link to a downloadable version of the dataset or simulation environment. \answerYes{See 
the included \texttt{yaib-cohorts} code in order to generate the datasets (additionally, the \acrshort{yaib} repository contains preprocessed demo datasets)}
\item For new data collected, a complete description of the data collection process, such as instructions to annotators and methods for quality control. \answerNA{We did not collect new data.}
\end{itemize}
For all shared code related to this work, check if you include:
\begin{itemize}
\item Specification of dependencies. \answerYes{This can be found in \autoref{app:exp_setup} and at our GitHub repository (below).}
\item Training code. \answerYes{This can be found in the main \acrshort{yaib} code (\texttt{cross-validation.py} and \texttt{train.py}. 
Details for the used parameters can be found in \autoref{app:exp_setup}}
\item Evaluation code. \answerYes{See answer above.}
\item (Pre-)trained model(s). \answerYes{Pretrained models for each task are publicly available and can be provided upon request.} 
\item README file includes table of results accompanied by precise command to run to produce those results.\answerYes{We include the results in the main text and \autoref{app:ext-results}. Additionally, we have a reproducibility 
document (\texttt{PAPER.md}) in our repository that follows common guidelines.}
\end{itemize}
For all reported experimental results, check if you include:
\begin{itemize}
\item  The range of hyper-parameters considered, method to select the best hyper-parameter configuration, and specification of all hyper-parameters used to generate results.\answerYes{For the method see \autoref{subsec:training}, for the hyperparameter configuration, see: \autoref{app:hyperparams}. For the individual hyperparameter configurations per model, dataset, and task, see \autoref{subapp:dl} and \autoref{subapp:ml} for ml and dl models, respectively.}
\item The exact number of training and evaluation runs. \answerYes{See \autoref{subsec:exp_settings}.}
\item A clear definition of the specific measure or statistics used to report results.\answerYes{See \autoref{subsec:exp_settings}.}
\item A description of results with central tendency (e.g. mean) \& variation (e.g. error bars). \answerYes{We report the standard deviation of all results, to accurately reflect the precision of our results.}
\item The average runtime for each result, or estimated energy cost.\answerYes{See \autoref{app:exp_setup}.}
\item A description of the computing infrastructure used.\answerYes{See \autoref{app:exp_setup}.}
\end{itemize}

\end{document}